\documentclass[lettersize,journal]{IEEEtran}

\IEEEoverridecommandlockouts                              

\usepackage[T1]{fontenc}

\usepackage{amsmath} 
\usepackage{amssymb} 
\usepackage{amsbsy}

\usepackage{mathtools}
\usepackage{arydshln} 

\usepackage{enumerate}
\usepackage{listings}

\usepackage{bm}

\usepackage{graphicx} 

\usepackage{subcaption} 
\usepackage[font=footnotesize,labelfont=bf]{caption}

\usepackage{hyperref} 

\usepackage{tcolorbox} 

\usepackage{cite}
\usepackage{cleveref} 

\usepackage[ruled,vlined]{algorithm2e}

\usepackage{grffile}

\usepackage{threeparttable} 
\usepackage{tabularx} 
\usepackage{booktabs} 
\usepackage{multirow} 
\usepackage{makecell}  

\usepackage{lipsum} 

\usepackage[switch]{lineno} 

\usepackage{cancel} 
\usepackage[normalem]{ulem} 

\usepackage{circledsteps} 
\newcommand{\CircledOrange}[2][]{%
  \CircledParamOpts{inner color=white, outer color=orange!0!black, fill color=orange!90!black, #1}{1}{#2}%
}

\newcommand{\CircledGreen}[2][]{%
  \CircledParamOpts{inner color=white, outer color=green!0!black, fill color=green!50!black, #1}{1}{#2}%
}

\newcommand{\CircledCyan}[2][]{%
  \CircledParamOpts{inner color=white, outer color=cyan!0!black, fill color=cyan!90!black, #1}{1}{#2}%
}

\newcommand{\CircledMagenta}[2][]{%
  \CircledParamOpts{inner color=black, outer color=magenta!0!black, fill color=magenta!50!white, #1}{1}{#2}%
}

\newcommand{\hyperfootnote}[1][]{\def\ArgI\hyperfootnoteRelay}
\newcommand\hyperfootnoteRelay[2][]{\href{#1#2}{\ArgI}\footnote{\href{#1#2}{#2}}}

\usepackage{orcidlink}

\newif\ifANONREVIEW
\ANONREVIEWfalse 

\begin{document}

\title{Planning and Control for Deformable Linear Object Manipulation}

\author{%
\thanks{Manuscript received 99 Marcober 9999; accepted 99 Aprober 999. Date
of publication 99 Janbery 9999; date of current version 99 Julner 9999. This article was recommended for publication by Associate Editor X. Xxxxx and Editor Y. Yyyyyyy upon evaluation of the reviewers’ comments. %
\ifANONREVIEW
\else
    \textit{(Corresponding author: Burak Aksoy.)}
\fi
} 
\ifANONREVIEW
\else
    Burak Aksoy\orcidlink{0009-0000-1942-3976}~\IEEEmembership{Student Member,~IEEE,},
    John T.~Wen\orcidlink{0000-0002-5123-5411}~\IEEEmembership{Fellow,~IEEE,}
    \thanks{B. Aksoy and J. T.~Wen are with Electrical, Computer, and Systems Engineering, Rensselaer Polytechnic Institute. {\tt \{aksoyb2, wenj\} @rpi.edu}}
\fi
\thanks{This paper has supplementary downloadable multimedia material available
at http://ieeexplore.ieee.org provided by the authors. 
}
\thanks{Digital Object Identifier XX.XXXX/TASE.9999.9999999}
}

\ifANONREVIEW
\else
    \markboth{Submitted to IEEE Transactions on Automation Science and Engineering,~Vol.~XX, No.~X, Augmber~9999}%
    {Aksoy \MakeLowercase{\textit{et al.}}: Planning and Control for Deformable Linear Object Manipulation}
\fi

\IEEEpubid{0000--0000/00\$00.00~\copyright~9999 IEEE}

\maketitle



\begin{abstract} 
\looseness=-1
Manipulating a deformable linear object (DLO) such as wire, cable, and rope is a common yet challenging task due to their high degrees of freedom and complex deformation behaviors, especially in an environment with obstacles. 
Existing local control methods are efficient but prone to failure in complex scenarios, while precise global planners are computationally intensive and difficult to deploy.
This paper presents an efficient, easy-to-deploy framework for collision-free DLO manipulation using mobile manipulators. 
We demonstrate the effectiveness of leveraging standard planning tools for high-dimensional DLO manipulation without requiring custom planners or extensive data-driven models.
Our approach combines an off-the-shelf global planner with a real-time local controller. The global planner approximates the DLO as a series of rigid links connected by spherical joints, enabling rapid path planning without the need for problem-specific planners or large datasets.
The local controller employs control barrier functions (CBFs) to enforce safety constraints, maintain the DLO integrity, prevent overstress, and handle obstacle avoidance. It compensates for modeling inaccuracies by using a state-of-the-art position-based dynamics technique that approximates physical properties like Young's and shear moduli.
We validate our framework through extensive simulations and real-world demonstrations. 
In complex obstacle scenarios—including tent pole transport, corridor navigation, and tasks requiring varied stiffness—our method achieves a 100\% success rate over thousands of trials, with significantly reduced planning times compared to state-of-the-art techniques. 
Real-world experiments include transportation of a tent pole and a rope using mobile manipulators.
We share our ROS-based implementation to facilitate adoption in various applications.


\end{abstract}

\def\abstractname{Note to Practitioners}
\begin{abstract}
\looseness=-1
This work addresses a core challenge in industrial automation: controlling deformable, cable-like objects (DLOs) in confined or cluttered environments, such as wiring in aircraft fuselages or automotive cable-harness assembly. Traditional methods struggle because DLO motion is sensitive to bending and contact forces; purely local controllers often fail in complex scenarios, and specialized global planners can be difficult to implement while also requiring rapid planning speeds.
Our approach combines an off-the-shelf path planner with a real-time controller that enforces collision and stress constraints using physics-based simulations. Practitioners can deploy it on standard robotic arms or mobile manipulators by specifying the workspace geometry and approximate DLO stiffness parameters. The global planner provides a coarse path, while the local controller corrects real-world deviations. Implemented in ROS with open-source code, this is, to our knowledge, the first demonstration of multiple mobile manipulators collaborating smoothly on a single DLO.
Engineers should note that firm grasping is crucial to prevent slippage, and the global planner does not explicitly account for gravity. In many production tasks, this is sufficient if some safety margin around obstacles is defined, since the local controller and simulations factor in gravity and handle nontrivial shape estimation. Future implementations could incorporate active shape feedback and strengthening robustness against dynamic effects. 

\end{abstract}


\section{Introduction}
\label{sec:introduction}
\looseness=-1 


\begin{figure}[!tp]
    \centering
    \includegraphics[width=0.9\columnwidth]{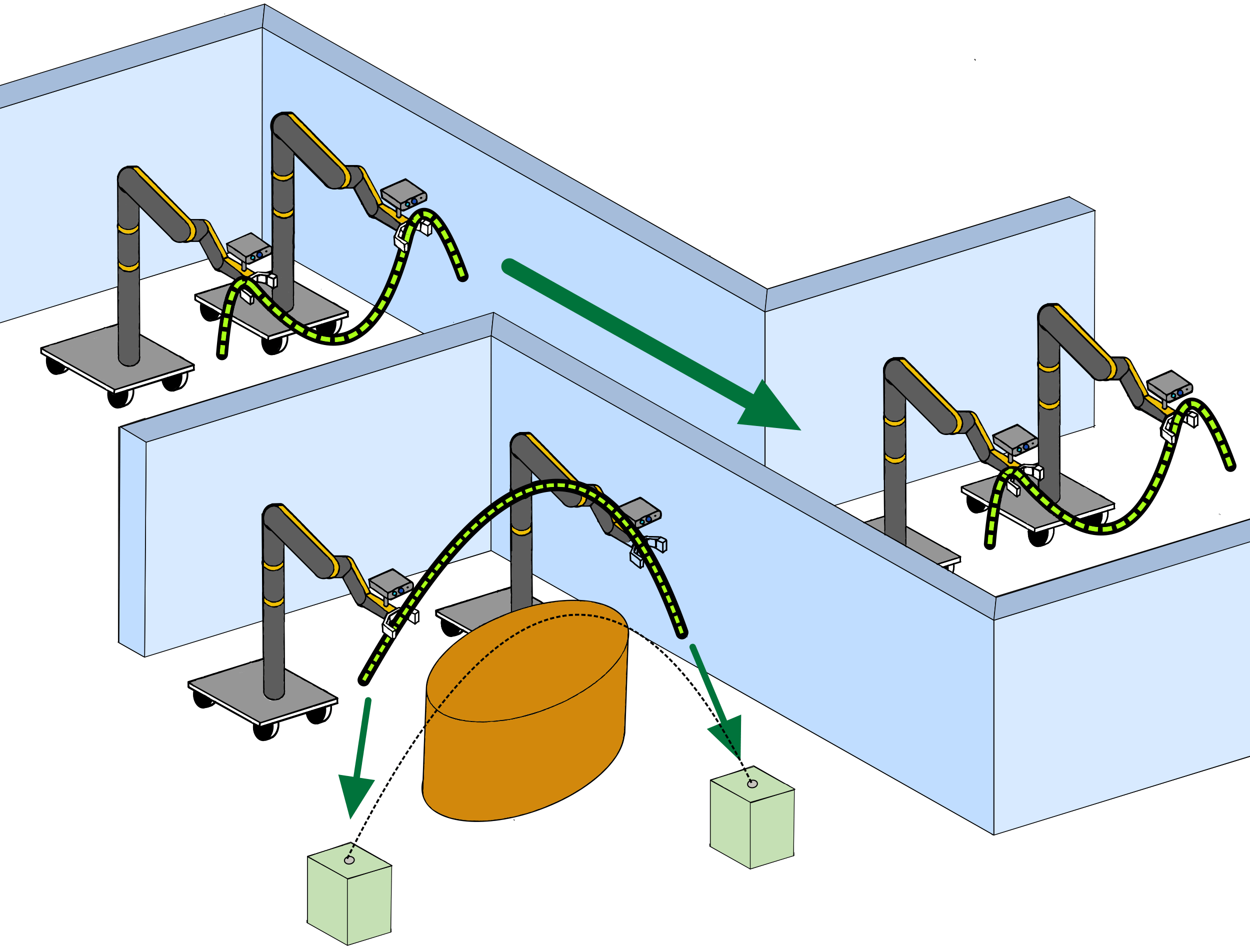}
    \caption{Example deformable linear object manipulation tasks: tent building and heavy cable transportation around obstacles and narrow passageways.} 
    \label{fig:DLO-cartoon}
    \vspace{-0.3cm}
\end{figure}

\IEEEPARstart{C}{onstructing} a tent using flexible poles around obstacles or maneuvering long, heavy cables through narrow passageways (Fig.~\ref{fig:DLO-cartoon}) 
frequently arises in daily life and industrial applications. 
Manipulating deformable linear objects (DLOs), such as rods, wires, cables, and ropes, remains challenging due to the high dimensionality of the problem and variability in material properties\cite{2022zhu}. 
Although multiple robotic manipulators can make these tasks more manageable, developing generalizable and robust solutions remains difficult~\cite{2018sanchez}, especially when accurate prediction of DLO motion is required to match real-world behaviors~\cite{2021yin}.

\IEEEpubidadjcol
This article focuses on safe DLO transportation and shaping in obstacle-rich environments. 
Although deformable objects can adapt their shapes to navigate around obstacles, designing a single planning-and-control framework that accommodates various materials—while satisfying safety requirements like preserving DLO integrity—remains challenging~\cite{2024yu}. 
Existing research has primarily targeted deformation modeling, perception, and shape control \cite{2024liang,2023yu,2024zhaole,2023xiang}, with fewer studies addressing environments with obstacles \cite{2023yu2}. 
\cite{2024aksoy} proposed a local controller using Control Barrier Functions (CBFs) for real-time collision and overstretch avoidance, but noted that task completion is not guaranteed in more complex scenes.
\cite{2023desai} proposed an auxiliary control for CBFs to steer the system away from deadlocks. However, since they did not consider multiple constraints when determining the auxiliary control direction, the system can be steered into other unsafe states. 
Recent attempts to combine local controllers with global planners show promise~\cite{2020mcconachiea,2022mishani,2020roussel,2020sintov,2023yu2,2024yu}, yet these solutions often rely on highly precise global planners that are computationally demanding or make aggressive simplifications that compromise accuracy and generalizability.
In this paper, we propose an efficient, easy-to-deploy, robust, and generalizable framework for DLO manipulation in 3D environments with obstacles. 
Building on the concept of integrating a global offline planner and a local real-time controller, our approach removes the need for developing custom planners. 
Detailed descriptions of the global planning method, local control design, and implementation appear in the following sections.

A key contribution is demonstrating that off-the-shelf planning techniques can effectively and efficiently solve the high-dimensional DLO manipulation problem, without resorting to custom problem-specific planners or data-intensive deformation models. 
We validate our framework through extensive simulations and real-world demonstrations in challenging 3D environments. 
In simulations, we handle tasks such as tent-pole transport, corridor navigation, and other obstacle-dense scenarios involving objects with varied stiffness, achieving a 100\% success rate over thousands of trials and significantly reduced planning times compared to state-of-the-art methods. 
Real-world experiments involve transporting both a tent pole and a rope using mobile manipulators, showcasing the robustness of the framework across different DLOs. 
All necessary planner, controller, and related code implementations are released as ROS packages to support the research community.\footnote{
\ifANONREVIEW
\else
    GitHub link will be here after the review process.
\fi
}

This paper extends the previous work \cite{2024aksoy} with the following improvements: 
(1) Enhanced the ROS-integrated, PBD-based deformable object simulation to handle contacts in complex 3D scenes, provide real-time stress and F/T readings, and report minimum distances to obstacles simultaneously;
(2) introduced the proposed global planning scheme to mitigate the risk of deadlock in difficult environments;
(3) extended the controller to consider grasping point orientations that affect induced stress, and improved overstretch avoidance by using actual stress estimations from advanced simulations and real F/T measurements;
(4) refined online obstacle avoidance by considering all nearby obstacles within a threshold, rather than focusing solely on the closest obstacle; and
(5) conducted more challenging 3D simulations and provided real-world experiments.


\section{Related Work}
\label{sec:related_work}
\looseness=-1 
\subsection{Deformable Linear Object Modeling}
Modeling DLOs undergoing large deformations generally falls into two categories: \textit{analytical models} and \textit{learning-based models}.
\textit{Learning-based approaches} employ neural networks to capture the relationship between control actions and object states without explicit formulations \cite{2021yang, 2020yan, 2022wang, 2021mitrano}. 
While they can model complex DLO behaviors, they require substantial training data and often fail to generalize to untrained DLOs or novel scenarios. For instance, \cite{2023yu} updated a pre-trained state-action Jacobian using online data to track planned paths but relied on 60k simulated samples for offline training.
\textit{Analytical models} such as mass-spring systems, Position-Based Dynamics (PBD), finite-element methods, and As-Rigid-As-Possible (ARAP) \cite{2007sorkine}, extensively reviewed in \cite{2021yin}. 
Notable advancements in PBD, such as XPBD \cite{2016macklin}, convergence improvements with small-step updates \cite{2019macklin}, and developments summarized by \cite{2023fang}, have prompted researchers to adopt PBD in robotic DLO manipulation.
\cite{2024aksoy} integrated these advancements with a discretized Cosserat rod model \cite{2018deul}, achieving realistic simulations with physical parameters like Young's and shear moduli.
While PBD-based models accurately simulate dynamic DLO behaviors, robotics applications have been mostly quasi-static. Recently, \cite{2024floren} addressed dynamic DLO behaviors with PBD, showing promise for more accurate simulation in robotic tasks.

\subsection{Real-to-Sim Matching and Parameter Identification}
\label{subsec:dlo_param_identfication}
Successful DLO manipulation hinges on identifying object properties such as length, mass, and stiffness. While length and mass are straightforward to measure, stiffness parameters (e.g., Young’s and shear moduli) require dedicated identification techniques.
\cite{2024yu} employed Particle Swarm Optimization (PSO) \cite{1995kennedyPSO} for offline parameter identification, using an energy-based DLO model separate from their trained model. While generalizable, this added complexity to their framework. 
\cite{2024tabata} and \cite{2024caporali} used mass-spring models; \cite{2024tabata} iteratively refined parameters using randomly generated values, while \cite{2024caporali} trained a neural network for online estimation. 
Within PBD-based frameworks, \cite{2022liu} implemented XPBD in PyTorch \cite{pytorch} leveraging auto-differentiation for parameter estimation. \cite{2024liang} extended this approach for online estimation in robotic surgery. \cite{2023stuyck} derived analytical differentiation methods for XPBD, enabling efficient gradient computation to minimize the gap between real and simulated behaviors. More recently, \cite{2024floren} estimated DLO parameters from dynamic observations via PBD.

\subsection{Global Planning guided DLO Control}
To address the limitations of local controllers in complex environments, integrating global planning with local control has been explored. Researchers have focused on efficient global planning methods to find stable DLO configurations.\cite{2020mcconachiea, 2020mcconachieb} proposed an open-loop elastic band approach, simplifying models by stretching deformable objects. \cite{2022mishani} introduced a planning method suitable for high-stiffness DLOs, representing configurations on a 6D manifold, which \cite{2020roussel} utilized with a variant of RRT, though with high computation times. \cite{2020sintov} pre-computed a roadmap of stable configurations to reduce online computation, but this lacked adaptability to new DLOs. \cite{2023yu2} found stable configurations by locally minimizing deformation energy but did not consider robot holding point orientations or gravity effects. \cite{2024yu} addressed these limitations by adding an energy-based DLO model alongside a data-driven model, increasing overall complexity. 
In contrast, our approach adopts off-the-shelf planning tools with minimal model complexity, enabling efficient and generalizable DLO manipulation avoiding both custom planning algorithms and large datasets.


\section{Problem Statement}
\label{sec:problem_statement}
\begin{figure}[!tp]
    \centering
    \includegraphics[width=0.9\columnwidth]{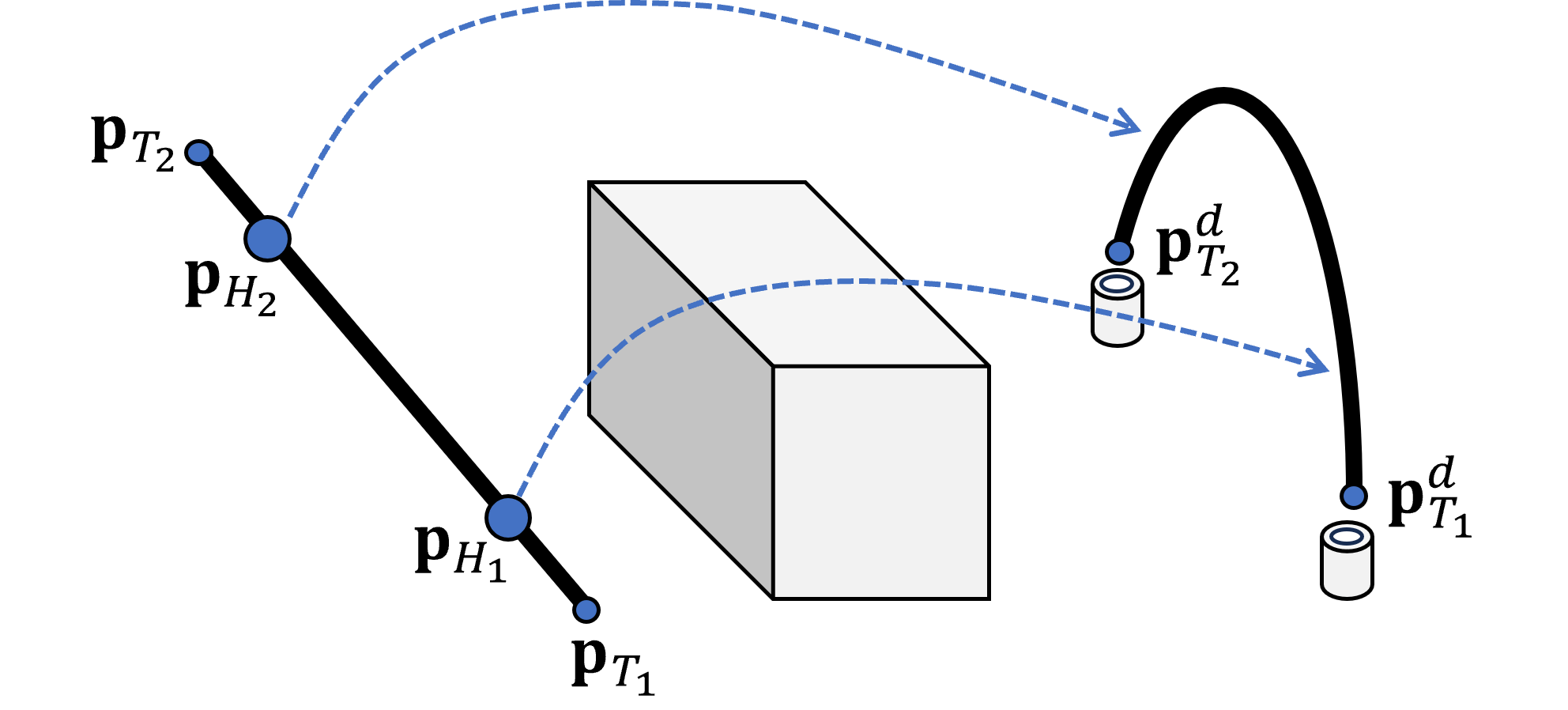}
    \caption{\textbf{Problem description:} Two robots hold a DLO internally, guiding tips to desired poses under stress and obstacle constraints.} 
    \label{fig:problem-definition}
    \vspace{-0.3cm}
\end{figure}

Given an initial state (\( \bm{\Gamma_0} \)) of a DLO in 3D space with obstacles, 
consider two mobile manipulators rigidly grasping the DLO at holding points
\( \mathbf{p}_{H_1}, \mathbf{p}_{H_2} \in SE(3) \) 
(see Fig. \ref{fig:problem-definition}).
The objective is to collaboratively drive the tip points \( \mathbf{p}_{T_1}, \mathbf{p}_{T_2} \in SE(3) \) to desired poses \( \mathbf{p}_{T_1}^d, \mathbf{p}_{T_2}^d \), 
while satisfying safety constraints: avoiding collisions between the DLO and environment and maintaining safe stress levels for both robots and the DLO.
Each robot \(j\) operates on spatial velocity input
\( \mathbf{u}_j = \dot{\mathbf{p}}_{H_j} \in \mathbb{R}^6 \), which controls both linear and angular velocities of the holding point \( \mathbf{p}_{H_j} \). 
The task performance is quantified by pose errors between the tip points and their targets:
\begin{align}
{\mathbf{e}_{T_i} = [\mathbf{e_{\mathbf{x}_i}^\top}, \mathbf{e}_{\mathbf{R}_i}^\top]^\top}\in \mathbb{R}^6,  \quad i=1,2,
\label{eqn:err}
\end{align}
where \(\mathbf{e}_{\mathbf{x}_i} = \mathbf{x}_{T_i} - \mathbf{x}_{T_i}^d\), \(\mathbf{e}_{\mathbf{R}_i} = \sigma ({\mathbf{R}_{T_i}^{d\top}} {\mathbf{R}_{T_i}})\) are position and orientation errors, and \(\sigma\) is a minimal rotation representation. 

We make the following assumptions: 
robot poses are always known via kinematics and sensor fusion;
the DLO state can be estimated in real-time using RGB-D cameras and simulations;
the manipulation process is quasi-static neglecting inertial effects;
either elastic or plastic deformations may occur, but elastoplastic deformations are not considered;
holding points can be anywhere along the DLO, relaxing the common assumption that they must be at the DLO’s endpoints.\footnote{This relaxation is particularly useful for insertion tasks.};
the precise properties of the DLO are not known beforehand; and
obstacles can be of any shape (i.e., no convexity assumption).

\section{Framework Overview}
\label{sec:method_overview}
\begin{figure}[!tp]
    \centering
    \includegraphics[width=1.00\columnwidth]{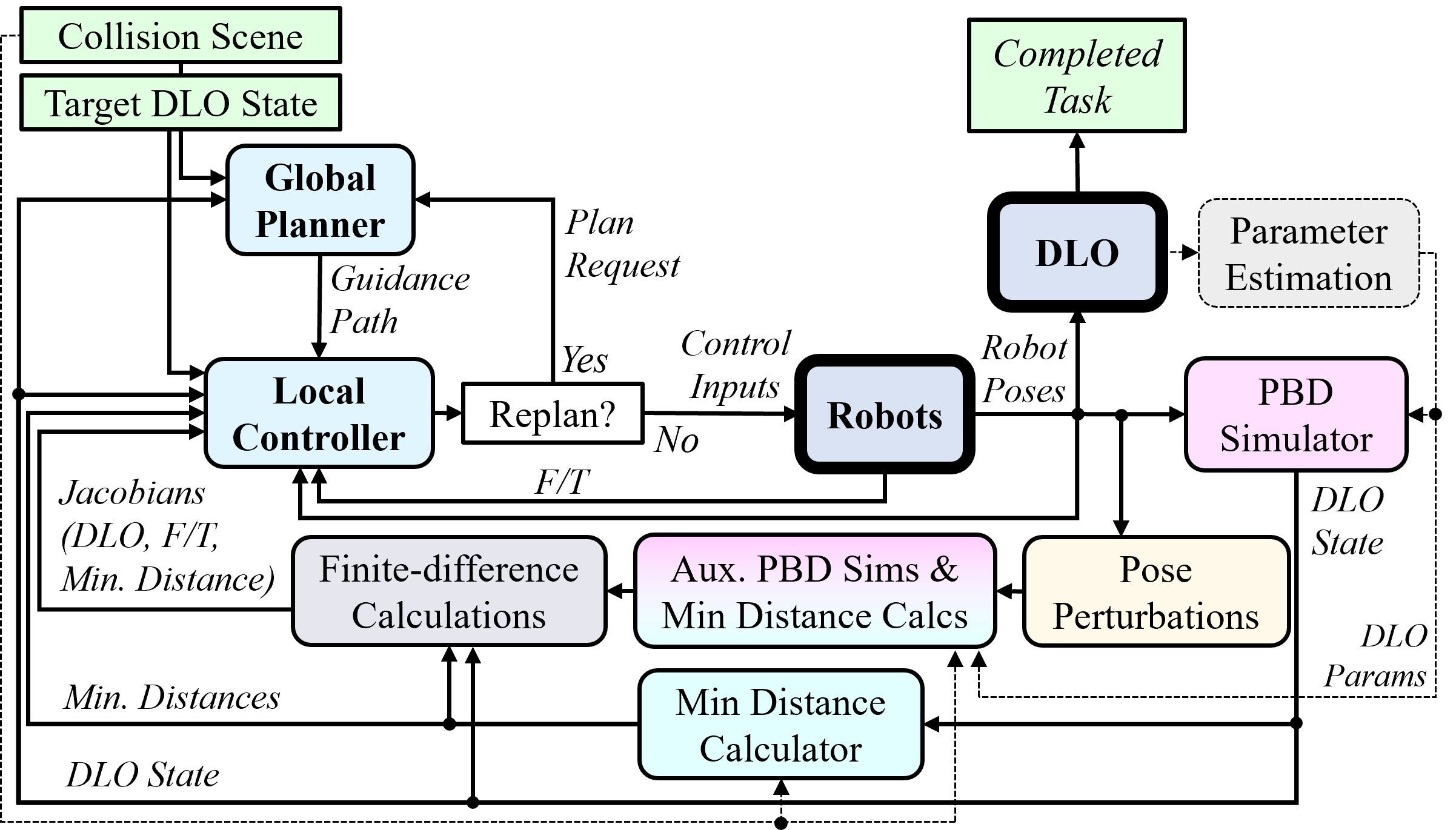}
    \caption{Proposed \textbf{DLO manipulation framework} combining global planning and local control. The local controller uses PBD-based simulation and the guidance path to compute safe robot commands.} \label{fig:method-overview}
    \vspace{-0.3cm}
\end{figure}

\looseness=-1
Our framework integrates a global offline planner and a local real-time controller within a unified ROS-based system (Fig.~\ref{fig:method-overview}), aiming for efficient, safe, and generalizable DLO manipulation in 3D obstacle-rich environments. Its main components are:
\paragraph{Global Planner}
Given a collision scene and desired object states, the global planner rapidly computes a coarse guidance path. Implemented in Python, it leverages the Tesseract \cite{tesseract} framework to deploy popular planning algorithms such as RRT \cite{1998lavalle_rrt} and TrajOpt \cite{2013schulman_trajopt}. By approximating the DLO as a minimal series of rigid links connected by spherical joints, the planner achieves swift path computation without relying on specialized planners or extensive datasets.
\paragraph{Local Controller}
Also implemented in Python, the local controller enforces safety and accuracy in real time using Quadratic Programming (QP) combined with CBFs to ensure obstacle avoidance, maintain DLO integrity and prevent overstress. To compensate for planning inaccuracies, it relies on a PBD simulation that approximates material properties such as Young’s and torsion moduli. The simulation provides stress estimations and minimum distance calculations, allowing the controller to handle both positional and orientational adjustments—crucial for stiffer objects. If the system enters a deadlock, the local controller requests a replan from the global planner.

\begin{figure}[!hp]
    \centering
    \includegraphics[width=0.80\columnwidth]{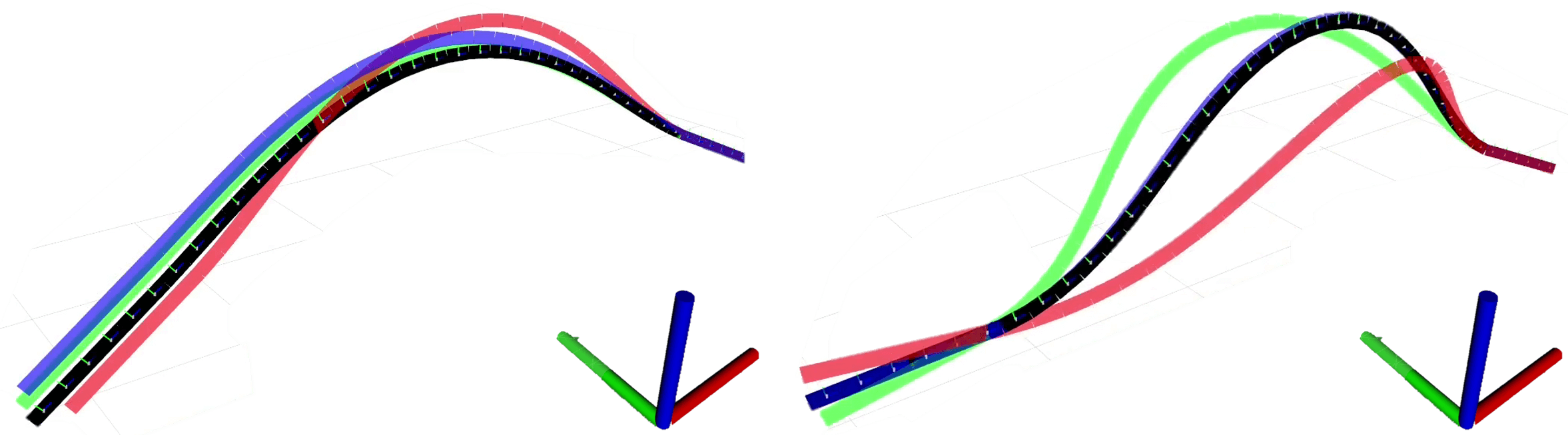}
    \caption{\textbf{Auxiliary DLO states} used for Jacobian computations at a single holding point (left end) via six parallel simulations. 
    Left: linear perturbations ($\Delta d$) along local X (red), Y (green), and Z (blue) axes. 
    Right: rotational perturbations ($\Delta \theta$) about local X (red), Y (green), and Z (blue).}
    \label{fig:aux-dlo-example}
    \vspace{-0.4cm}
\end{figure}

\paragraph{PBD-Based Simulations}
PBD simulations estimate the DLO state and predict its response to manipulator motions. For each manipulated holding point, $6N$ auxiliary simulations are run in parallel for $N$ agents. Each auxiliary simulation applies a small linear or angular perturbation to the holding point along or around the principal axes, as depicted in Fig.~\ref{fig:aux-dlo-example}. These perturbed states enable finite-difference approximations of key Jacobians that relate robot motions to tip poses, obstacle distances, and stress values—essential for the local controller’s safety and accuracy constraints. 
To maintain real-time performance, minimum distance computations and stress/force-torque estimations are embedded in the PBD simulation loop, implemented in C++ for efficiency.\footnote{While the controller can rely solely on simulation-based estimates, it may also incorporate external sensor data (e.g., F/T sensors) for enhanced accuracy and robustness.}

By pairing off-the-shelf global planning for quick path generation with a sophisticated local controller and embedded PBD simulations for state estimation and stress feedback, our framework achieves robust, efficient, and scalable DLO manipulation.

\section{PBD Based Modeling of DLOs}
\label{sec:modeling}

\subsection{DLO model}
\begin{figure}[!tp]
    \centering
    \includegraphics[width=0.6\columnwidth]{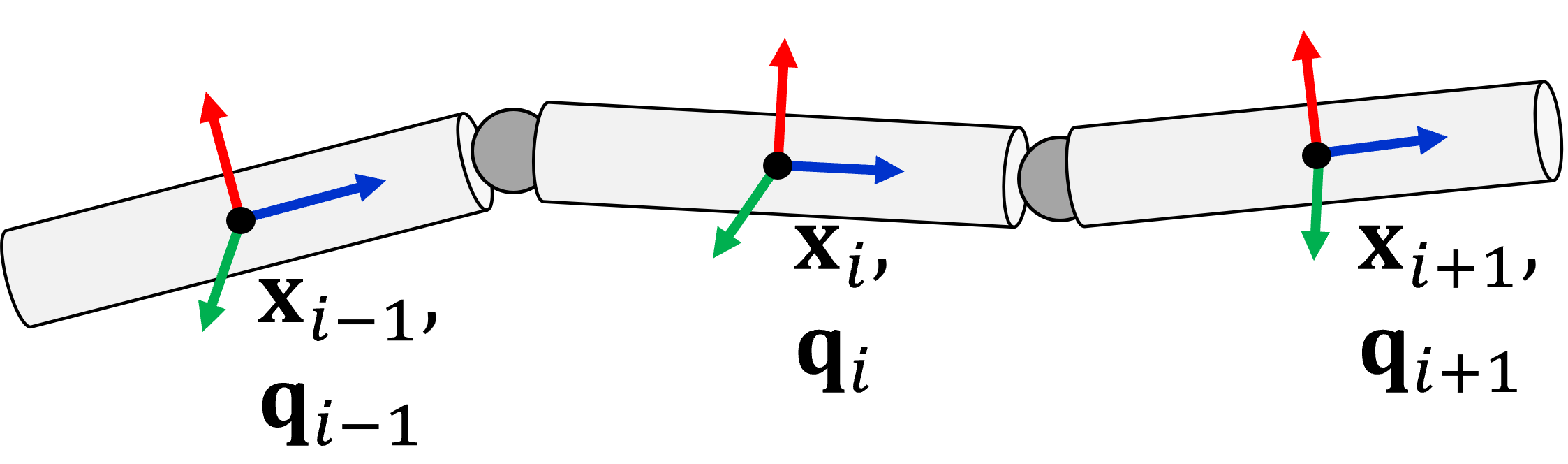}
    \caption{\textbf{DLO modeled} as $N$ discretized segments, each with positions, orientations, and stiffness embedded at segment connections and relative poses.} \label{fig:dlo-modeling}
    \vspace{-0.3cm}
\end{figure}

In PBD, deformable objects are modeled by discretizing them into $N$ particles. To represent a DLO of length $L$, we follow \cite{2018deul} by dividing it into $N$ line segments of equal length $l_i = L/N$. The state of DLO is then defined as $\mathbf{\Gamma} = \{\mathbf{x}, \mathbf{q} \} \in SE(3)^{N}$ where \( \mathbf{x} = (\mathbf{x}_1, \cdots, \mathbf{x}_N) \in \mathbb{R}^{3N} \) represents segment center positions (feature points), and \(\mathbf{q} = (\mathbf{q}_1, \cdots, \mathbf{q}_N) \in SO(3)^{N} \) represents the orientations of each segment (Fig.~\ref{fig:dlo-modeling}). Thus, the holding point of agent \(j\) corresponds to particle \(i\) is \(\mathbf{p}_{H_j} = (\mathbf{x}_i, \mathbf{q}_i) \).

\subsection{PBD based implementation}
Our PBD implementation (Algorithm~\ref{alg:pbd}), extends \cite{2024aksoy} by incorporating contact handling, real-time stress reporting, and simultaneous minimum distance checks to obstacles. The DLO is simulated with rigid body segments supporting both translational and rotational dynamics. We adopt \emph{small substeps} updates from \cite{2019macklin} and \emph{zero-stretch bending twisting} constraint from \cite{2018deul}, derived from Cosserat rod theory. This allows to simulate a range of material properties by adjusting stiffness parameters (e.g., Young's/shear moduli, and zero-stretch stiffness). Although these parameters do not necessarily match exact physical values, they provide visual fidelity and accurate force feedback. The DLO is treated as quasi-static and include damping to rapidly mitigate oscillations. 

\begin{algorithm}[!bp]
\scriptsize
\DontPrintSemicolon
\While{\textit{simulating}}{
    $h \gets \Delta t / \text{numSubsteps}$\;

    MinDistances \(\gets \) calculateMinDistances(DLO, Scene)\;
    
    \For{\textit{numSubsteps}}{
        \(\mathbf{\lambda} \gets \mathbf{0} \)\;
        
        \For{\textit{each body/particle}}{
            $\mathbf{x}_{\text{prev}} \gets \mathbf{x}$\;
            
            $\mathbf{v} \gets \mathbf{v} + h \mathbf{f}_{\text{ext}}/m$\;
            
            $\mathbf{x} \gets \mathbf{x} + h \mathbf{v}$\;
            
            $\mathbf{q}_{\text{prev}} \gets \mathbf{q}$\;
            
            $\mathbf{\omega} \gets \mathbf{\omega} + h \mathbf{I}^{-1} (\tau_{\text{ext}} - (\mathbf{\omega} \times (\mathbf{I}\mathbf{\omega})))$\;

            $\mathbf{q} \gets \mathbf{q} + h \frac{1}{2} [\omega_x,\omega_y,\omega_z,0]\mathbf{q}$\;

            $\mathbf{q} \gets \mathbf{q}/ | \mathbf{q} |$\;
        }
            
        \For{\textit{numSteps} (usually $= 1$)}{
            \((\mathbf{F},\mathbf{T}) \gets \mathbf{0} \)\;
            
            Contacts \(\gets \) detectContacts(DLO, Scene)\;

            \(\mathbf{x}, \mathbf{q} \gets \) solveContactPositionConstraints(DLO, Contacts, $h$)\;

            \(\mathbf{x}, \mathbf{q}, \mathbf{\Delta\lambda} \gets \) solveStretchBendTwistConstraints(DLO,$h,\mathbf{\lambda}$)\;

            \(\mathbf{\lambda} \gets \mathbf{\lambda} + \mathbf{\Delta\lambda} \) \;
            
            \((\mathbf{F},\mathbf{T}) \gets (\mathbf{F},\mathbf{T}) + \) CalculateDeltaFT($h,\mathbf{\lambda}$)
        }
        
        \For{\textit{each body/particle}}{
            $\mathbf{v} \gets (\mathbf{x} - \mathbf{x}_{\text{prev}}) /h$\;

            $ \Delta \mathbf{q} \gets \mathbf{q}\mathbf{q}^{-1}_{\text{prev}}  $\;

            $ \mathbf{\omega} \gets 2 [\Delta \mathbf{q}_x,\Delta \mathbf{q}_y,\Delta \mathbf{q}_z]/h$\;

            $\mathbf{\omega} \gets \Delta\mathbf{q}_w \geq 0\ ?\ \mathbf{\omega}: -\mathbf{\omega}$\;

            $(\mathbf{v}, \mathbf{\omega}) \gets $ addDamping($\mathbf{v}, \mathbf{\omega}$)\;
        }

        $(\mathbf{v}, \mathbf{\omega}) \gets $ solveContactVelocityConstraints(DLO, Contacts, $h$)\;
        
    }
}
\caption{PBD-based DLO Simulations}
\label{alg:pbd}
\end{algorithm}

\subsubsection{Reading Force-Torque values}
We estimate F/T at each holding point and internal segment by accessing the Lagrange multipliers ($\lambda$) computed at each solver iteration \cite{2016macklin}. This enables overstress avoidance using physical quantities rather than geometric constraints, crucial when manipulating stiff objects. For example, significant torques induced by gripper orientations when manipulating a tent pole cannot be detected without considering the object's shape. The simulation method thus incorporates holding point orientations into safety constraints.

\subsubsection{Minimum distance reporting \& Contact handling}
Using primitive shapes or custom meshes along Signed Distance Fields from the Discregrid library \cite{discregrid}, we efficiently calculate closest points and distances between the DLO and obstacles. This addresses the performance bottleneck noted in \cite{2024aksoy} and enables contact handling in simulation. By detecting environmental contact, the system can identify hooked scenarios for replanning and account for friction effects in dynamic simulations—guiding future extensions of our method.

\subsection{Parameter Identification}
\label{subsec:param_identification}
For real-world deployment of our DLO model, we focus on identifying only the Young’s and shear moduli. Our straightforward offline procedure computes parameter gradients in the PBD simulation using finite differences and minimizes the discrepancy between simulated and observed DLO behavior.
Practically, the DLO is held by two robots, one end remaining fixed while the other follows a simple rotation–translation trajectory designed to highlight the object’s stiffness properties. During this motion, the DLO is recorded via an RGB-D camera. Although more sophisticated techniques  (Section~\ref{subsec:dlo_param_identfication}) may yield higher global accuracy, our method is suffices for many tasks. When minimal twisting is involved, estimating Young’s modulus from a few representative configurations and assigning shear modulus via an isotropic relation (e.g., $E \approx 2G(1+\nu)$) can be adequate—particularly if conservative safety margins are adopted.



\section{Global Planner Pipeline}
\label{sec:global_planner}
\begin{figure}[!bp]
    \centering
    \includegraphics[width=1.00\columnwidth]{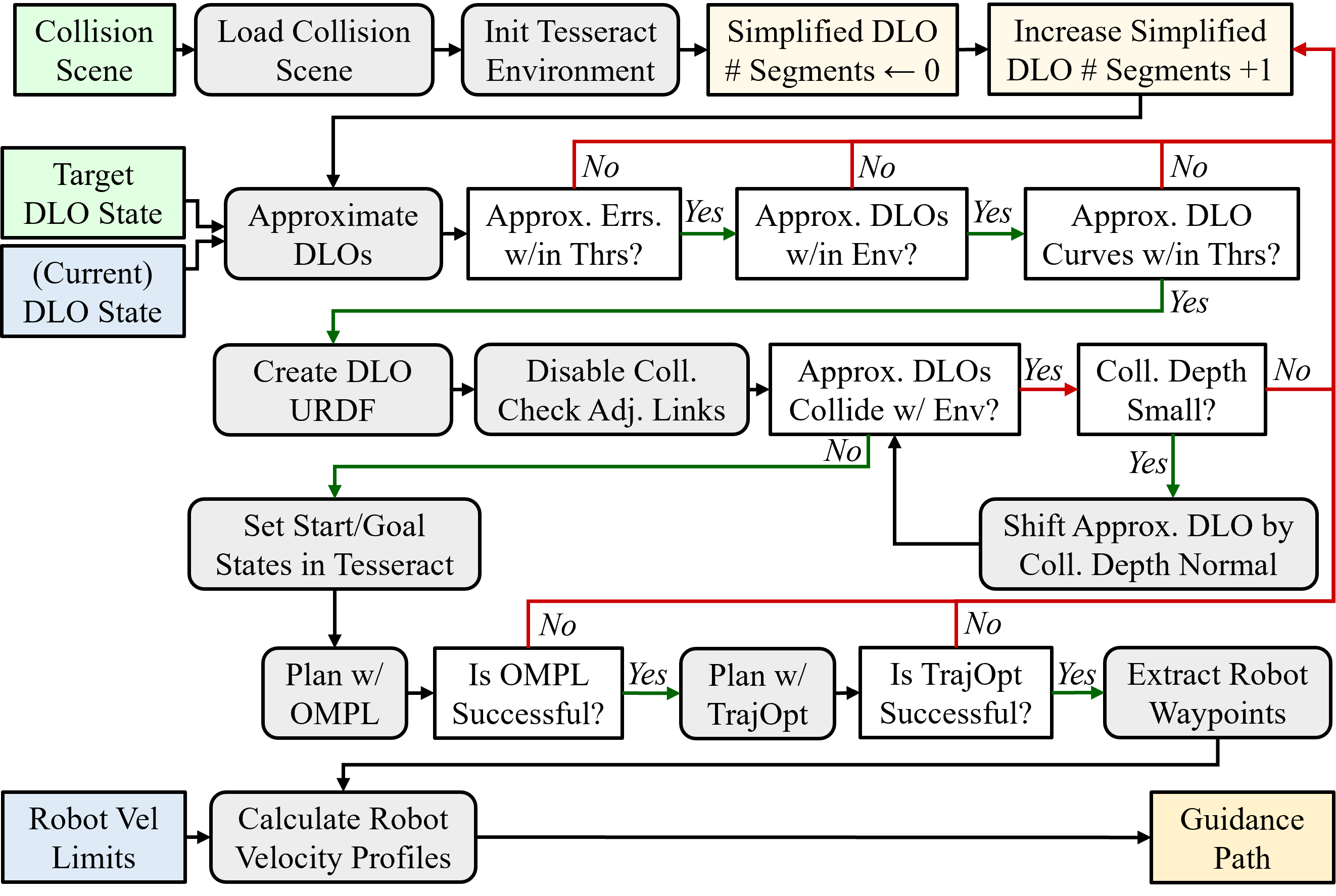}
    \caption{The \textbf{Global Planner Pipeline} (Fig.\ref{fig:method-overview}) rapidly computes a guidance trajectory for robots moving from current to target DLO states in a 3D collision scene.} \label{fig:global-planner}
    \vspace{-0.3cm}
\end{figure}

\looseness=-1
This section details our global planner pipeline, illustrated in Fig.\ref{fig:global-planner} and introduced in Section~\ref{sec:method_overview}. 

\subsection{DLO Approximation} 
\label{subsec:dlo-approximation}
After loading the collision scene from an URDF file into the Tesseract environment, the planner computes a guidance path by treating the DLO as a high-DoF articulated chain composed of rigid line segments connected by spherical joints. Although using more segments yields a closer DLO approximation, it also increases path planning complexity and computation time. To achieve efficient performance, we use fewer segments during planning, relying on the local controller to handle discrepancies between this simplified model and the full-resolution DLO state. 
For example, when transporting a DLO between relatively straight configurations, using only a few segments may suffice to produce a feasible coarse path. The local controller then compensates for the simplifications, ensuring accuracy and robustness. If needed, the approximation can be improved by adding more segments and adjusting curvature and angle limits—either manually chosen or mapped from the DLO’s stiffness parameters.

Given an original DLO representation $\mathbf{\Gamma}$ with $N$ segments, we approximate it as $\mathbf{\Gamma}_\text{apr}$ with $N_\text{apr} < N$ segments of equal length, preserving the total length $L$. To approximate segment $i_\text{apr}$, we choose a subset of the original indices:
\begin{equation}
\begin{aligned} 
i_{\text{start}} = \text{round}\left(i_{\text{apr}}  \tfrac{N}{N_{\text{apr}}}\right),
\ i_{\text{end}} = \text{round}\left((i_{\text{apr}}+1)  \tfrac{N}{N_{\text{apr}}}\right). 
\end{aligned}
\end{equation}

We set $\mathbf{x}_{\text{apr},i}$ as the average of $\mathbf{x}_{i_\text{start}:i_\text{end}}$, and compute $\mathbf{q}_{\text{apr},i}$ by averaging the corresponding quaternions using Markley’s method \cite{2007markley}. Assuming local alignment along the DLO’s $z$-axis, we derive each segment's start and end points from its known segment length $l_{i_\text{apr}} = L/N_\text{apr}$.

\begin{figure}[!tp]
    \centering
    \begin{subfigure}[t]{0.8137\columnwidth}
        \centering
        \includegraphics[width=\textwidth]{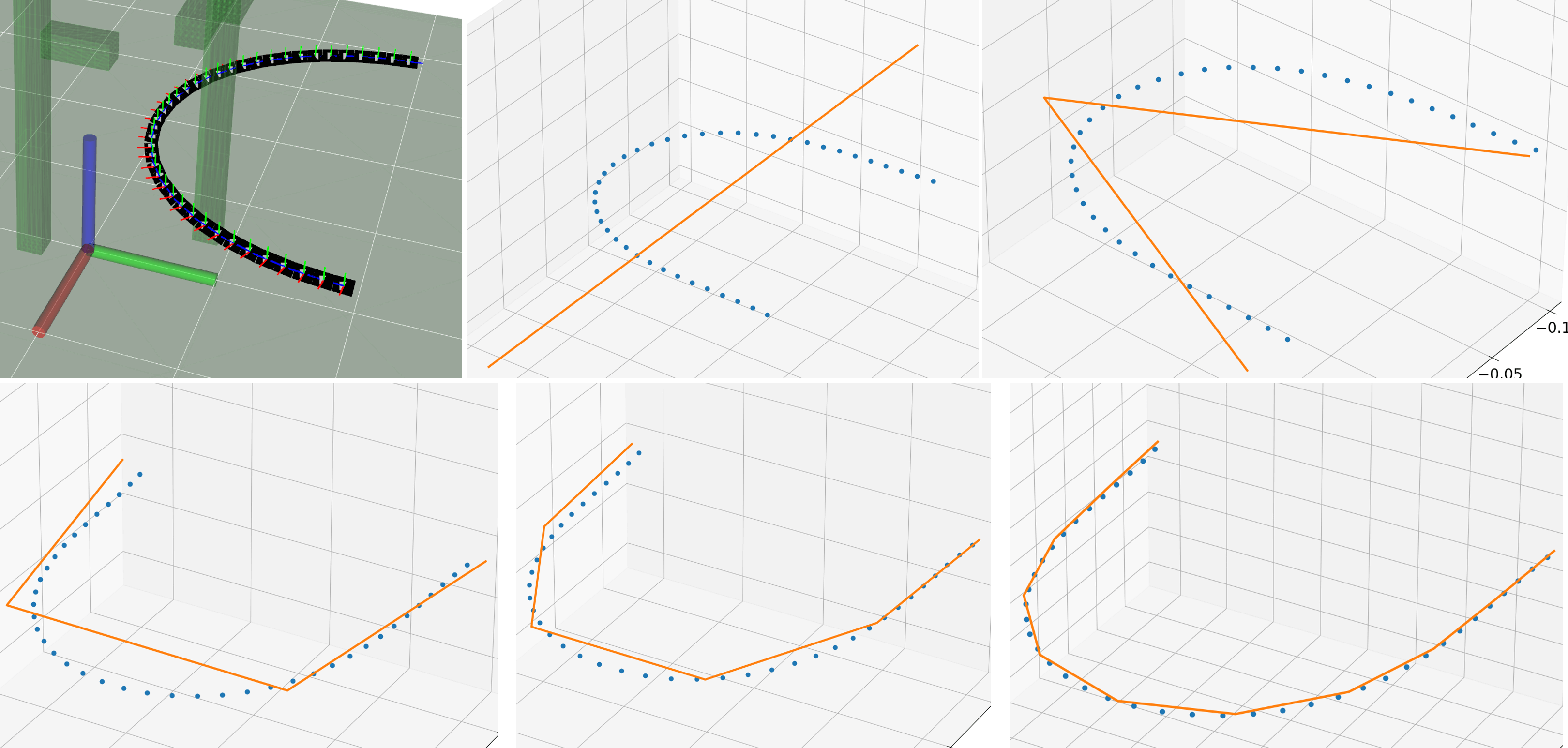}
        \caption{} \label{fig:dlo-approximation}
    \end{subfigure} \hfill
    \begin{subfigure}[t]{0.1737\columnwidth}
        \centering
        \includegraphics[width=\textwidth]{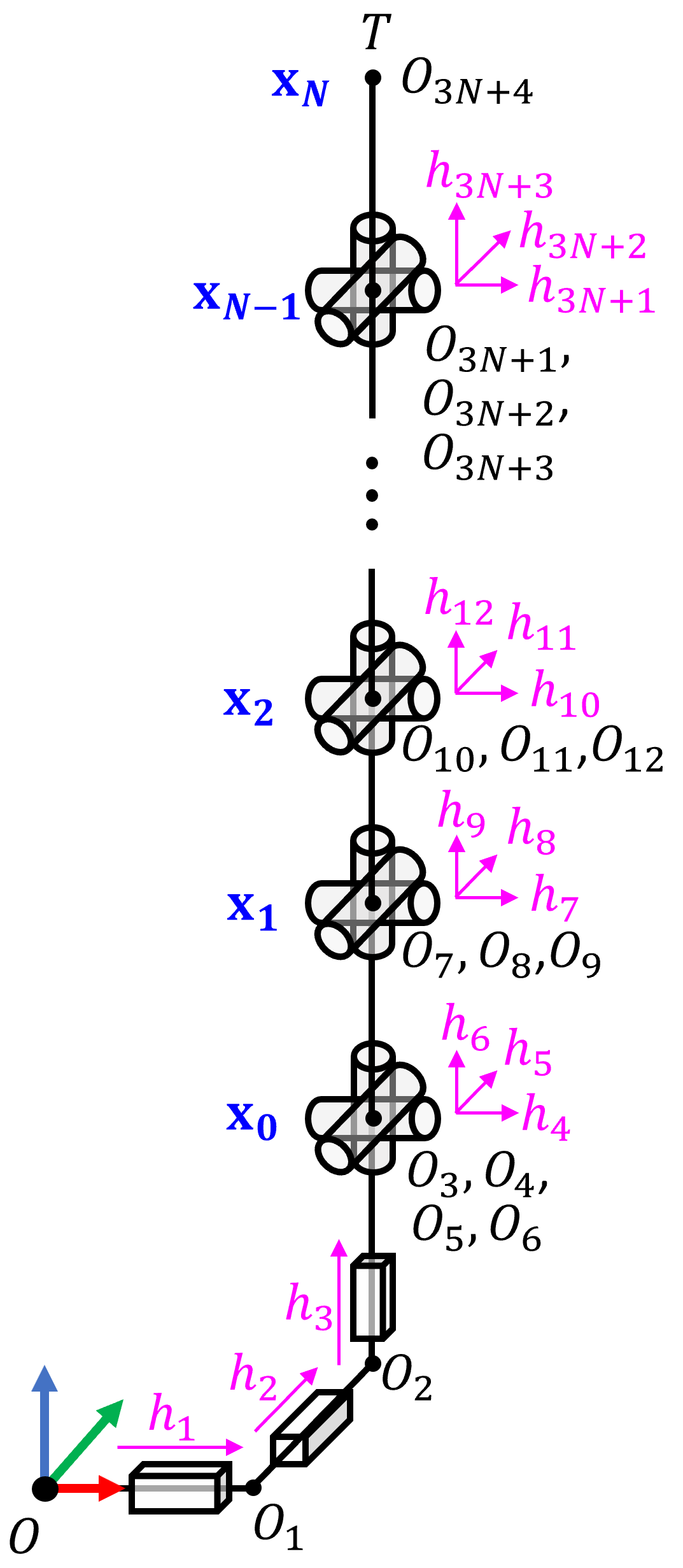}
        \caption{} \label{fig:dlo-zero-config}
    \end{subfigure}
    
    \caption{\textbf{(a)} Example \textbf{DLO approximations} for the global planner. The original 40-segments DLO (top left) is approximated with \(N=1,2,3,5,10\) segments (blue dots: original segment centers; orange lines: approximated DLOs). Approximations maintain equal segment lengths, preserve total length, and capture essential position and orientation characteristics. \textbf{(b)} Zero-configuration of the approximated \textbf{DLO kinematic model} as a chain of spherical joints.} 
    \vspace{-0.3cm}
\end{figure}

As $N_\text{apr}$ increases, the approximated DLO more closely matches the original state, reducing curvature errors and improving accuracy. Unlike naive interpolation methods---which can shorten the DLO or distort its orientation distribution\footnote{For example in Fig.\ref{fig:dlo-approximation}, a 1-link approximation with a naive interpolation would result in a line segment from the first point to the last point which neither maintains the total length nor preserves the position and orientation distribution of the DLO.}, our method preserves the DLO length and maintain the geometric and orientation characteristics of the original DLO. 
Although one may formulate the approximation as a nonlinear optimization, our approach is computationally efficient with complexity $\mathcal{O}(N_\text{apr})$, relying on averaging rather than iterative solvers. 

Fig.\ref{fig:dlo-approximation} shows various approximation levels. As $N_\text{apr}$ grows, the approximated representation converges to the original DLO geometry, striking a suitable balance between computational efficiency and fidelity.

\subsection{DLO URDF Creation and Kinematics}
\label{subsec:urdf-creation-n-kinematics}
Most path planning frameworks (e.g. Tesseract, MoveIt) require robot descriptions in URDF/Xacro formats—usually static for manufactured robots.
By contrast, our framework dynamically generates these descriptions for the approximated DLO, based on the selected number of segments. A Python script creates the URDF, modeling the DLO as cylindrical links connected by spherical joints, with one end attached to the world frame using virtual prismatic joints (Fig.~\ref{fig:dlo-zero-config})
\footnote{Instead of treating the DLO as a floating body, one could position it with robot bodies including their collision objects. We omit these details for simplicity.}. Adjacent DLO links in the URDF are excluded from self-collision checks.

Planners such as bidirectional RRT require collision-free initial and target states before sampling feasible paths. Determining collisions at these states requires the corresponding joint angles of the approximated DLO. We treat this as an inverse kinematics (IK) problem: given the simplified state $\mathbf{\Gamma}_\text{apr}$, solve for the joint vector 
$\mathbf{\Theta}= [\theta_x,\theta_y,\theta_z,\theta_{x_1},\theta_{y_1},\theta_{z_1},\cdots,\theta_{x_{N_{\text{apr}}}},\theta_{y_{N_{\text{apr}}}},\theta_{z_{N_{\text{apr}}}}] \in \mathbb{R}^{3N_{\text{apr}}+3}$. 

The first three joint values correspond to the prismatic base positioning (i.e. $\mathbf{x}_0$), and the subsequent triples $\theta_{x_{\text{i}}},\theta_{y_{\text{i}}},\theta_{z_{\text{i}}}$ define spherical joints between consecutive links. 
For each link $i=1,\cdots,N_{\text{apr}}$,
\begin{equation}
    \begin{aligned}
        \mathbf{R}_{i} &=\mathbf{R}_{i-1} \mathbf{R}_{XYZ}(\theta_{x_{\text{i}}},\theta_{y_{\text{i}}},\theta_{z_{\text{i}}}), \\
        \mathbf{x}_{i} &= \mathbf{x}_{i-1} + \mathbf{R}_{i} l_{i_{\text{apr}}}\mathbf{e}_z,
    \end{aligned}
    \label{eqn:dlo_inv_kin}
\end{equation}
where $\mathbf{x}_{i}$ and $\mathbf{R}_{i}$ come from the DLO approximation, $\mathbf{R}_{XYZ}(\theta_{x},\theta_{y},\theta_{z}) = \mathbf{R}_x(\theta_x)\mathbf{R}_y(\theta_y)\mathbf{R}_z(\theta_z)$ is the spherical joint rotation, and $\mathbf{e}_z = [0,0,1]^\top$ is the local axis of the DLO. 
We solve for these angles using the Paden–Kahan subproblems \cite{2024elias}. First, $\theta_{x_{\text{i}}}$ and 
$\theta_{y_{\text{i}}}$ are found by matching $\mathbf{R}_{XYZ}\mathbf{e}_z$ to the DLO's local $z$-axis (\textit{Subproblem 2}).
Then, given those angles, $\theta_{z_{\text{i}}}$ is computed by comparing $\mathbf{R}_{XYZ}\mathbf{e}_x$ with a rotated $\mathbf{e}_x$ (\textit{Subproblem 1}). 

\subsection{Setting Planner Start \& Goal States}
By dynamically generating the DLO URDF and solving the IK problem, we can quickly verify whether the approximated DLO’s initial and target configurations are collision-free. If both states pass collision checks, the planner proceeds. Otherwise, if collision depths exceeds a threshold, we assume a more accurate approximation is needed; incrementing the number of segments and repeating collision checks.
For small collision depths, a local pose shift can often resolve the issue---using collision data from Tesseract (points and normals) to translate the approximated DLO accordingly. If shifting fails, we again refine the approximation. Once collision-free states are confirmed, the planner proceeds to generate the guidance path. 
Minor adjustments in the approximated states are acceptable because the local controller compensates for small mismatches during execution.

\subsection{Path Planning \& Robot Trajectory Generation}
\label{subsec:robot-traj-gen-from-path-planning}
With collision-free start and goal states established, we initiate planning using a Tesseract-integrated OMPL planner, specifically RRT-Connect~\cite{2000kuffner-RRT-Connect}. As a probabilistic method,  RRT-Connect does not guarantee a solution, so we retry if the initial attempt fails.
After finding a feasible path, we smooth it with TrajOpt~\cite{2013schulman_trajopt}, optimizing path length, velocity, acceleration limits, and any custom cost functions. A final feasibility check confirms the smoothed plan remains collision-free. 
We then compute the 3D waypoints for each robot by applying forward kinematics on the approximated DLO's planned joint angles.
To generate synchronized robot trajectories, we assign velocity profiles respecting each robot’s speed limits. For each path segment, we calculate the time required for both robots and set the segment duration to the maximum of these times. Since the final plan serves only as a coarse guide for the local controller, we do not implement advanced velocity profiling (e.g., trapezoidal). The local controller further refines the motion during execution.


\section{Local Controller Design}
\label{sec:controller_design}
\begin{figure}[!tp]
    \centering
    \includegraphics[width=1.00\columnwidth]{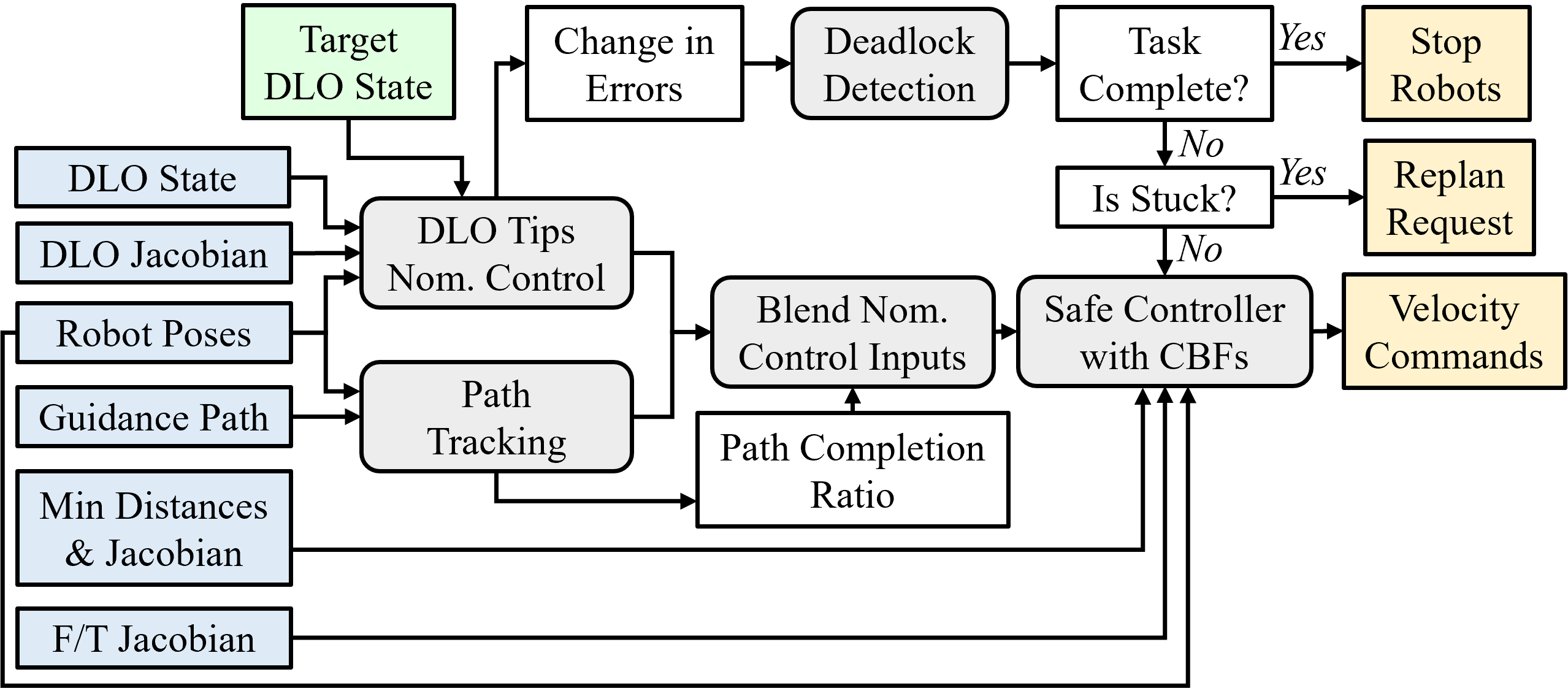}
    \caption{The \textbf{Local Controller} (Fig.\ref{fig:method-overview}) uses control feedback and DLO Jacobians to compute safe real-time robot inputs. It blends path tracking with tip control, detects stuck states, and triggers replanning if needed.} \label{fig:local-controller}
    \vspace{-0.3cm}
\end{figure}

This section details the local controller, shown in Fig.\ref{fig:local-controller} and introduced in Section~\ref{sec:method_overview}.
\subsection{Nominal Controllers}
\looseness=-1
\subsubsection{Tip Pose Controller}
We begin with a nominal controller to guide the DLO tips toward their desired poses, minimizing tip pose errors:
\begin{equation}
    \begin{aligned}
        \mathbf{u}^{\text{tip}}_{\text{nom}} &= - \mathbf{K}^{\text{tip}}_p \mathbf{J}_{\text{tip}}^{\dagger} \mathbf{e}_{\text{tip}} \in \mathbb{R}^{12},
    \end{aligned}    
    \label{eqn:nom_controller}
\end{equation}
where 
\(\mathbf{e}_{\text{tip}} = [\mathbf{e}_{T_1}^\top \ \mathbf{e}_{T_2}^\top]^\top \) is the concatenated tip pose error, 
\( \mathbf{K}^{\text{tip}}_p\) is a control gain, and
\( \mathbf{J}_{\text{tip}}\) is the Jacobian relating tip velocities to control inputs:
\begin{equation}
    \begin{aligned}
        \dot{\mathbf{p}}_{\text{tip}} &= \begin{bmatrix}
 \dot{\mathbf{p}}_{T_1}\\
 \dot{\mathbf{p}}_{T_2}
\end{bmatrix} = \mathbf{J}_{\text{tip}} \mathbf{u} = \begin{bmatrix}
 \mathbf{J}_{T_1H_1} &\mathbf{J}_{T_1H_2} \\
 \mathbf{J}_{T_2H_1} &\mathbf{J}_{T_2H_2} 
\end{bmatrix} \begin{bmatrix}
 \mathbf{u}_{1}\\
 \mathbf{u}_{2}
\end{bmatrix}.
    \end{aligned}    
    \label{eqn:tip_jacobian}
\end{equation}
Here \(\mathbf{J}_{T_iH_j} \in \mathbb{R}^{6\times6}\) is approximated via finite differences in simultaneous PBD simulations \cite{2024aksoy} and includes both orientation and position effects.\footnote{Note that, if \(\mathbf{p}_{H_1} \) is in between \({\mathbf{p}}_{T_1} \) and \(\mathbf{p}_{H_2}\) along the DLO, then \( \mathbf{J}_{T_1H_2} = \mathbf{J}_{T_2H_1} = \mathbf{0}\) in (\ref{eqn:tip_jacobian}). Otherwise, \( \mathbf{J}_{T_1H_1} = \mathbf{J}_{T_2H_2} = \mathbf{0}\).}
\footnote{This formulation targets only tip poses for a fast controller. For a general DLO shape control with \(N\) segments, (\ref{eqn:nom_controller}) and (\ref{eqn:tip_jacobian}) extend to 
\( \mathbf{u}^{\text{dlo}}_{\text{nom}} = - \mathbf{K}_p \mathbf{J}_{\text{dlo}}^{\dagger} \mathbf{e}_{\text{dlo}} \) and 
\( \dot{\mathbf{p}}_{\text{dlo}} = \mathbf{J}_{\text{dlo}} \mathbf{u} \) respectively, where 
\(\mathbf{e}_{\text{dlo}} \in \mathbb{R}^{6N}\) and 
\( \mathbf{J}_{\text{dlo}} \in \mathbb{R}^{6N\times 12} \).}.
\subsubsection{Path Tracking}
\label{subsubsec:path-tracking}
Because the globally planned path is coarse, our controller allows deviations for safety and model inaccuracies. The path thus serves as guidance rather than a strict trajectory. We define:
\begin{equation}
    \begin{aligned}
        \mathbf{u}^{\text{path}}_{\text{nom}} &= [\mathbf{u}^{\text{path}\top}_{\text{nom,}1} \quad \mathbf{u}^{\text{path}\top}_{\text{nom,}2}]^\top \in \mathbb{R}^{12},
    \end{aligned}
    \label{eqn:nom_controller_path_tracking}
\end{equation}
where \(\mathbf{u}^{\text{path}}_{\text{nom,}j}\) is the nominal path tracking input for each robot \(j=1,2\), 
defined as:
\begin{equation}
    \begin{aligned}
        \mathbf{u}^{\text{path}}_{\text{nom,}j} &:= [\mathbf{u}^{\text{path}\top}_{\text{nom,}\mathbf{x}_j} \quad                                                                                     \mathbf{u}^{\text{path}\top}_{\text{nom,}\mathbf{R}_j}]^\top \in \mathbb{R}^{6},
    \end{aligned}
    \label{eqn:nom_controller_path_tracking2}
\end{equation}
where \(\mathbf{u}^{\text{path}}_{\text{nom,}\mathbf{\phi}} \in \mathbb{R}^{3}\) designates linear or angular components with \(\mathbf{\phi} \in \{ \mathbf{x}_j,\mathbf{R}_j \}\) as:
\begin{equation}
    \begin{aligned}
        \mathbf{u}^{\text{path}}_{\text{nom,}\mathbf{\phi}} &:=\begin{cases}
                                                     \mathbf{u}^{\text{path}}_{p,\phi\perp} + \mathbf{u}^{\text{path}}_{\text{ff},\phi} , &\text{if}\ ({\mathbf{u}^{\text{path}}_{p,\phi}}^{\top}\mathbf{u}^{\text{path}}_{\text{ff},\phi})>0, \\
                                                     \mathbf{u}^{\text{path}}_{p,\phi}, &\text{otherwise}.
                                                   \end{cases}
    \end{aligned}
    \label{eqn:nom_controller_path_tracking3}
\end{equation}
Here \(\mathbf{u}^{\text{path}}_{p,\phi}\) is the proportional term calculated with gains \(\mathbf{K}_{p,\mathbf{x}_j}\) and \(\mathbf{K}_{p,\mathbf{R}_j}\) as
\begin{equation}
    \begin{aligned}
        \mathbf{u}^{\text{path}}_{p,\mathbf{x}_j} &= -\mathbf{K}_{p,\mathbf{x}_j} \ \mathbf{e}_{\mathbf{x}_{H_j}} ,\\
        \mathbf{u}^{\text{path}}_{p,\mathbf{R}_j} &= -\mathbf{K}_{p,\mathbf{R}_j} \ \mathbf{e}_{\mathbf{R}_{H_j}} ,
    \end{aligned}
    \label{eqn:nom_controller_path_tracking4}
\end{equation}
where \(\mathbf{e}_{\mathbf{x}_{H_j}},\mathbf{e}_{\mathbf{R}_{H_j}}\)are position and orientation errors between the current waypoint \(\mathbf{p}^{\text{path}}_{j,k}\) and \(\mathbf{p}_{H_j}\). 
The feed-forward term \( \mathbf{u}^{\text{path}}_{\text{ff},\phi}\) is a velocity vector aligned with the path direction to synchronize robot motions according to speed limits.
The perpendicular part \( \mathbf{u}^{\text{path}}_{p,\phi\perp}\) is: 
\begin{equation}
    \begin{aligned}
        \mathbf{u}^{\text{path}}_{p,\phi\perp} &= \mathbf{u}^{\text{path}}_{p,\phi} - \frac{ \mathbf{u}^{\text{path}}_{\text{ff},\phi}} {\|  \mathbf{u}^{\text{path}}_{\text{ff},\phi} \|} ({\mathbf{u}^{\text{path}}_{p,\phi}}^{\top}\mathbf{u}^{\text{path}}_{\text{ff},\phi}).
    \end{aligned}
    \label{eqn:nom_controller_path_tracking5}
\end{equation}

We increment the waypoint index \(k\) only when both robots lie within a threshold distance or have passed their respective target waypoints along the path direction. 
This conditional approach uses the feed-forward term to reduce longitudinal errors while the perpendicular part of the proportional control handles lateral adjustments. If one robot reaches or surpasses the waypoint ahead of the other, the feed-forward term is disabled and the controller relies on proportional control to reduce the remaining error.

\subsubsection{Blending Nominal Control Inputs}
\label{subsubsec:blending-nom-control}
When a guidance path is available, we prioritize path tracking over tip control until later in the motion. We blend both nominal controllers:
\begin{equation}
    \begin{aligned}
        \mathbf{u}_{\text{nom}} &= (1-w)\mathbf{u}^{\text{tip}}_{\text{nom}} + w\mathbf{u}^{\text{path}}_{\text{nom}}
    \end{aligned}    
    \label{eqn:nom_controller_blended}
\end{equation}
where \(w = \beta(\overline{l_{\text{path}}}, l_{\text{up}}, l_{\text{low}}) \in[0,1]\) is a weight function:
\begin{equation}
\beta(l, l_{\text{up}}, l_{\text{low}})  = 
\begin{cases}
	0,  &  l \leq l_{\text{low}}, \\
	0.5  [ 1 -\cos( \frac{  l - l_{\text{low}} } {l_{\text{up}}-l_{\text{low}}}\pi) ], & l_{\text{low}} < l < l_{\text{up}}, \\
	1, & l_{\text{up}} \leq l.
\end{cases}
\label{eqn:transition}
\end{equation}
Here, \(\overline{l_{\text{path}}}\) is the average remaining path length for the robots, while \(l_{\text{up}}\) and \(l_{\text{low}}\) are user-specified thresholds. The controller thus transitions smoothly from path tracking (\(w=1\)) to tip control (\(w=0\)).

\subsection{Safe Controller with CBFs}
\label{sec:controller_design:cbf}
The nominal controller \(\mathbf{u}_{\text{nom}}\) handles path tracking and tip control but does not address safety considerations (e.g., collision avoidance, overstress prevention). To ensure safe operation, we incorporate CBFs, known for their robustness and flexibility~\cite{2022herguedas}. CBFs preserve key performance objectives while enforcing crucial safety properties.

\subsubsection{Collision Avoidance}
Let $f(\mathbf{\Gamma})$ represent the minimum distance between the DLO and environment, assumed locally Lipschitz. We define the candidate CBF:
\begin{equation}
    h^\text{coll} (\mathbf{\Gamma}) = f(\mathbf{\Gamma}) - d^\text{ofs},
    \label{eqn:h_coll_avoid}
\end{equation}
where $d^\text{ofs}$ is a user-defined offset distance. 
Differentiating (\ref{eqn:h_coll_avoid}) with respect to time yields:
\begin{equation}
    \begin{aligned}
        \dot{h}^\text{coll}(\mathbf{\Gamma},\mathbf{u}) &= \frac{\partial h^{\text{coll}}}{\partial \mathbf{\Gamma}} \dot{\mathbf{\Gamma}}= \nabla f(\mathbf{\Gamma}) g(\mathbf{\Gamma},\mathbf{u}), 
    \end{aligned}
\end{equation} 
where $g$ defines the system dynamics as $\dot{\mathbf{\Gamma}} = g(\mathbf{\Gamma},\mathbf{u})$. 
For $h^\mathrm{coll}$ to be a valid CBF \cite{2019ames}, we require $\nabla f(\mathbf{\Gamma}) \neq \mathbf{0}$ when $h^\mathrm{coll} = 0$, ensuring that at least one DLO segment influences the minimum distance at $d_\mathrm{offset}$.
With this assumption and from the CBF definition, the collision avoidance imposes:
\begin{equation}
    \begin{aligned}
        \nabla f(\mathbf{\Gamma}) g(\mathbf{\Gamma},\mathbf{u})  &\geq -\alpha_{\text{coll}}(h^\text{coll}(\mathbf{\Gamma})),    
    \end{aligned}
\end{equation} 
where $\alpha_{\text{coll}}$ is an extended class-$\mathcal{K}$ function (defined later). 

Approximating \(\nabla f(\mathbf{\Gamma})\,g(\mathbf{\Gamma},\mathbf{u})\approx \mathbf{J}^\text{coll}(\mathbf{\Gamma})\,\mathbf{u}\),
we obtain
\begin{equation}
    [\mathbf{J}_{H_1}^\text{coll},\,\mathbf{J}_{H_2}^\text{coll}]
    \,\mathbf{u}
    \;\ge\;
    -\,\alpha_{\text{coll}}\bigl(h^\text{coll}(\mathbf{\Gamma})\bigr),
    \label{eqn:lin_constraint_coll_avoidance}
\end{equation}
where \(\mathbf{J}_{H_j}^\text{coll}\in\mathbb{R}^{1\times6}\) maps the motion of robot \(j\) to the distance function \(f(\mathbf{\Gamma})\), computed in real time with finite differences using auxiliary PBD simulations. This linear constraint enforces collision avoidance in a robust, real-time manner. For multiple obstacles, each minimum distance introduces an additional linear constraint.

\subsubsection{Overstress Avoidance}
While collision avoidance protects the DLO from external hazards, it can push the system into high-deformation configurations or exceed the robot stress limits. To prevent overstress, we define CBFs using F/T readings at the holding points:

\begin{equation}
    \mathbf{h}^\text{strs}(\mathbf{\Gamma})
    = (\mathbf{w}^\text{max}-\mathbf{w}^\text{ofs})^2
      \;-\;
      \bigl(\mathbf{w}(\mathbf{\Gamma})\odot\mathbf{w}(\mathbf{\Gamma})\bigr),
    \label{eqn:h_overstress_avoid_general}
\end{equation}
where \(\mathbf{w}(\mathbf{\Gamma})\in\mathbb{R}^{12}\) is the concatenated F/T vector at robot holding points, \(\mathbf{w}^\text{max}\in\mathbb{R}^{12}\) specifies stress/F/T limits, \(\mathbf{w}^\text{ofs}\in\mathbb{R}^{12}\) is an offset vector, and \(\odot\) denotes elementwise multiplication. Differentiating this w.r.t.\ time,
\begin{equation}
\dot{\mathbf{h}}^\text{strs}(\mathbf{\Gamma},\mathbf{u}) = \frac{\partial \mathbf{h}^\text{strs}}{\partial \mathbf{\Gamma}}\dot{\mathbf{\Gamma}} 
= -2\,\mathbf{w}(\mathbf{\Gamma}) \odot \bigl(\nabla \mathbf{w}(\mathbf{\Gamma}) g(\mathbf{\Gamma},\mathbf{u})\bigr).
\end{equation}
We introduce a Jacobian
\begin{equation}
\mathbf{J}^\text{strs}(\mathbf{\Gamma}) = 
\begin{bmatrix}
\mathbf{J}_{H_1H_1}^\text{strs} & \mathbf{J}_{H_1H_2}^\text{strs} \\
\mathbf{J}_{H_2H_1}^\text{strs} & \mathbf{J}_{H_2H_2}^\text{strs} 
\end{bmatrix},
\end{equation}
to approximate $\nabla \mathbf{w}(\mathbf{\Gamma}) g(\mathbf{\Gamma},\mathbf{u}) \approx \mathbf{J}^\text{strs}(\mathbf{\Gamma}) \mathbf{u}$. Each block $\mathbf{J}_{H_iH_j}^\text{strs} \in \mathbb{R}^{6\times6}$ maps the motion of robot $j$ to the F/T at the holding point of robot $i$, computed in real-time via finite differences in auxiliary PBD simulations.
Substituting this approximation, the overstress avoidance constraints reduce to
\begin{equation}
-2\,\mathbf{w}(\mathbf{\Gamma}) \odot (\mathbf{J}^\text{strs}(\mathbf{\Gamma})\mathbf{u}) \succcurlyeq -\alpha_{\text{strs}}\bigl(\mathbf{h}^\text{strs}(\mathbf{\Gamma})\bigr),
\label{eqn:lin_constraint_stress_avoidance}
\end{equation}
where $\succcurlyeq$ is componentwise inequality, and $\alpha_{\text{strs}}$ is an extended class-$\mathcal{K}$ function. This formulation maintains safe stress levels for both the DLO and the robots without unduly constraining the nominal controller.

\subsection{QP Based Controller}
We combine the derived constraints (\ref{eqn:lin_constraint_coll_avoidance}) and (\ref{eqn:lin_constraint_stress_avoidance}) into a quadratic program (QP) to find a safe control input $\mathbf{u}$:
\begin{equation}
    \begin{aligned}
    \min_{\mathbf{u}} \quad & \frac{1}{2} \left\| \boldsymbol{\gamma} (\mathbf{u} -  \mathbf{u}_{\text{nom}})  \right\|_2^2, \\
    \text{s.t.:} \quad & \mathbf{J}^\text{coll}(\mathbf{\Gamma})\mathbf{u} \succcurlyeq -\alpha_{\text{coll}}(h^\text{coll}(\mathbf{\Gamma})), \\
                \quad & -2\mathbf{w}(\mathbf{\Gamma}) \odot (\mathbf{J}^\text{strs}(\mathbf{\Gamma})\mathbf{u}) \succcurlyeq -\alpha_{\text{strs}}\bigl(\mathbf{h}^\text{strs}(\mathbf{\Gamma})\bigr), \\
                 \quad & -\|\mathbf{u} \|_{\infty} \succcurlyeq  -\mathbf{u}_{\text{max}}.
    \end{aligned}    
    \label{eqn:qp_controller}
\end{equation}
Here, $\boldsymbol{\gamma}$ scales the relative importance of different velocity components, and $\mathbf{u}_{\text{max}}$ represents robot velocity limits. 
The QP thus finds the feasible input \(\mathbf{u}\) closest to \(\mathbf{u}_{\text{nom}}\) while respecting collision and overstress constraints.

\subsection{Selection of \texorpdfstring{$\alpha$}{α} Functions}
For valid CBF construction \cite{2019ames}, we choose extended class-$\mathcal{K}$ functions \(\alpha_{\text{coll}}\) and \(\alpha_{\text{strs}}\) piecewise:
\begin{align}
        -\alpha_{\text{coll}}(h) &= \begin{cases} 
                                         -c_{1}^{\text{coll}}h, &  h < 0,\\
                                         \frac{-c_{2}^{\text{coll}} h}{(d^\text{frz}-d^\text{ofs}) - h}, &  0 \leq h \leq (d^\text{frz}-d^\text{ofs}),\\
                                         -\infty, &  (d^\text{frz}-d^\text{ofs}) < h,
                                    \end{cases} \label{eqn:dlo-alpha-func-coll}\\
        -\alpha_{\text{strs}}(\mathbf{h}) &= \begin{cases} 
                                                -\mathbf{c}_{1}^{\text{strs}} \odot \mathbf{h}, & \mathbf{h} \prec \mathbf{0},\\
                                                \frac{-\mathbf{c}_{2}^{\text{strs}} \odot \mathbf{h}}{(\mathbf{w}^\text{max}-\mathbf{w}^\text{ofs})^2 - \mathbf{h}},  & \mathbf{0} \preccurlyeq \mathbf{h} .\\
                                                \end{cases}\label{eqn:dlo-alpha-func-stress}
\end{align} 

Here, \(d^\text{frz}\) defines a free zone distance at which the collision constraint becomes inactive, while \(d^\text{ofs}\) sets an offset for earlier activation.
Similarly, \(\bigl(\mathbf{w}^\text{max}-\mathbf{w}^\text{ofs}\bigr)^2\) specifies a stress threshold, and $c_{1}^{\text{coll}}, c_{2}^{\text{coll}}, \mathbf{c}_{1}^{\text{strs}}, \mathbf{c}_{2}^{\text{strs}}$ are positive user-tuned constants. 
This piecewise design ensures smooth, progressive constraint enforcement as the system approaches defined collision or stress limits.

\section{Results}
\label{sec:results}
\looseness=-1

This section presents experimental results validating our framework in both simulation and real-world scenarios. We focus on three main tasks: (1) tent-pole placement for tent building, (2) maneuvering a tent pole through an L-shaped corridor, and (3) benchmarking in challenging collision scenes against state-of-the-art methods.

All experiments use the global planner and local controller running on a laptop with an Intel Core i9-10885H CPU. The PBD simulations run on a separate laptop with an Intel Core i7-13800H CPU, reflecting a realistic setup in which each robot manages its own object simulations and shares only the relevant Jacobians with the main controller. Detailed parameter settings are provided in each subsection, and additional configurations can be found in our released code (see Section~\ref{sec:introduction}). Supplementary video recordings of the tested cases\footnote{
\href{https://drive.google.com/file/d/12cikIMmTpxi-WdGvOUpgPp8YTDApeuch/view?usp=drive_link}{https://bit.ly/3DkPZKE}
} are also available. All visualizations are generated in RViz using our ROS-based implementation.

\subsection{Local Controller Evaluations in Simulation}
We first evaluate the local controller in isolation (i.e., without the global planner) using a simplified tent-building task, described below. This experiment demonstrates the importance of stress avoidance and underscores the need for a global planner to handle more complex scenarios. For safety, these evaluations are conducted in simulation only.

\subsubsection{Tent Building Task}
Tent building is chosen because it involves manipulating 1D (poles) and 2D (tarp) deformable objects. In practice, poles are assembled, attached to the tent body, inserted into corner grommets, and so on. Here, we focus on transporting a tent pole as a DLO for placement into grommets.

\begin{figure}[!hbtp]
  \centering
  \begin{minipage}[c]{0.50\columnwidth}
    \includegraphics[width=0.99\textwidth]{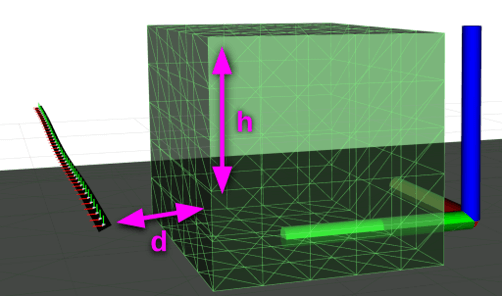}
  \end{minipage}
  \begin{minipage}[c]{0.45\columnwidth}
    \caption{DLO placement for tent-building experiments. The DLO is positioned at different distances $d$ behind the box and heights $h$ relative to its top, remaining straight and parallel to the ground.} 
    \label{fig:DLO-tent-building-d-h}
  \end{minipage}
  \vspace{-0.3cm}
\end{figure}

We simplify the task to two agents carrying a tent pole over a rectangular box obstacle and inserting its tips into grommet holes, as shown in Fig.~\ref{fig:problem-definition}. 
The obstacle is a $(1.0\times 1.0\times 1.5)$m box placed on the ground plane (XY-plane) centered at $(0,\,1,\,0)$m. 
The tent pole is $3.3528\,\mathrm{m}$ long, $7\,\mathrm{mm}$ thick, has a $1792.89 \,\mathrm{kg/m^3}$ density, $30\,\mathrm{GPa}$ and $10\,\mathrm{GPa}$ Young’s and shear moduli, respectively. 
It is placed at varying distances $d$ behind the box and heights $h$ relative to the box top (Fig.~\ref{fig:DLO-tent-building-d-h}), aligned straight and parallel to the ground. 
The grommets are located at $y=-1.0\,\mathrm{m}$ and $x=\pm 1.0\,\mathrm{m}$ with a desired orientation $5^\circ$ from the $+z$-axis. 
The DLO is discretized into 40 segments, with robots grasping segment indices 5 and 34 (approximately $46\,\mathrm{cm}$ from each tip). 
Gravity is $9.804\,\mathrm{m/s^2}$.

\vspace{-0.3cm}

\subsubsection{Case 1: DLO Tip Nominal Control Only} 
\begin{figure}[!tbp]
    \centering
    \begin{subfigure}[t]{0.4937\columnwidth}
        \centering
        \includegraphics[width=\textwidth]{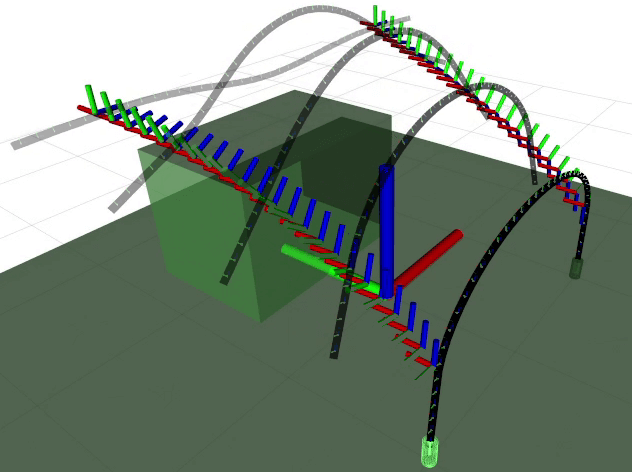}
        \caption{Starting $0.5\,\mathrm{m}$ above the box; snapshots at $t=\{0,3,7,13,53\}\,\mathrm{s}$.} 
    \end{subfigure} \hfill
    \begin{subfigure}[t]{0.4937\columnwidth}
        \centering
        \includegraphics[width=\textwidth]{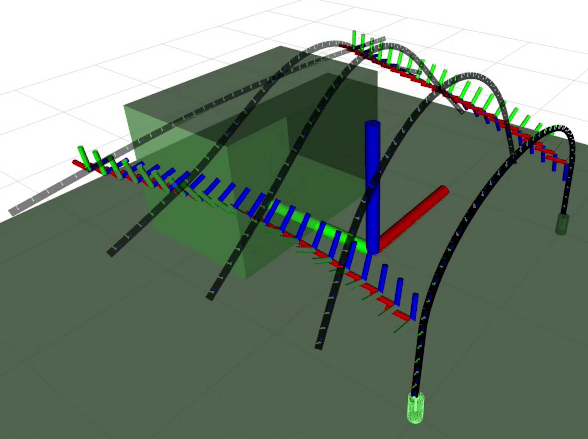}
        \caption{Starting on top of the box; snapshots at $t=\{0,2,5,11,41\}\,\mathrm{s}$.} 
    \end{subfigure}
    
    \caption{Verification of \textbf{DLO Tip Nominal Control} \eqref{eqn:nom_controller} in tent-building simulation. Safety features are disabled, so initial placements avoid collisions by design.}
    \label{fig:dlo-tip-nom-control-experiment-snapshots}
    \vspace{-0.3cm}
\end{figure}

To validate the tip-based nominal controller \eqref{eqn:nom_controller}, we enable only this control law (disabling all safety constraints). 
We test two naturally collision-free setups by placing the DLO either $0.5\,\mathrm{m}$ above the box or directly on top of it (i.e., $h=-0.5$ m or $h=0.0$ m, with $d=-0.5$ m). 
Fig.~\ref{fig:dlo-tip-nom-control-experiment-snapshots} shows how the proportional controller successfully drives the DLO tips to their targets using the tip Jacobian \eqref{eqn:tip_jacobian}.

\subsubsection{Case 2: Tip Control + Obstacle Avoidance}
\begin{figure}[!htbp]
    \centering
    \begin{subfigure}[t]{0.4937\columnwidth}
        \centering
        \includegraphics[width=\textwidth]{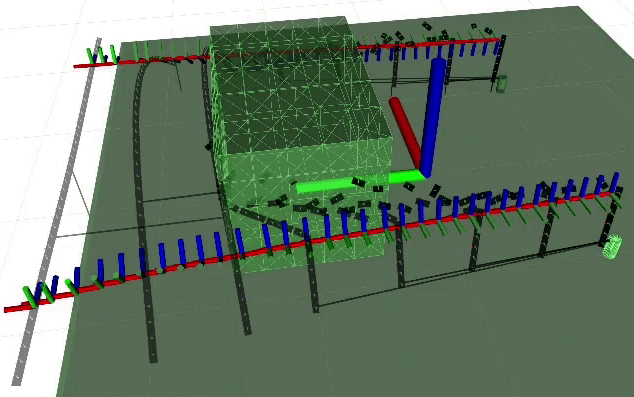}
        \caption{Collision avoidance OFF; snapshots at $t=\{0,1,2,3,5,7,10,20\}\,\mathrm{s}$.} 
        \label{fig:coll-avoid-off}
    \end{subfigure} \hfill
    \begin{subfigure}[t]{0.4937\columnwidth}
        \centering
        \includegraphics[width=\textwidth]{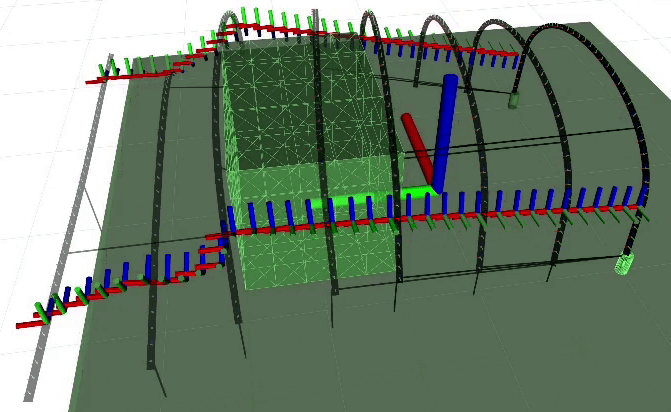}
        \caption{Collision avoidance ON; snapshots at $t=\{0,1,3,4,6,8,12,33\}\,\mathrm{s}$.} 
        \label{fig:coll-avoid-on}
    \end{subfigure}
    
    \caption{Verification of collision-avoidance constraints \eqref{eqn:lin_constraint_coll_avoidance} in a tent-building simulation. The DLO is placed behind the box, and collision avoidance is toggled on/off. Without avoidance (left), the DLO hooks on the box and fails. With avoidance (right), the robots lift it over the box to complete the task.}
    \label{fig:coll-avoid-on-vs-off-quick-verification}
    \vspace{-0.3cm}
\end{figure}

For the assessment of the collision-avoidance constraints \eqref{eqn:lin_constraint_coll_avoidance}, we position the DLO behind the box at $d=1.0\,\mathrm{m}$ and $h=0.5\,\mathrm{m}$, then run two tests: one with obstacle avoidance disabled and one with it enabled. As shown in Fig.~\ref{fig:coll-avoid-on-vs-off-quick-verification}, disabling collision avoidance (Fig.~\ref{fig:coll-avoid-off}) causes the nominal controller to drive the tips directly toward the target, hooking the DLO on the obstacle and failing the task. When collision avoidance is active, the controller minimally adjusts the nominal input to lift the DLO over the box (Fig.~\ref{fig:coll-avoid-on}). 
A closer look at the robot trajectories reveals only the +$z$-axis motion changes significantly, by approximately $20\,\mathrm{cm}$ on one robot and $40\,\mathrm{cm}$ on the other near $t=3\,\mathrm{s}$, allowing the DLO to clear the obstacle before descending to the target poses.

\begin{figure}[!hbtp]
  \begin{minipage}[c]{0.4\columnwidth}
    \includegraphics[width=0.95\columnwidth]{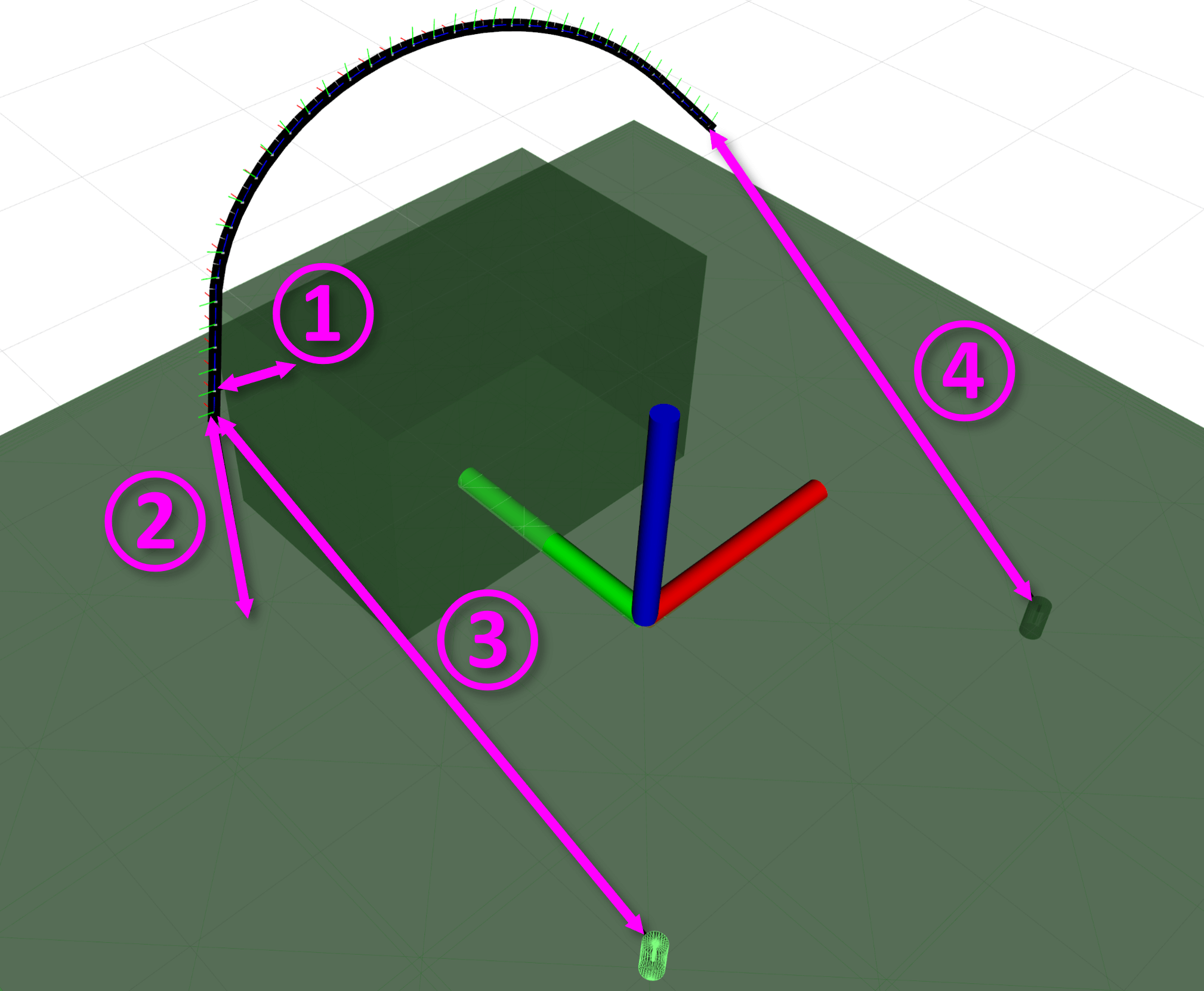}
  \end{minipage}\hfill
  \begin{minipage}[c]{0.6\columnwidth}
    \caption{%
      Reported minimum distances in the tent-building task. 
      With a single collision-avoidance constraint, only \Circled{1} (the overall minimum distance) is used in \eqref{eqn:qp_controller}. 
      By contrast, multiple constraints can include \Circled{1}, \Circled{2}, \Circled{3}, and \Circled{4}, 
      improving obstacle-avoidance performance for the local controller.
    }
    \label{fig:single-vs-multi-min-distances-snapshot}
  \end{minipage}
  \vspace{-0.3cm}
\end{figure}

\begin{figure}[!htbp]
    \centering
    \begin{subfigure}[t]{0.4437\columnwidth}
        \centering
        \includegraphics[width=\textwidth]{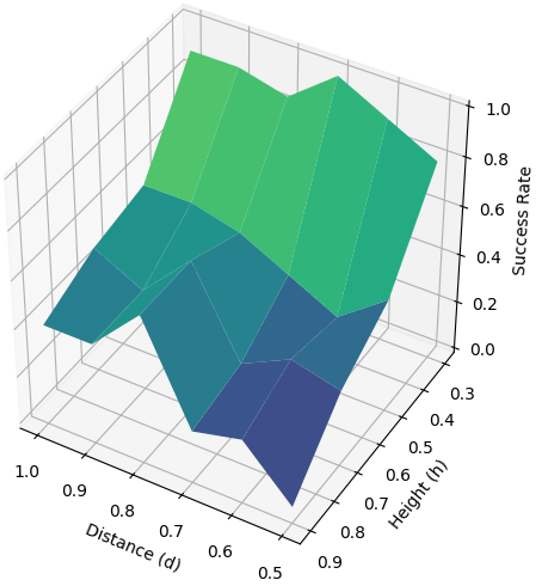}
        \caption{Single coll. avoidance constraint.} 
        \label{fig:coll-avoid-single-const-allowed}
    \end{subfigure} \hfill
    \begin{subfigure}[t]{0.5437\columnwidth}
        \centering
        \includegraphics[width=\textwidth]{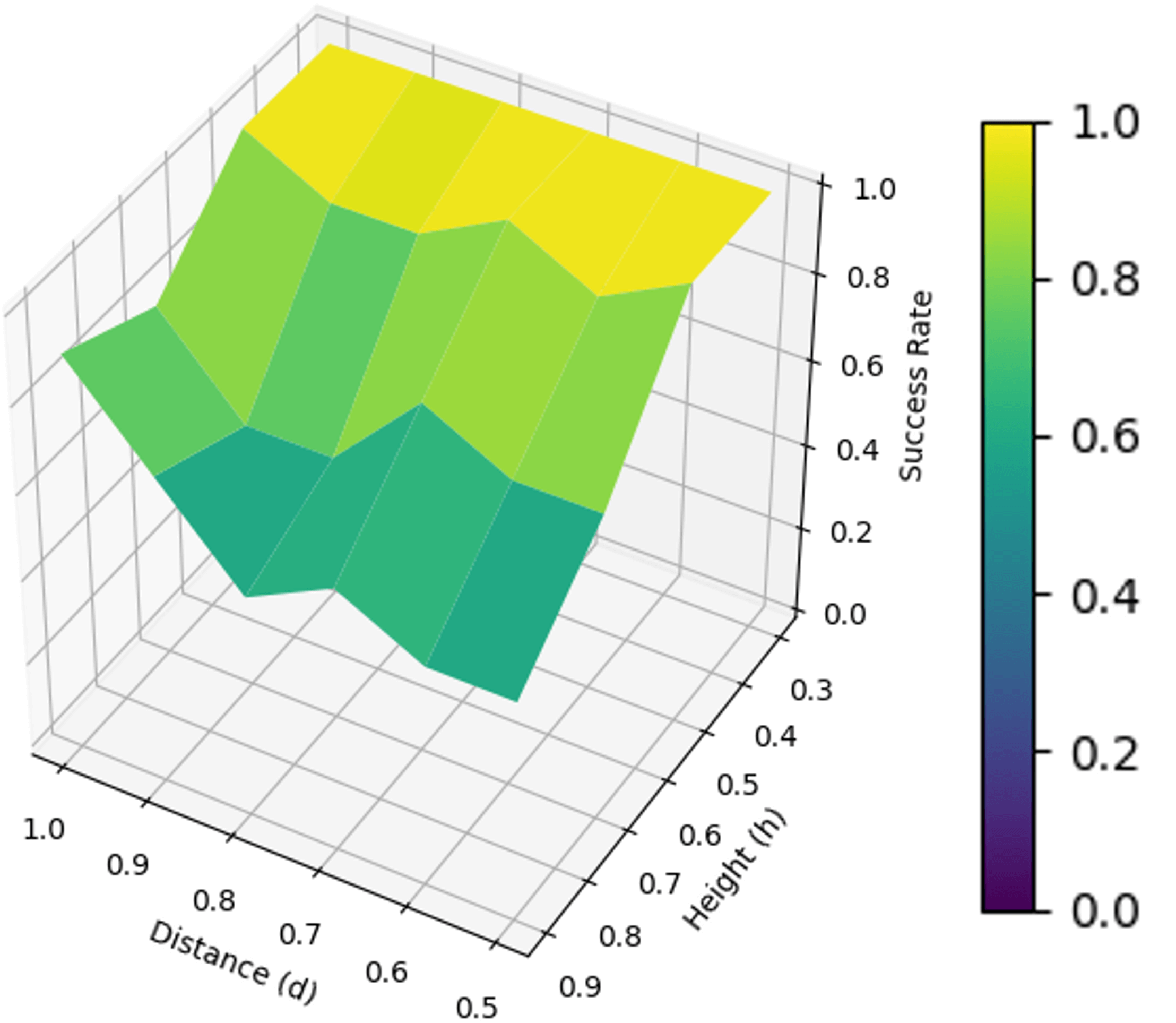}
        \caption{Multiple coll. avoidance constraints.} 
        \label{fig:coll-avoid-multi-const-allowed}
    \end{subfigure}
    
    \caption{Average success rates for \textbf{collision avoidance} in the tent-building task from various $(h,d)$ pairs, where $h\in\{0.3,0.5,0.7,0.9\}$ and $d\in\{0.5,0.6,0.7,0.8,0.9,1.0\}$. While avoidance generally works, success diminishes with smaller $d$ and larger $h$, illustrating local-control limitations. Multiple constraints (right) consistently outperform a single constraint (left).}
    \label{fig:coll-avoid-multi-vs-single-constraints}
    \vspace{-0.3cm}
\end{figure}

\paragraph{Single vs.\ Multiple Minimum Distances}
\label{parag:single-vs-multi-min-dist}
The controller’s “lift-up” action, instead of lowering the DLO, occurs because each minimum distance reported within a threshold proximity is treated as a separate collision-avoidance constraint in the QP \eqref{eqn:qp_controller}. As illustrated in Fig.~\ref{fig:single-vs-multi-min-distances-snapshot}, including only a single constraint based on the global minimum distance (\Circled{1} from the obstacle box), can initially drive the DLO downward to deactivate \Circled{1}. However, this activates another constraint (\Circled{2} from the ground), and the controller can get stuck toggling between these constraints. 
Allowing multiple collision-avoidance constraints keeps both \Circled{1} and \Circled{2} simultaneously active, generating a feasible motion that lifts the DLO. Although this approach resolves the specific limitation mentioned by \cite{2024aksoy}, it does not guarantee avoidance of local minima in all cases.
To quantify obstacle-avoidance performance and show that purely local control methods can still fail under certain conditions, we sample $h \in \{0.3,0.5,0.7,0.9\}\,\mathrm{m}$ and $d \in \{0.5,0.6,0.7,0.8,0.9,1.0\}\,\mathrm{m}$, testing each pair $(h,d)$ 10 times, with minor random variations in the initial poses. We compare a single-constraint approach (only the global minimum distance) to a multi-constraint approach (all reported minima). 
Fig.~\ref{fig:coll-avoid-multi-vs-single-constraints} shows the average success rates, defined by avoiding collisions (distance never reaching zero) and placing the DLO tips in the grommets (avoiding deadlocks). Although the collision avoidance  generally works, success rates decline as $d$ decreases and $h$ increases--highlighting the limitations of a local controller and the need for global planning. In all configurations, using multiple constraints consistently achieves higher or equal success rates compared to a single constraint.

\subsubsection{Case 3: Tip Control + Obstacle \& Overstress Avoidance}
\begin{figure}[!htbp]
    \centering
    \begin{subfigure}[t]{0.4937\columnwidth}
        \centering
        \includegraphics[width=\textwidth]{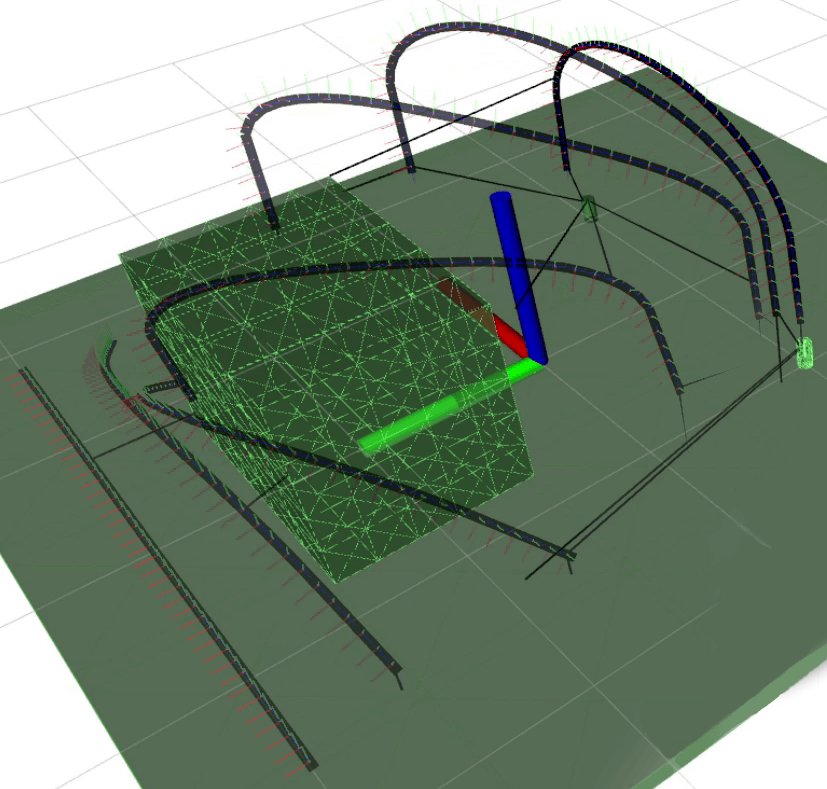}
        \caption{Overstress avoidance OFF; snapshots at $t=\{0,10,24,50,84,98,164\}\,\mathrm{s}$.}
        \label{fig:stress-avoid-off}
    \end{subfigure}\hfill
    \begin{subfigure}[t]{0.4937\columnwidth}
        \centering
        \includegraphics[width=\textwidth]{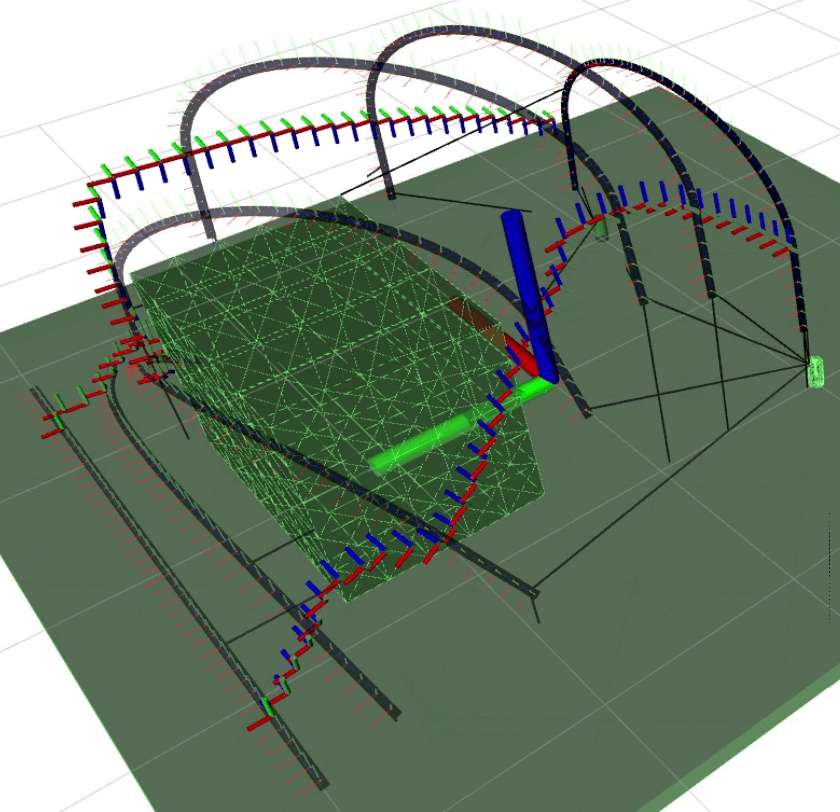}
        \caption{Overstress avoidance ON; snapshots at $t=\{0,7,18,33,47,60,120\}\,\mathrm{s}$.}
        \label{fig:stress-avoid-on}
    \end{subfigure}
    \caption{Verifying \textbf{stress-avoidance constraints} \eqref{eqn:lin_constraint_stress_avoidance} in a tent-building simulation. Placing the DLO at $d=0.5\,\mathrm{m}$ and $h=0.9\,\mathrm{m}$, we toggle overstress avoidance on/off. Without it (left), the obstacle is avoided, but the DLO bends severely (peak $F\approx5200\,\mathrm{N}$, $T\approx250\,\mathrm{Nm}$). With overstress avoidance (right), the shape remains smoother (peak $F\approx26\,\mathrm{N}$, $T\approx9\,\mathrm{Nm}$).}
    \label{fig:stress-avoid-on-vs-off-quick-verification}
    \vspace{-0.3cm}
\end{figure}

While collision avoidance prevents obstacle contact, the DLO can still be overstressed without explicit stress-avoidance constraints \eqref{eqn:lin_constraint_stress_avoidance}. Fig.~\ref{fig:stress-avoid-on-vs-off-quick-verification} demonstrates this by placing the DLO at $(d,h)=(0.5,\,0.9)\,\mathrm{m}$. Without overstress avoidance, peak F/T values reported by the PBD simulation reached $5200\,\mathrm{N}$ and $250\,\mathrm{Nm}$. Enabling overstress avoidance with $F^\text{max}=200\,\mathrm{N}$, $T^\text{max}=15\,\mathrm{Nm}$, $F^\text{ofs}=160\,\mathrm{N}$, and $T^\text{ofs}=11.5\,\mathrm{Nm}$ reduced these peaks to $26\,\mathrm{N}$ and $9\,\mathrm{Nm}$. Visually, the DLO maintained a smoother shape (Fig.~\ref{fig:stress-avoid-on}), indicating lower stress.

\begin{figure}[!hbtp]
  \begin{minipage}[c]{0.55\columnwidth}
    \includegraphics[width=0.95\columnwidth]{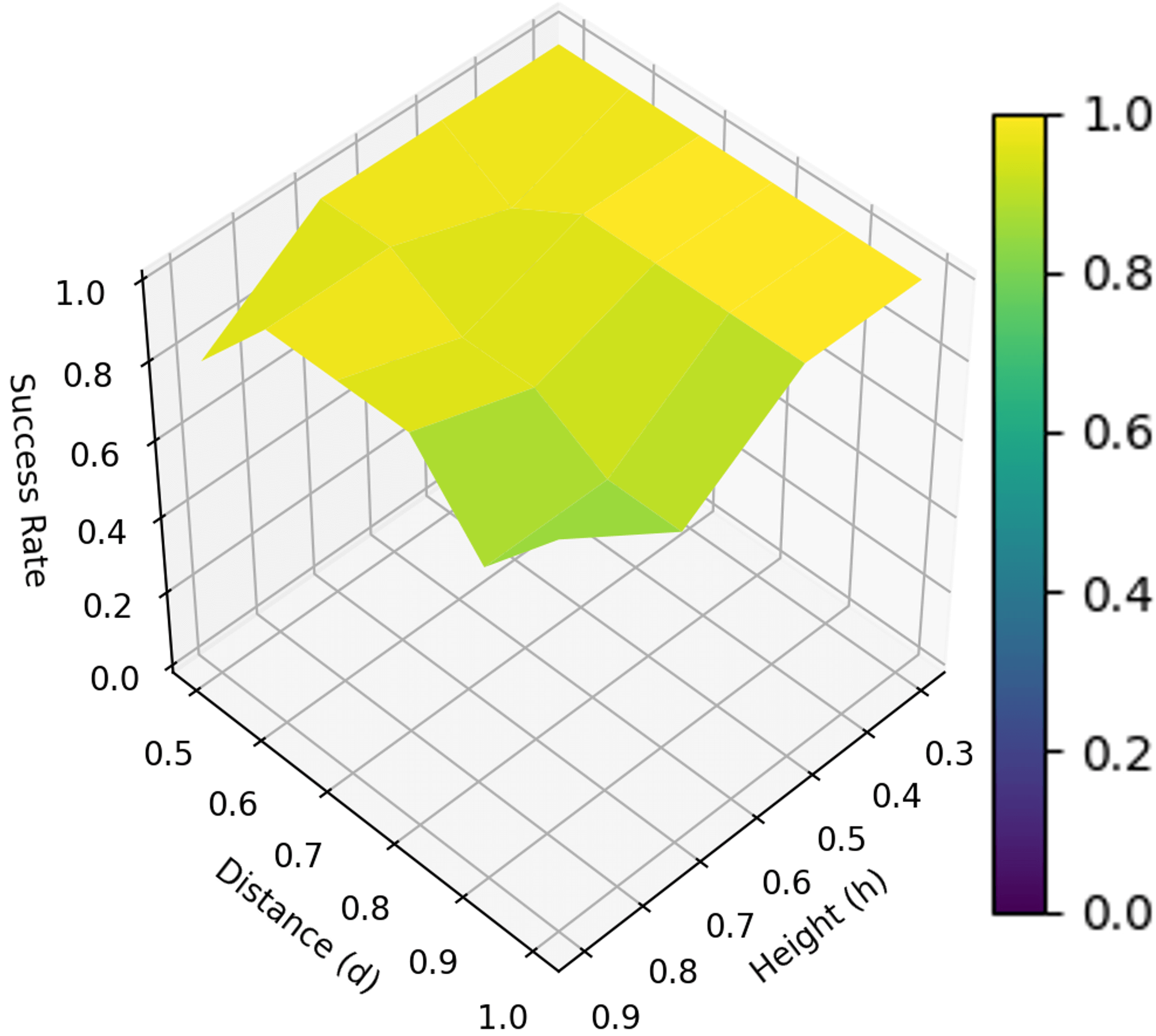}
  \end{minipage}\hfill
  \begin{minipage}[c]{0.45\columnwidth}
    \caption{Average success rates with \textbf{stress and collision avoidance} both enabled. The DLO starts from various $(h,d)$ pairs, where $h\in\{0.3,0.5,0.7,0.9\}$ and $d\in\{0.5,0.6,0.7,0.8,0.9,1.0\}$. Compared to Fig.~\ref{fig:coll-avoid-multi-const-allowed} (collision avoidance only), stress avoidance further improves overall success.}
    \label{fig:stress-avoid-success-rates}
  \end{minipage}
  \vspace{-0.3cm}
\end{figure}

To further evaluate overstress avoidance, we repeated the sampling of $h$ and $d$ from the previous subsection (Fig.~\ref{fig:coll-avoid-multi-vs-single-constraints}) and measured overall task success (obstacle avoidance \emph{and} no overstress). Fig.~\ref{fig:stress-avoid-success-rates} shows that adding stress-avoidance constraints further improves success rates compared to collision avoidance alone.

Regarding overstress statistics, high overstress events occurred most often for smaller $d$ and larger $h$, where the controller attempted more aggressive bends to avoid obstacles. Specifically, around $(d,h)=(0.5,\,0.9)\,\mathrm{m}$, $80\%$ of trials experienced overstress without the stress-avoidance feature. In other $(d,h)$ combinations, overstress occurred in about $10\%$ of trials. With stress avoidance enabled, no overstress was observed in any sampled cases (i.e., $100\%$ overstress avoidance).

\subsubsection{Case 4: Blending Path Tracking \& Tip Control with Safety}
We now demonstrate combining path tracking and tip control (\S\ref{subsubsec:path-tracking}--\ref{subsubsec:blending-nom-control}) with both collision and overstress avoidance. A simple mock path is formed by linearly interpolating three waypoints between the initial DLO position and the grommet holes. The DLO is placed at $(d,h)=(0.1,\,0.9)\,\mathrm{m}$, very close to the obstacle—where the local controller alone would fail without global guidance.

\begin{figure}[!htbp]
    \centering
    \begin{subfigure}[t]{0.4937\columnwidth}
        \centering
        \includegraphics[width=\textwidth]{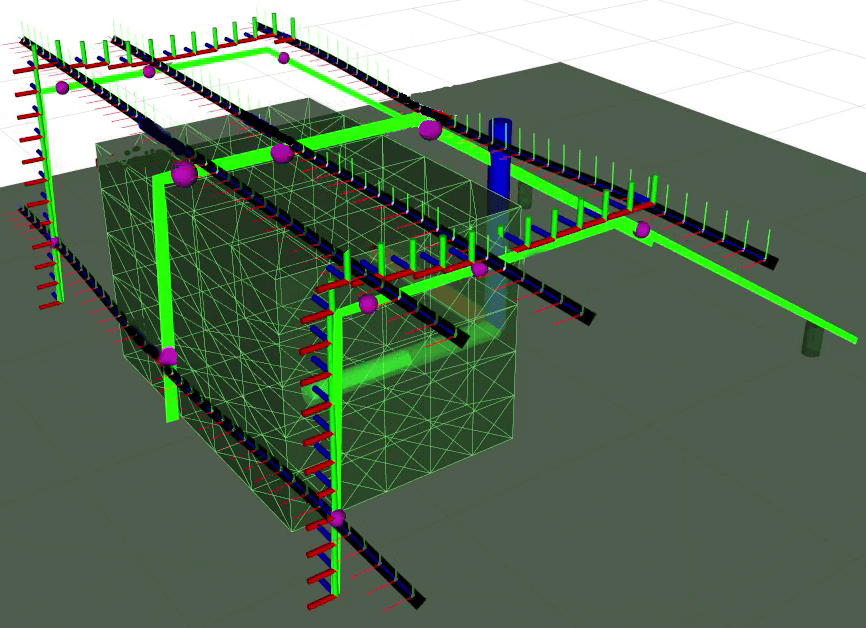}
        \caption{Snapshots at $t=\{4,20,28,40\}\,\mathrm{s}$.}
    \end{subfigure}
    \begin{subfigure}[t]{0.4937\columnwidth}
        \centering
        \includegraphics[width=\textwidth]{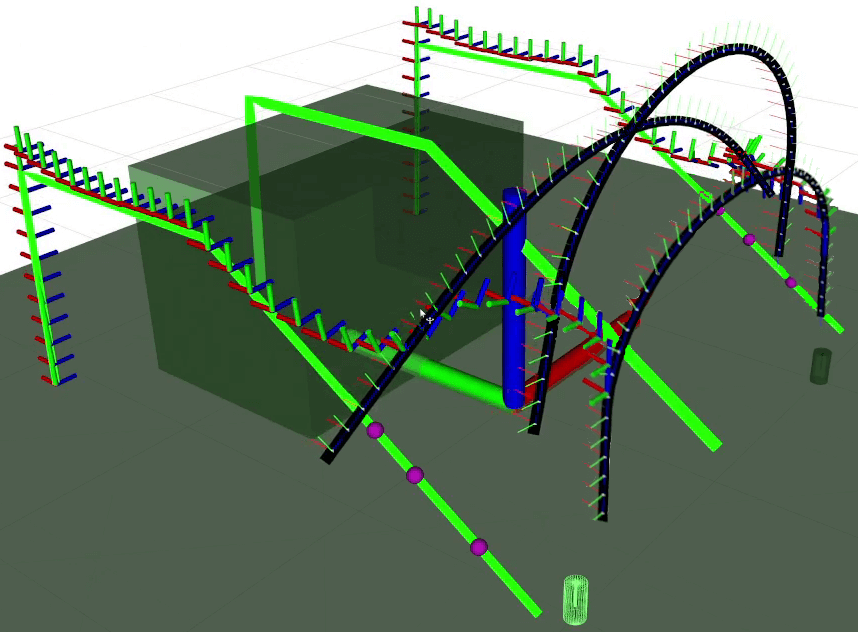}
        \caption{Snapshots at $t=\{56,72,132\}\,\mathrm{s}$.}
    \end{subfigure}
    \caption{Path tracking \eqref{eqn:nom_controller_path_tracking5} and control blending \eqref{eqn:nom_controller_blended} 
  in a tent-building simulation with $(d,h)=(0.1,\,0.9)\,\mathrm{m}$. 
  A simple path of three linearly interpolated waypoints (green lines) guides the robot holding points, 
  while magenta spheres mark the current targets and traces show the executed paths. 
  Initially, the controller follows the path closely. As collision/overstress constraints activate, 
  minor deviations occur, and near the end the control transitions from path tracking to tip control, 
  enabling a collision-free approach to the grommet holes.}
    \label{fig:test-path-tracking-and-blending}
    \vspace{-0.3cm}
\end{figure}

This mock path assumes the DLO remains ``straight,'' akin to a rigid link. As seen in Fig.~\ref{fig:test-path-tracking-and-blending}, the robots initially follow the path closely because no safety constraints are activated. Near the first “kink” at the obstacle, collision avoidance momentarily deviates the controller’s motion. While crossing the top surface of the box, minor deviations persist until the robots are about $1\,\mathrm{m}$ from the path end. At that point, the blending scheme shifts from path tracking to tip control with a smooth transition. To avoid ground contact, the robots lift the DLO before descending into the grommet holes, completing the task safely.

\subsubsection{Local Controller Runtime Performance}
\label{subsubsec:local-controller-runtime-performance}
Table~\ref{tab:local-controller-execution-times} lists average controller loop rates under various constraint configurations in the tent-building task. Two QP solvers in Python (OSQP~\cite{osqp} and Clarabel~\cite{clarabel} via CVXPY~\cite{cvxpy}) are tested, spanning scenarios from nominal control only (collision/stress avoidance off) to collision avoidance (single or multiple distances) and overstress avoidance.
Overall, OSQP outperforms Clarabel in all tested setups, although Clarabel may excel in non-linear constraints (e.g., quadratic velocity limits). The simplest case (nominal control only) achieves the highest rate of $50.3\,\mathrm{Hz}$. Enabling collision avoidance with multiple distance constraints yields about $39.6\,\mathrm{Hz}$, and adding overstress avoidance slightly decreases this to $39.2\,\mathrm{Hz}$. These results indicate that the controller remains real-time capable even with added safety constraints. For environments containing numerous collision objects, limiting simultaneous constraints and ignoring distant obstacles can preserve real-time performance.

\begin{table}[!htbp]
    \centering
    \resizebox{0.9\columnwidth}{!}{
    \begin{tabular}{@{}l l c c c@{}}
        \toprule
        \textbf{Solver} 
          & \makecell{\textbf{Overstress}\\\textbf{Avoidance}}
          & \makecell{\textbf{Collision}\\\textbf{OFF}}
          & \makecell{\textbf{Collision ON}\\(\textbf{Multiple Dist.})} 
          & \makecell{\textbf{Collision ON}\\(\textbf{Single Dist.})} \\
        \midrule
        \multirow{2}{*}{\textbf{OSQP}}
          & OFF & 50.3 & 39.6 & 41.7 \\
          & ON  & 46.4 & 39.2 & -- \\
        \midrule
        \multirow{2}{*}{\textbf{Clarabel}}
          & OFF & 43.4 & 36.0 & --  \\
          & ON  & 40.0 & 34.3 & -- \\
        \bottomrule
    \end{tabular}
    }
    \caption{%
      Average local controller loop rates (Hz) under different solver and constraint setups. 
      ``Collision ON (Single Dist.)'' uses only the global minimum-distance constraint, whereas ``Collision ON (Multiple Dist.)'' includes multiple reported distances. ``--'' indicates untested combinations.
    }
    \label{tab:local-controller-execution-times}
    \vspace{-0.3cm}
\end{table}

\subsection{Global Planner + Local Controller Evaluations}
We next integrate the local controller 
with the global planner 
to demonstrate our hybrid method in an L-shaped corridor maneuvering task and compare its performance against recent state-of-the-art approaches.

\subsubsection{L-shaped Corridor Maneuvering Task}
\begin{figure}[!hbtp]
    \centering
    \includegraphics[width=0.80\columnwidth]{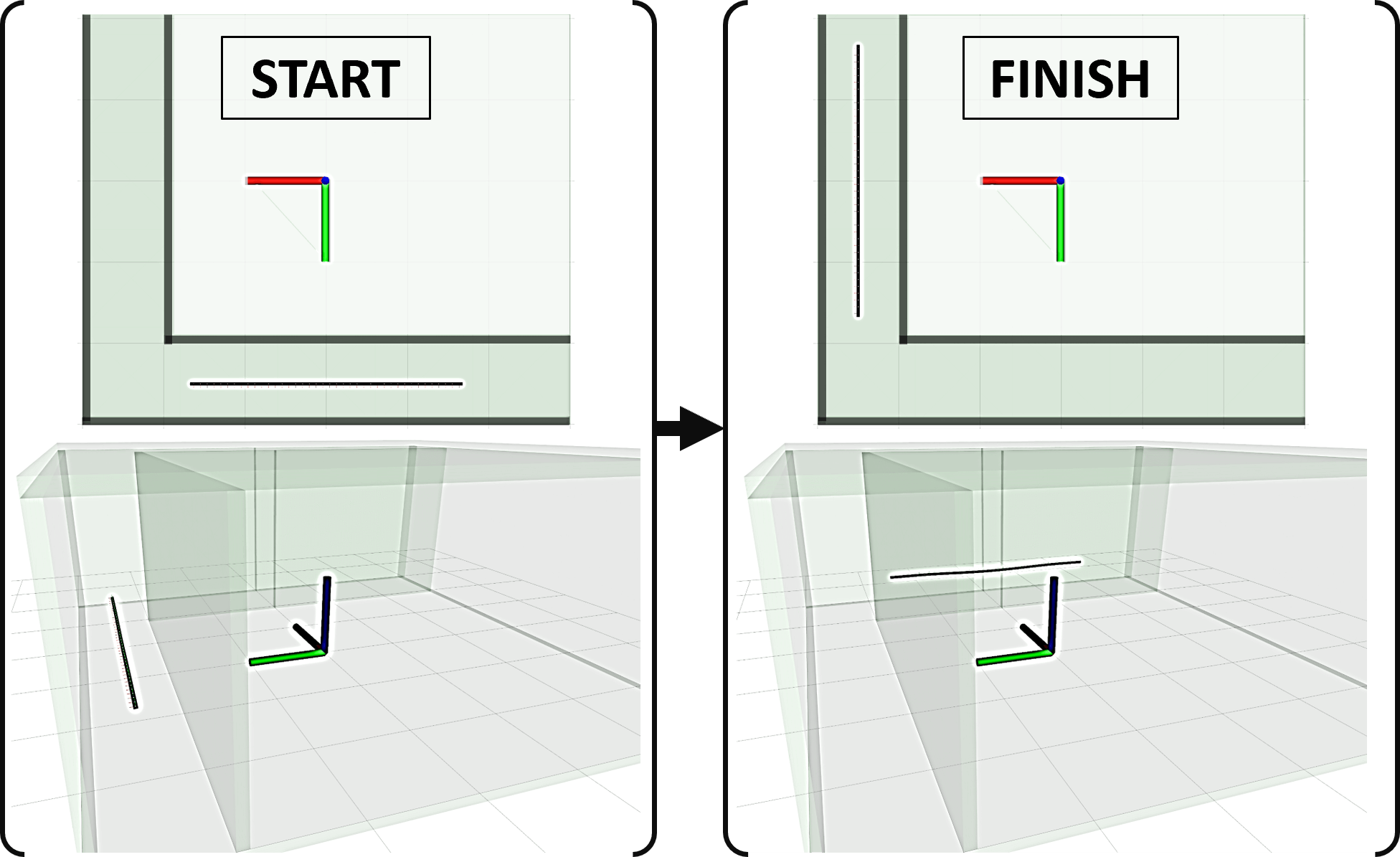}
    \caption{\textbf{L-shaped corridor maneuvering task.} An initially straight DLO is guided through a right-angle corridor and reoriented by $90^\circ$ at the exit. Top: top view. Bottom: oblique view.}
    \label{fig:corridor-maneuvering-task}
\end{figure}

\begin{figure}[!hbtp]
  \begin{minipage}[c]{0.55\columnwidth}
    \includegraphics[width=0.95\columnwidth]{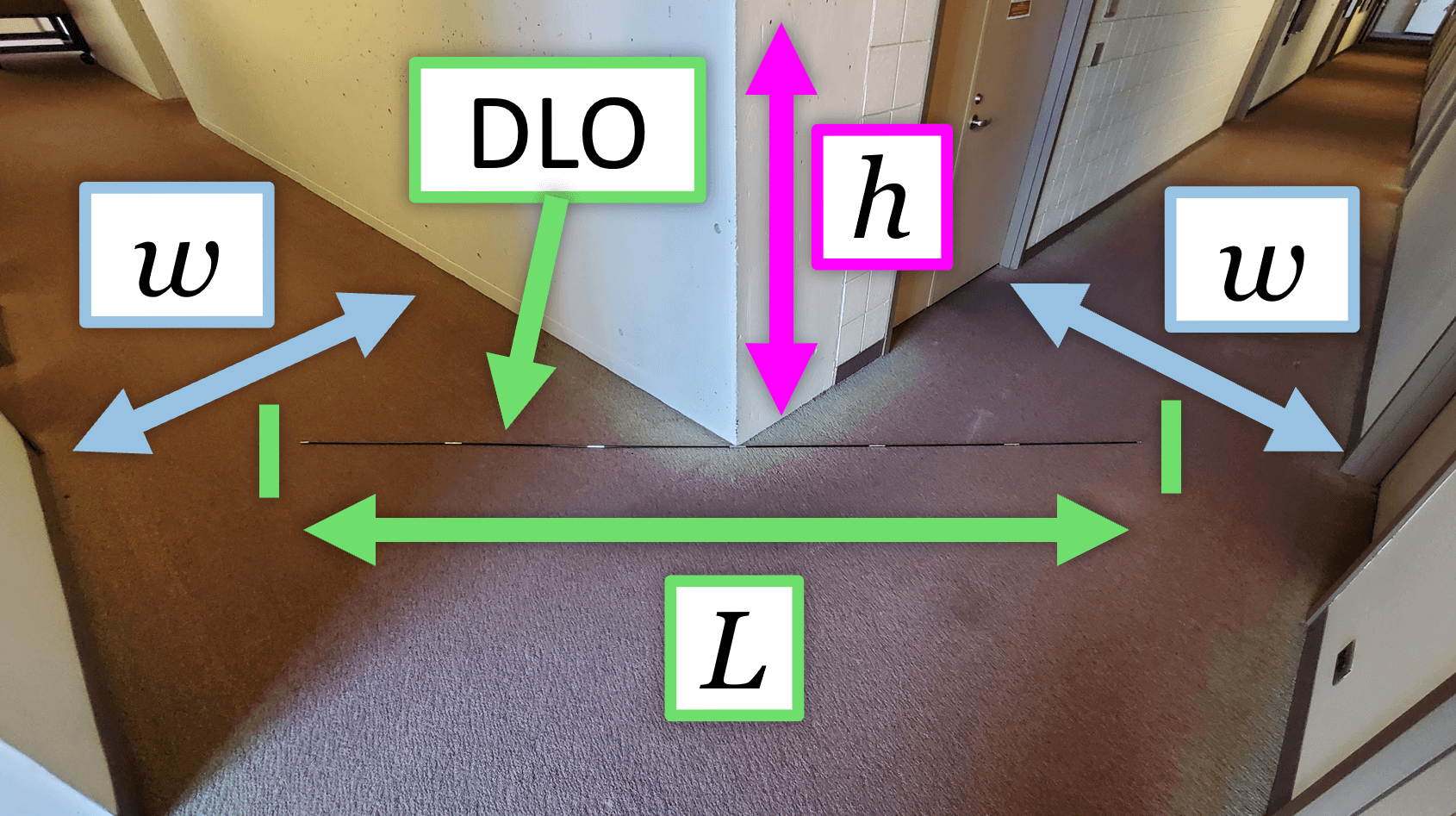}
  \end{minipage}\hfill
  \begin{minipage}[c]{0.45\columnwidth}
    \caption{A corridor with equal widths $w$ and ceiling height $h$. A rigid link of length $L$ can only pass around the corner without bending if $w \ge \sqrt{\tfrac{L^2 - h^2}{8}}$. Otherwise, flexibility is crucial.}
    \label{fig:corridor-pic}
  \end{minipage}
\end{figure}

In this scenario, an initially straight DLO must traverse a right-angle corridor and emerge in a $90^\circ$ rotated, straight configuration (Fig.~\ref{fig:corridor-maneuvering-task}). Such a setup models narrow passageways (see Fig.~\ref{fig:DLO-cartoon}), where a purely rigid link of length $L$ can only pass if the corridor width $w$ and ceiling height $h$ satisfy
\(
w \;\ge\; \sqrt{\tfrac{L^2 - h^2}{8}}.
\)
When $w$ is smaller than this threshold, bending is necessary, illustrating the importance of combining global planning with local control for DLO manipulation in tight spaces (Fig.~\ref{fig:corridor-pic}).


\subsubsection{Case 1: Verify Local Controller Alone is Insufficient}
\begin{figure}[!htbp]
    \centering
    \begin{subfigure}[t]{0.3437\columnwidth}
        \centering
        \includegraphics[width=\textwidth]{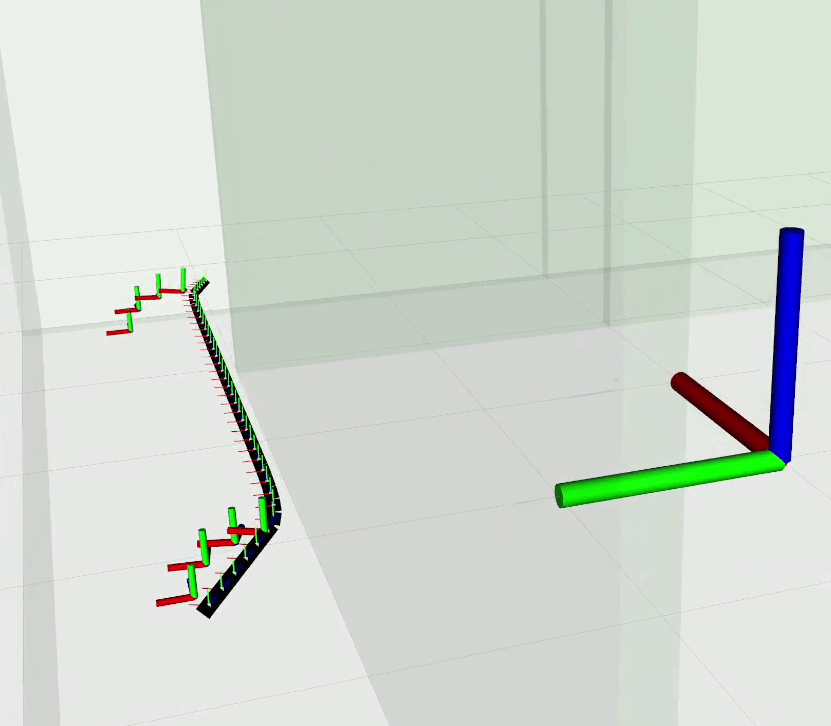}
        \caption{DLO stuck shortly after control begins.}
        \label{fig:corridor-w1.0-local-contr-stuck}
    \end{subfigure}
    \begin{subfigure}[t]{0.5437\columnwidth}
        \centering
        \includegraphics[width=\textwidth]{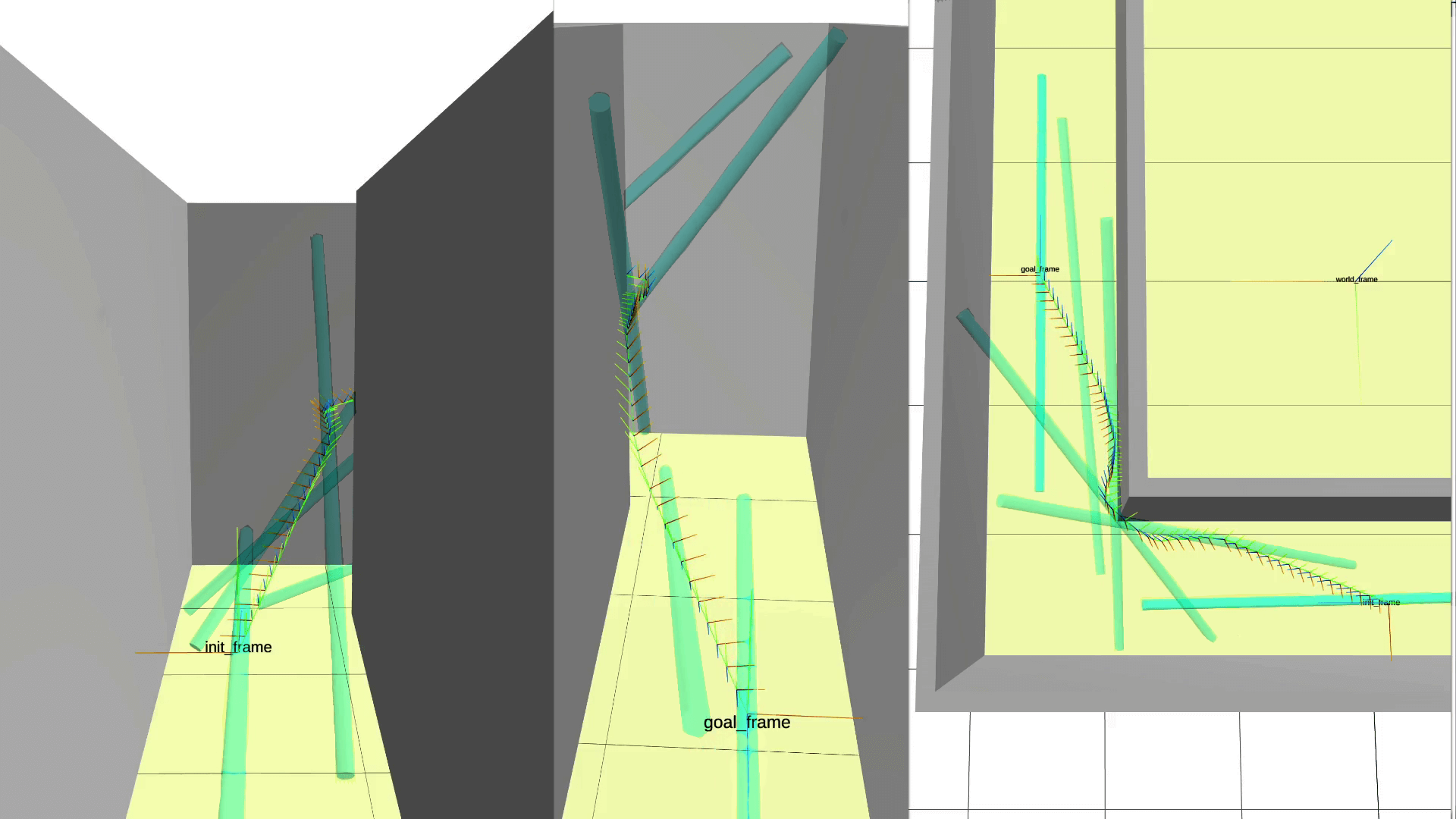}
        \caption{Snapshots of the planner-generated path for a single rigid link. Left: view from the start; middle: target corridor end; right: top view.}
        \label{fig:corridor-w1.0-single-link-plan}
    \end{subfigure}
    \caption{Necessity of global planning in an L-shaped corridor of width $w=1.0\,\mathrm{m}$. Although the DLO’s length is within rigid feasibility limits, the local controller alone becomes stuck (left), while the global planner easily finds a collision-free path (right).}
    \label{fig:corridor-verify-local-control-not-sufficient}
    \vspace{-0.1cm}
\end{figure}

Consider the same DLO from the previous subsection in a corridor of width $w=1.0\,\mathrm{m}$ and ceiling height $h=2.4\,\mathrm{m}$. Since a rigid link of length $L$ could navigate this corridor if $w\ge0.83\,\mathrm{m}$, the task is theoretically feasible without bending. However, relying solely on the local controller quickly leads the DLO to become stuck against the nearest wall, as shown in Fig.~\ref{fig:corridor-w1.0-local-contr-stuck}. Although collisions and overstress are prevented, the local controller lacks a mechanism to explore alternate paths and thus fails to reach the goal.
By contrast, the global planner easily finds a path for a rigid-link approximation of the same length (Fig.~\ref{fig:corridor-w1.0-single-link-plan}), confirming that a high-level plan is indispensable, even when the corridor is theoretically wide enough for a rigid link.

\subsubsection{Case 2: Verifying the Framework in a Narrow Corridor}
\begin{figure}[!hbtp]
  \centering
  \includegraphics[width=0.95\columnwidth]{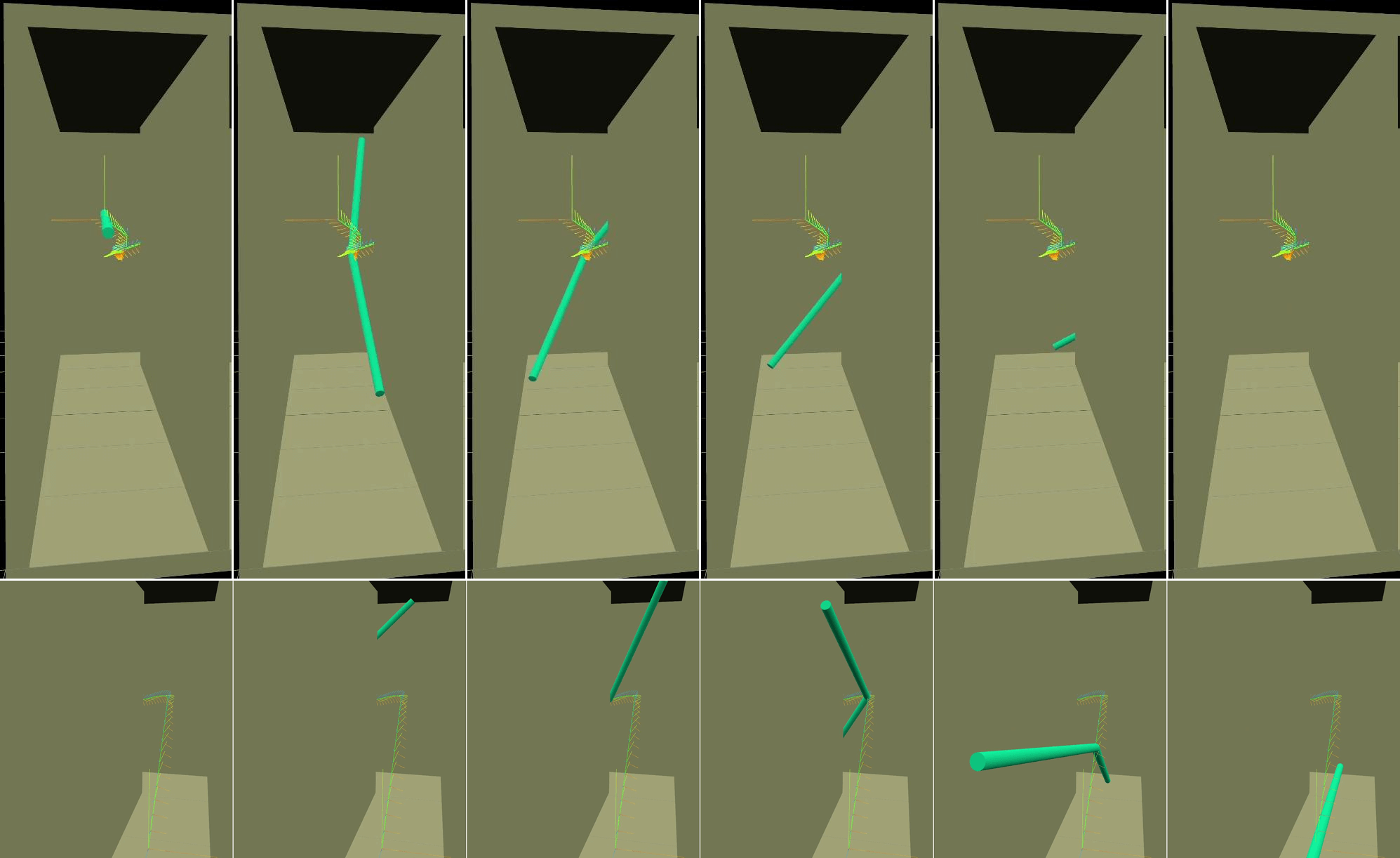}
  \caption{Planner-generated path using two rigid links in an $L$-shaped corridor of width $w=0.8\,\mathrm{m}$ (requires bending, so a single-link model fails). Top: view from the start; Bottom: view from the target end.}
  \label{fig:narrow-corridor-plan}
\end{figure}

We reduce the corridor width to $w=0.8\,\mathrm{m}$, below the $0.83\,\mathrm{m}$ threshold for a rigid link of zero thickness. As described in Section~\ref{sec:global_planner}, the planner first attempts a single rigid link and fails, as expected. It then increases the number of links to two, connected by a spherical joint (\S\ref{subsec:dlo-approximation}), and regenerates the URDF (\S\ref{subsec:urdf-creation-n-kinematics}). This two-link approximation successfully finds a path, shown in Fig.~\ref{fig:narrow-corridor-plan}.

\begin{figure}[!hbtp]
    \centering
    \includegraphics[width=0.90\columnwidth]{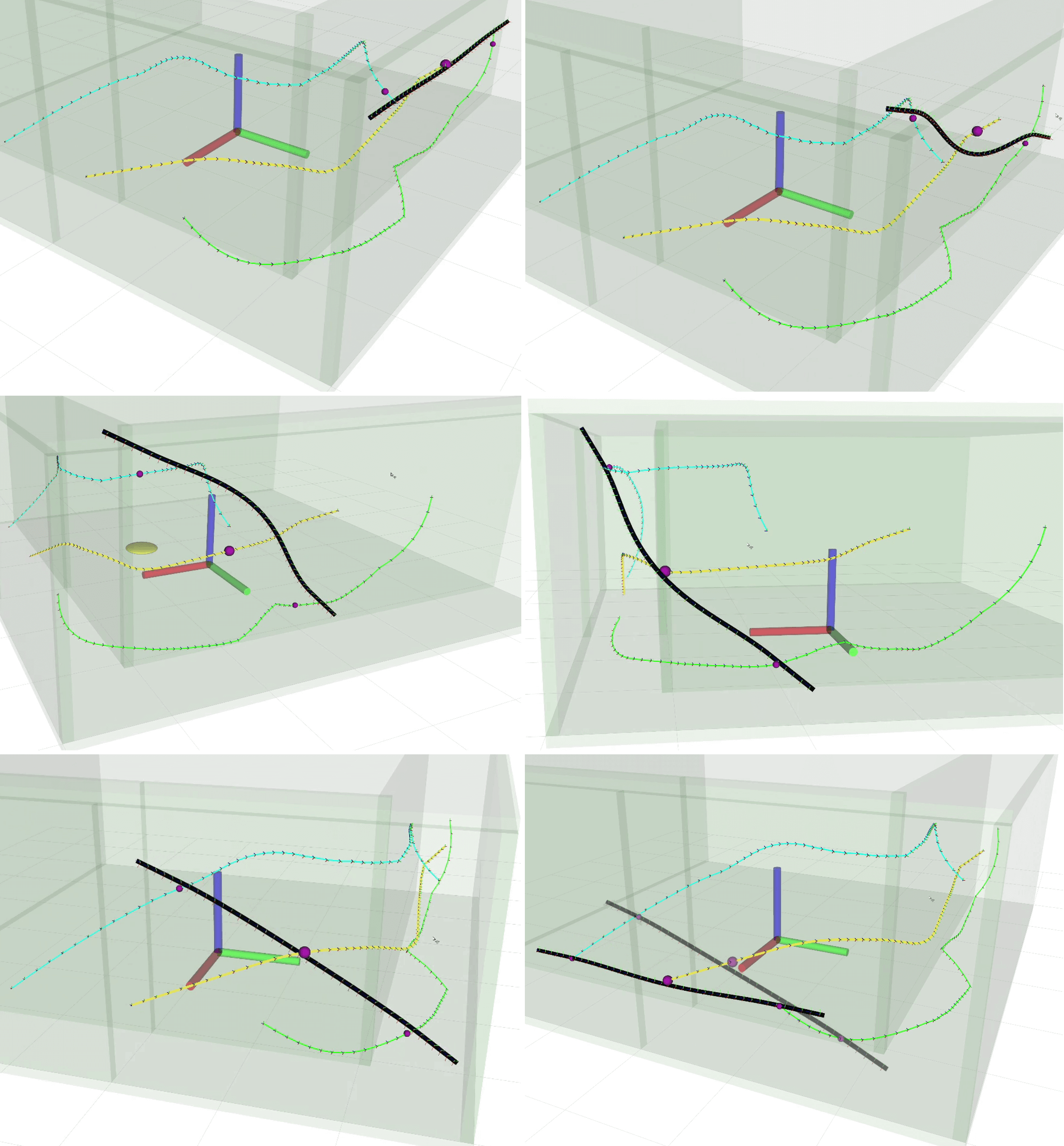}
    \caption{(Left to right, top to bottom) Execution of the planned path in the narrow-corridor task at $t=\{0,7,38,58,86,100,110\}\,\mathrm{s}$. Green and cyan curves depict the guidance paths; the yellow curve (center) shows the centroid of the two-link approximation for visualization.}
    \label{fig:narrow-corridor-plan-exec}
    \vspace{-0.1cm}
\end{figure}

After planning, we extract robot waypoints (\S\ref{subsec:robot-traj-gen-from-path-planning}) for the local controller, as seen in Fig.~\ref{fig:narrow-corridor-plan-exec}. The robots closely follow the plan, with no safety constraints triggered. Notably, the PBD simulation in the local controller provides higher-fidelity DLO states than the simplified two-link planner model. This highlights an important insight of our approach: the planner does not require detailed knowledge of the DLO characteristics; a simple model often suffices to generate a viable coarse path.

\subsubsection{Benchmarks in Challenging Simulation Scenes}
\begin{figure}[!htbp]
  \centering
  \includegraphics[width=0.99\columnwidth]{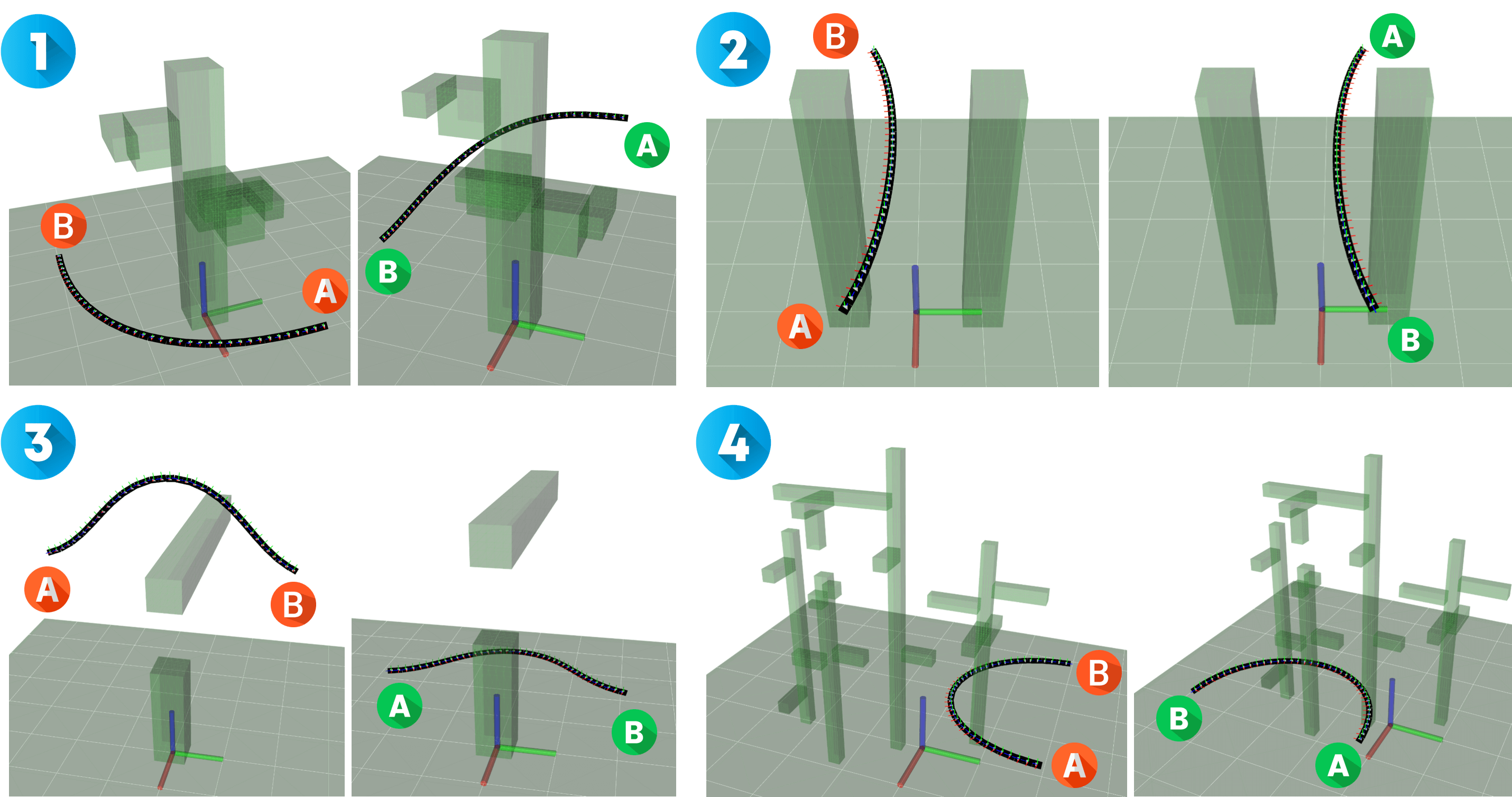}
  \caption{Four benchmark scenes reproduced from \cite{2024yu}. The DLO must travel from the initial poses (left) to the final poses (right), held at its ends (\CircledOrange{A} and \CircledOrange{B} to \CircledGreen{A} and \CircledGreen{B}). Some tasks require flipping the DLO.}
  \label{fig:benchmark-scenes}
  \vspace{-0.3cm}
\end{figure}

We compare our method against recent state-of-the-art techniques~\cite{2024yu,2023yu2,2020mcconachiea} by replicating their collision scenes and tasks from \cite{2024yu}, labeled \CircledCyan{1}--\CircledCyan{4} in Fig.~\ref{fig:benchmark-scenes}. Tasks \CircledCyan{1}--\CircledCyan{3} are relatively straightforward, while \CircledCyan{4} is highly cluttered.
We use the same material properties as in the tent-pole tasks, but shorten the DLO length to $0.5\,\mathrm{m}$. Following \cite{2024yu}, both ends of the DLO are grasped within a $(1.0 \times 1.0 \times 1.5)\,\mathrm{m}$ workspace. Each task is repeated 100 times with automatic segment selection in the planner, and the DLO is discretized into 40 segments in PBD. The automatic segment selection used $4, 2, 3,$ and $5$ link approximations for tasks \CircledCyan{1}--\CircledCyan{4}, respectively. For comparison to \cite{2024yu}, which uses 10 segments, we also fix our planner to 10 segments and run an additional 100 trials.
\begin{table*}[!htbp]
  \centering
  \resizebox{\textwidth}{!}{
    \begin{threeparttable}[b]
      \begin{tabular}{@{}l l c c c c c c c c c c@{}}
          \toprule
          \textbf{Task} 
            & \makecell{\textbf{Method}}
            & \makecell{\textbf{Avr. Num.}\\\textbf{of}\\\textbf{Segments}}
            & \makecell{\textbf{DLO}\\\textbf{Apprx.}\\\textbf{Err. (cm)}}
            & \makecell{\textbf{Planning}\\\textbf{Success}\\\textbf{Rate}}
            & \makecell{\textbf{DLO}\\\textbf{Simplification}\\\textbf{Time (s)\tnote{a}}}
            & \makecell{\textbf{DLO}\\\textbf{URDF Gen.}\\\textbf{Time (s)\tnote{a}}}
            & \makecell{\textbf{Feasible}\\\textbf{Path Plan.}\\\textbf{Time (s)\tnote{a}}}
            & \makecell{\textbf{Smoothing}\\\textbf{Path}\\\textbf{Time (s)\tnote{a}}}
            & \makecell{\textbf{TOTAL}\\\textbf{Planning}\\\textbf{Time (s)\tnote{a}}}
            & \makecell{\textbf{Feasible}\\\textbf{Path}\\\textbf{Length (m)\tnote{a}}}
            & \makecell{\textbf{Smoothed}\\\textbf{Path}\\\textbf{Length (m)\tnote{a}}} \\
          \midrule
          \multirow{5}{*}{\CircledCyan[]{1}}
            & \cite{2014bretl}\tnote{b}       & 10 (Fixed)              & --   & 100/100 & --                & --                & --              & --              & 13.21 $\pm$ 6.58          & --              & 2.02 $\pm$ 0.58 \\
            & \cite{2020mcconachiea}\tnote{b} & 10 (Fixed)              & --   &  78/100 & --                & --                & --              & --              & --                        & --              & --              \\
            & \cite{2024yu}\tnote{b}          & 10 (Fixed)              & --   & 100/100 & --                & --                & 1.92 $\pm$ 0.66 & 2.33 $\pm$ 0.4  &  4.25 $\pm$ 0.77          & 1.94 $\pm$ 0.39 & 1.04 $\pm$ 0.20 \\
            & \textbf{Ours}                   & 10 (Fixed)              & 0.13 & 100/100 & 0.016 $\pm$ 0.001 & 0.02 $\pm$1e-4    & 0.66 $\pm$ 0.22 & 1.24 $\pm$ 0.34 &  \textbf{1.94 $\pm$ 0.41} & 1.09 $\pm$ 0.17 & 0.96 $\pm$ 0.14 \\
            & \textbf{Ours}                   & 4 (Auto)                & 0.65 & 100/100 & 0.015 $\pm$ 0.001 & 0.01 $\pm$ 0.01   & 0.3 $\pm$ 0.11  & 0.38 $\pm$ 0.12 &  \textbf{0.71 $\pm$ 0.16} & 1.11 $\pm$ 0.17 & 0.96 $\pm$ 0.16 \\
          \midrule
          \multirow{5}{*}{\CircledCyan[]{2}}
            & \cite{2014bretl}\tnote{b}       & 10 (Fixed)              & --   & 100/100 & --                & --                & --              & --              &  7.00 $\pm$ 4.64          & --              & 1.22 $\pm$ 0.30 \\
            & \cite{2020mcconachiea}\tnote{b} & 10 (Fixed)              & --   & 100/100 & --                & --                & --              & --              & --                        & --              & --              \\
            & \cite{2024yu}\tnote{b}          & 10 (Fixed)              & --   & 100/100 & --                & --                & 1.43 $\pm$ 0.56 & 2.73 $\pm$ 0.34 &  4.16 $\pm$ 0.65          & 1.71 $\pm$ 0.36 & 0.86 $\pm$ 0.12 \\
            & \textbf{Ours}                   & 10 (Fixed)              & 0.02 & 100/100 & 0.016 $\pm$1e-4   & 0.02 $\pm$1e-4    & 0.17 $\pm$ 0.18 & 1.19 $\pm$ 0.27 &  \textbf{1.39 $\pm$ 0.35} & 1.03 $\pm$ 0.13 & 0.87 $\pm$ 0.06 \\
            & \textbf{Ours}                   & 2 (Auto)                & 0.29 & 100/100 & 0.004 $\pm$1e-4   & 0.01 $\pm$ 0.01   & 0.03 $\pm$ 0.03 & 0.18 $\pm$ 0.08 &  \textbf{0.23 $\pm$ 0.09} & 1.1  $\pm$ 0.11 & 0.89 $\pm$ 0.05 \\
            \midrule
            \multirow{5}{*}{\CircledCyan[]{3}}
            & \cite{2014bretl}\tnote{b}       & 10 (Fixed)              & --   & 100/100 & --                & --                & --              & --              & 21.01 $\pm$ 12.72         & --              & 1.29 $\pm$ 0.42 \\
            & \cite{2020mcconachiea}\tnote{b} & 10 (Fixed)              & --   & 91/100  & --                & --                & --              & --              & --                        & --              & --              \\
            & \cite{2024yu}\tnote{b}          & 10 (Fixed)              & --   & 100/100 & --                & --                & 3.18 $\pm$ 1.61 & 2.78 $\pm$ 0.45 &  5.96 $\pm$ 1.67          & 1.59 $\pm$ 0.39 & 0.79 $\pm$ 0.17 \\
            & \textbf{Ours}                   & 10 (Fixed)              & 0.16 & 100/100 & 0.015 $\pm$1e-4   & 0.02 $\pm$1e-4    & 0.07 $\pm$ 0.04 & 1.72 $\pm$ 0.71 &  \textbf{1.83 $\pm$ 0.71} & 0.78 $\pm$ 0.17 & 0.52 $\pm$ 0.12 \\
            & \textbf{Ours}                   & 3 (Auto)                & 1.29 & 100/100 & 0.008 $\pm$1e-4   & 0.01 $\pm$ 0.01   & 0.03 $\pm$ 0.02 & 0.33 $\pm$ 0.15 &  \textbf{0.37 $\pm$ 0.15} & 0.72 $\pm$ 0.16 & 0.52 $\pm$ 0.13 \\
            \midrule
          \multirow{5}{*}{\CircledCyan[]{4}}
            & \cite{2014bretl}\tnote{b}       & 10 (Fixed)             & --   &  93/100 & --                 & --                & --              & --              & 46.02 $\pm$ 43.88         & --              & 1.81 $\pm$ 0.31 \\
            & \cite{2020mcconachiea}\tnote{b} & 10 (Fixed)             & --   & 0/100   & --                 & --                & --              & --              & --                        & --              & --              \\
            & \cite{2024yu}\tnote{b}          & 10 (Fixed)             & --   & 100/100 & --                 & --                & 8.22 $\pm$ 3.34 & 2.61 $\pm$ 0.44 & 10.83 $\pm$ 3.37          & 2.28 $\pm$ 0.55 & 1.33 $\pm$ 0.27 \\
            & \textbf{Ours}                   & 10 (Fixed)             & 0.11 & 100/100 & 0.016 $\pm$ 0.002  & 0.02 $\pm$1e-4    & 2.76 $\pm$ 1.08 & 1.83 $\pm$ 0.42 &  \textbf{4.63 $\pm$ 1.09} & 1.60 $\pm$ 0.34 & 1.54 $\pm$ 0.33 \\
            & \textbf{Ours}                   & 5.03 $\pm$ 0.17 (Auto) & 0.32 & 100/100 & 0.023 $\pm$ 0.002  & 0.02 $\pm$ 0.02   & 1.36 $\pm$ 0.43 & 1.12 $\pm$ 0.80 &  \textbf{2.52 $\pm$ 0.93} & 1.54 $\pm$ 0.43 & 1.50 $\pm$ 0.41 \\
          \bottomrule
      \end{tabular}
      \begin{tablenotes}
        \item[a] Mean value \(\pm\) standard deviation are provided.
        \item[b] Results taken from \cite{2024yu}. ``--'' indicates data not reported in the original paper.
      \end{tablenotes}
    \end{threeparttable}
  }
  \caption{%
    \textbf{Planning performance} for tasks \CircledCyan{1}--\CircledCyan{4}, compared with state-of-the-art methods. 
    Our planner finds feasible paths for all tasks and substantially reduces total planning time. 
    DLO approximation and URDF generation overhead is minimal compared to path planning. 
    CPU benchmark differences \cite{cpu-monkey-i9-10885h-i7-10700, technical-city-i9-10885h-i7-10700, userbenchmark-i9-10885h-i7-10700} suggest these speedups would be even greater on equivalent hardware.
  }
  \label{tab:benchmark-results-planning}
  \vspace{-0.3cm}
\end{table*}
\begin{figure*}[!htbp]
  \centering
  \begin{subfigure}[t]{0.99\textwidth}
      \centering
      \includegraphics[width=\textwidth]{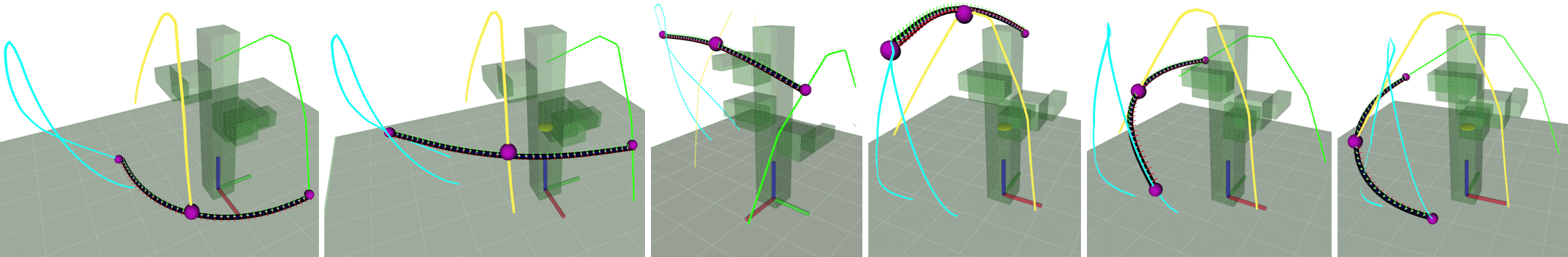}
      \caption{Task \CircledCyan{1}, snapshots at $t=\{0,6,12,17,23,31\}\,\mathrm{s}$.}
      \label{fig:benchmark-execution-snapshots-Scene1}
  \end{subfigure}
  \begin{subfigure}[t]{0.99\textwidth}
    \centering
    \includegraphics[width=\textwidth]{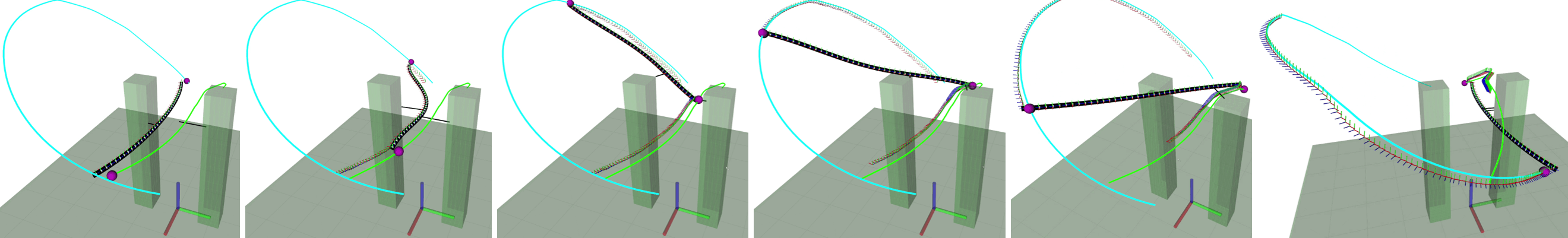}
    \caption{Task \CircledCyan{2}, snapshots at $t=\{0,4,11,20,29,44\}\,\mathrm{s}$.}
    \label{fig:benchmark-execution-snapshots-Scene2}
  \end{subfigure}
  \begin{subfigure}[t]{0.99\textwidth}
    \centering
    \includegraphics[width=\textwidth]{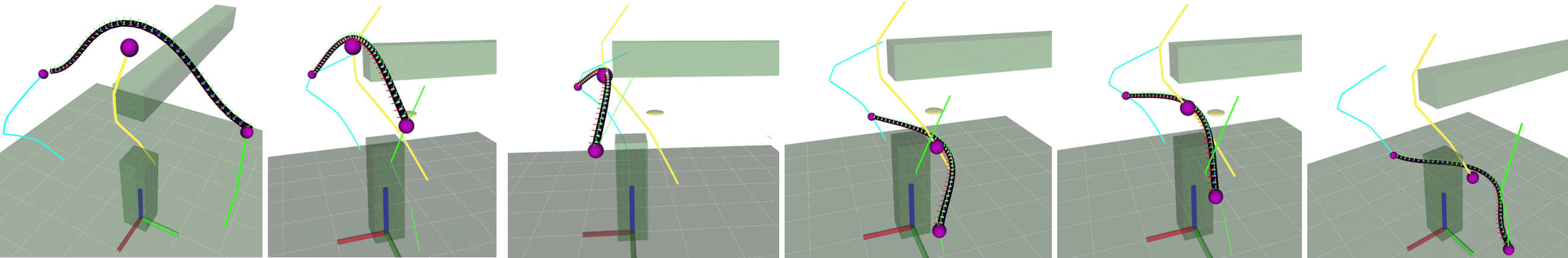}
    \caption{Task \CircledCyan{3}, snapshots at $t=\{0,6,10,13,18,25\}\,\mathrm{s}$.}
    \label{fig:benchmark-execution-snapshots-Scene3}
  \end{subfigure}
  \begin{subfigure}[t]{0.99\textwidth}
    \centering
    \includegraphics[width=\textwidth]{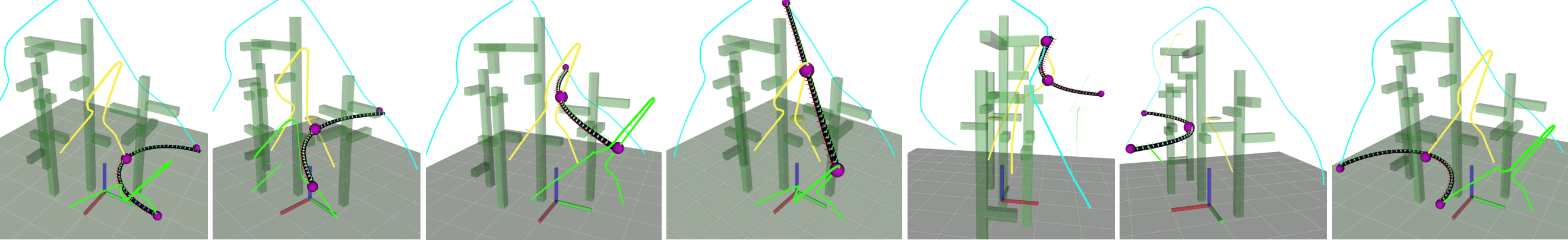}
    \caption{Task \CircledCyan{4}, snapshots at $t=\{0,10,19,27,37,49,60\}\,\mathrm{s}$.}
    \label{fig:benchmark-execution-snapshots-Scene4}
  \end{subfigure}
  \caption{Sample \textbf{execution snapshots} for tasks \CircledCyan{1}--\CircledCyan{4}. The DLO is guided from start to goal while held at its ends, avoiding collisions and overstress. Green and cyan lines depict the guidance paths; the yellow line shows the centroid of the planner’s approximated DLO.}
  \label{fig:benchmark-execution-snapshots}
  \vspace{-0.3cm}
\end{figure*}
\begin{table*}[!htbp]
  \centering
  \resizebox{\textwidth}{!}{
  \begin{threeparttable}[b]
    \begin{tabular}{@{}l l c c c c c c c c c@{}}
        \toprule
        \textbf{Task} 
          & \makecell{\textbf{Method}}
          & \makecell{\textbf{Avr. Num.}\\\textbf{of}\\\textbf{Segments}}
          & \makecell{\textbf{Final}\\\textbf{Task}\\\textbf{Error (mm)}}
          & \makecell{\textbf{Manipulation}\\\textbf{Success}\\\textbf{Rate}}
          & \makecell{\textbf{Number}\\\textbf{of}\\\textbf{Replanning}}
          & \makecell{\textbf{Execution}\\\textbf{Time}\\\textbf{(s)\tnote{a}}}
          & \makecell{\textbf{Executed}\\\textbf{Path}\\\textbf{Lenght (m)\tnote{a}}}
          & \makecell{\textbf{Control}\\\textbf{Rate}\\\textbf{(Hz)\tnote{a}}}
          & \makecell{\textbf{Min. Dist.}\\\textbf{to Obstacles}\\\textbf{(mm)\tnote{a}}}
          & \makecell{\textbf{Num. of}\\\textbf{Stress}\\\textbf{Violations}}\\
        \midrule
        \multirow{5}{*}{\CircledCyan[]{1}}
          & \cite{2014bretl}\tnote{b}       & 10 (Fixed)              & --              &  62/100        & --                & 119.8 $\pm$ 23.5  & --              & --              & --                       & --              \\
          & \cite{2020mcconachiea}\tnote{b} & 10 (Fixed)              & --              &  66/78         & 10                & --                & --              & --              & --                       & --              \\
          & \cite{2024yu}\tnote{b}          & 10 (Fixed)              & 0.10 $\pm$ 0.03 & 100/100        & 0                 & 47.9 $\pm$ 9.1    & --              & 5               & --                       & --              \\
          & \textbf{Ours}                   & 10 (Fixed)              & 0.13 $\pm$ 0.03 & 100/100        & 0                 & 54.1 $\pm$ 5.1    & 1.08 $\pm$ 0.14 & 34.4 $\pm$ 0.7  & 8.8 $\pm$ 3.7            & 0/100           \\
          & \textbf{Ours}                   & 4 (Auto)                & 0.12 $\pm$ 0.01 & 100/100        & 0                 & 51.0 $\pm$ 6.1    & 1.08 $\pm$ 0.16 & 36.3 $\pm$ 0.9  & 8.6 $\pm$ 3.3            & 0/100           \\
        \midrule
        \multirow{5}{*}{\CircledCyan[]{2}}
          & \cite{2014bretl}\tnote{b}       & 10 (Fixed)              & --              & 92/100         & --                & 76.3 $\pm$ 14.8   & --              & --              & --                        & --              \\
          & \cite{2020mcconachiea}\tnote{b} & 10 (Fixed)              & --              & 80/100         & 53                & --                & --              & --              & --                        & --              \\
          & \cite{2024yu}\tnote{b}          & 10 (Fixed)              & 0.20 $\pm$ 0.08 & 100/100        & 0                 & 41.0 $\pm$ 7.4    & --              & 5               & --                        & --              \\
          & \textbf{Ours}                   & 10 (Fixed)              & 0.08 $\pm$ 0.01 & 100/100        & 0                 & 56.6 $\pm$ 5.4    & 1.08 $\pm$ 0.08 & 36.6 $\pm$ 0.5  & 9.2 $\pm$ 2.7             & 0/100           \\
          & \textbf{Ours}                   & 2 (Auto)                & 0.13 $\pm$ 0.05 & 100/100        & 0                 & 56.3 $\pm$ 4.8    & 1.22 $\pm$ 0.07 & 35.6 $\pm$ 0.8  & 9.3 $\pm$ 3.0             & 0/100           \\
          \midrule
          \multirow{5}{*}{\CircledCyan[]{3}}
          & \cite{2014bretl}\tnote{b}       & 10 (Fixed)              & --              & 100/100        & --                & 101.5 $\pm$ 20.1  & --              & --              & --                        & --              \\
          & \cite{2020mcconachiea}\tnote{b} & 10 (Fixed)              & --              & 87/91          & 57                & --                & --              & --              & --                        & --              \\
          & \cite{2024yu}\tnote{b}          & 10 (Fixed)              & 0.44 $\pm$ 0.11 & 100/100        & 0                 & 37.5 $\pm$ 6.9    & --              & 5               & --                        & --              \\
          & \textbf{Ours}                   & 10 (Fixed)              & 0.33 $\pm$ 0.11 & 100/100        & 0                 & 34.4 $\pm$ 4.8    & 0.56 $\pm$ 0.13 & 36.1 $\pm$ 0.5  & 10.8 $\pm$ 3.4            & 0/100           \\
          & \textbf{Ours}                   & 3 (Auto)                & 0.30 $\pm$ 0.08 & 100/100        & 0                 & 38.4 $\pm$ 5.6    & 0.59 $\pm$ 0.15 & 38.5 $\pm$ 1.1  & 7.2 $\pm$ 3.6             & 0/100           \\
          \midrule
        \multirow{5}{*}{\CircledCyan[]{4}}
          & \cite{2014bretl}\tnote{b}       & 10 (Fixed)              & --              & 74/93          & --                & 125.3 $\pm$ 18.4  & --              & --              & --                       & --              \\
          & \cite{2020mcconachiea}\tnote{b} & 10 (Fixed)              & --              & 0/0            & --                & --                & --              & --              & --                       & --              \\
          & \cite{2024yu}\tnote{b}          & 10 (Fixed)              & 0.13 $\pm$ 0.11 & 100/100        & 1                 & 62.5 $\pm$ 12.6   & --              & 5               & --                       & --              \\
          & \textbf{Ours}                   & 10 (Fixed)              & 0.17 $\pm$ 0.06 & 100/100        & 0                 & 124.3 $\pm$ 32.3  & 1.86 $\pm$ 0.41 & 25.5 $\pm$ 1.5  & 4.2 $\pm$ 1.9            & 0/100           \\
          & \textbf{Ours}                   & 5.03 $\pm$ 0.17 (Auto)  & 0.21 $\pm$ 0.09 & 100/100        & 0                 & 116.3 $\pm$ 20.8  & 1.89 $\pm$ 0.40 & 25.5 $\pm$ 1.1  & 4.4 $\pm$ 1.6            & 0/100           \\
        \bottomrule
    \end{tabular}
    \begin{tablenotes}
      \item[a] Mean value \(\pm\) standard deviation are provided.
      \item[b] Results taken from \cite{2024yu}. ``--'' indicates data not reported in the original paper.
    \end{tablenotes}
  \end{threeparttable}
  }
  \caption{%
    \textbf{Execution performance} for tasks \CircledCyan{1}--\CircledCyan{4}, compared with state-of-the-art methods. 
    Our local controller successfully completes all tasks without collisions or overstress, guided by the planner’s paths. 
    Minor deviations from these paths increase total path length but maintain safe offset distances from obstacles. 
    No replanning was required, confirming the effectiveness of the initial plans.
  }
  \label{tab:benchmark-results-executions}
  \vspace{-0.2cm}
\end{table*}
\begin{figure*}[!htbp]
  \centering
  \includegraphics[width=0.99\textwidth]{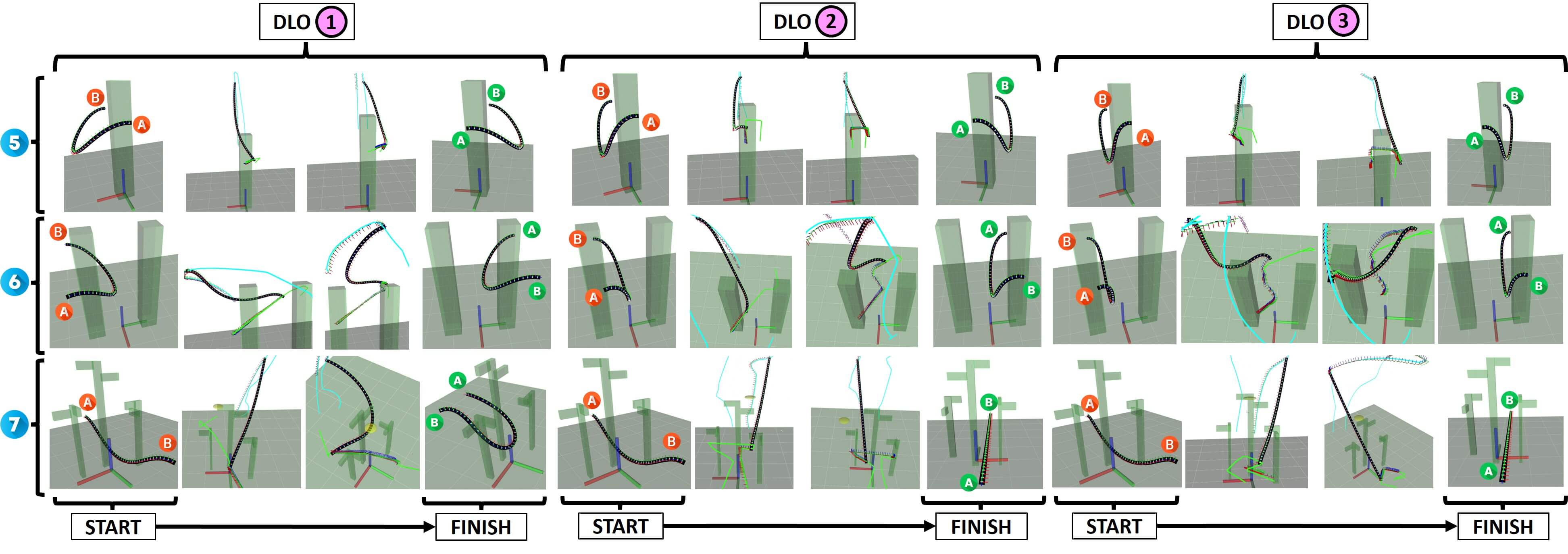}
  \caption{%
    Tasks \CircledCyan{5}--\CircledCyan{7}, used for DLOs \CircledMagenta{1}--\CircledMagenta{3}. 
    Each DLO travels from the initial pose (left) to the target pose (right), held at its ends. 
    \CircledCyan{5} is a simple block-bypass scenario; 
    \CircledCyan{6} requires flipping the DLO; 
    \CircledCyan{7} passes a narrow gap.
  }
  \label{fig:different-stiffness-tasks}
  \vspace{-0.3cm}
\end{figure*}

Table~\ref{tab:benchmark-results-planning} summarizes the planning results. Our method achieves a 100\% success rate, matching \cite{2024yu}. In task \CircledCyan{4}, three out of 100 trials employed six links rather than five, yielding an average of $5.03 \pm 0.17$. While our initial path lengths are slightly shorter than those in \cite{2024yu}, the final smoothed lengths are comparable, regardless of how many segments the planner uses.
Our primary advantage is reduced planning time, showing $54\%$, $67\%$, $69\%$, and $57\%$ faster performance (fixed 10 segments) and $83\%$, $94\%$, $94\%$, and $76\%$ faster (automatic segment selection) for tasks \CircledCyan{1}--\CircledCyan{4}, respectively. DLO approximation and URDF generation overhead is negligible compared to total planning time. While using fewer segments yields larger approximation errors, it does not reduce success rates, thanks to threshold checks in Section~\ref{sec:global_planner}.

Fig.~\ref{fig:benchmark-execution-snapshots} shows sample execution snapshots, and Table~\ref{tab:benchmark-results-executions} summarizes average execution results for each task. All trials succeed without collisions or overstress. For tasks \CircledCyan{1}--\CircledCyan{3}, we maintain an offset distance $d^\mathrm{ofst}=5\,\mathrm{mm}$; in \CircledCyan{4}, the minimum obstacle distance averages about $4\,\mathrm{mm}$ due to denser clutter. Task \CircledCyan{4} also takes longer to execute than \cite{2024yu} because the safety constraints slow the motion to avoid collisions from all sides. Nonetheless, no deadlocks occurred, and no replanning was triggered, indicating the planner’s paths were sufficient. Although additional collision constraints reduce the control rate in the cluttered environment of \CircledCyan{4}, bounding the number of active constraints (e.g., ignoring distant obstacles) can improve the controller rate (\S\ref{subsubsec:local-controller-runtime-performance}). As expected, the final executed trajectories exceed the planned paths due to safety-related deviations, yet final task errors remain sub-millimeter, comparable to \cite{2024yu}.

\subsubsection{Tests with Different Stiffness Parameters}
\label{subsubsec:tests-with-different-stiffness}
\begin{table}[!hbtp]
  \centering
    \begin{threeparttable}[b]
      \begin{tabular}{@{}c@{}c@{}c c c@{}}
          \toprule
          \makecell{\textbf{DLO}}
            & \makecell{\textbf{Type}}
            & \makecell{\textbf{Qualitative}\\\textbf{Stiffness\tnote{a}}}
            & \makecell{\textbf{Young's}\\\textbf{Modulus}\\\textbf{(MPa)}}
            & \makecell{\textbf{Shear}\\\textbf{Modulus}\\\textbf{(MPa)}}\\
          \midrule
          \CircledMagenta[]{0}
            & Tent Pole       & 5/5            & 30000      & 10000    \\
          \CircledMagenta[]{1}
            & TPU Rubber      & 3/5            & 10         & 100      \\
          \CircledMagenta[]{2}
            & Nylon Rope      & 2/5            & 3          & 100      \\
          \CircledMagenta[]{3}
            & Hemp Rope       & 1/5            & 1          & 100      \\
          \bottomrule
      \end{tabular}
      \begin{tablenotes}
        \item[a] Estimated by human judgment.
      \end{tablenotes}
    \end{threeparttable}
  \caption{%
    Parameters for different DLO stiffnesses. 
    \CircledMagenta{0} is tent-pole stiffness from prior experiments, 
    while \CircledMagenta{1}--\CircledMagenta{3} mimic TPU/rubber, nylon rope, and hemp rope.
  }
  \label{tab:stiffness-parameters}
\end{table}
\begin{figure}[!hbtp]
  \centering
  \begin{minipage}[c]{0.60\columnwidth}
    \includegraphics[width=0.99\textwidth]{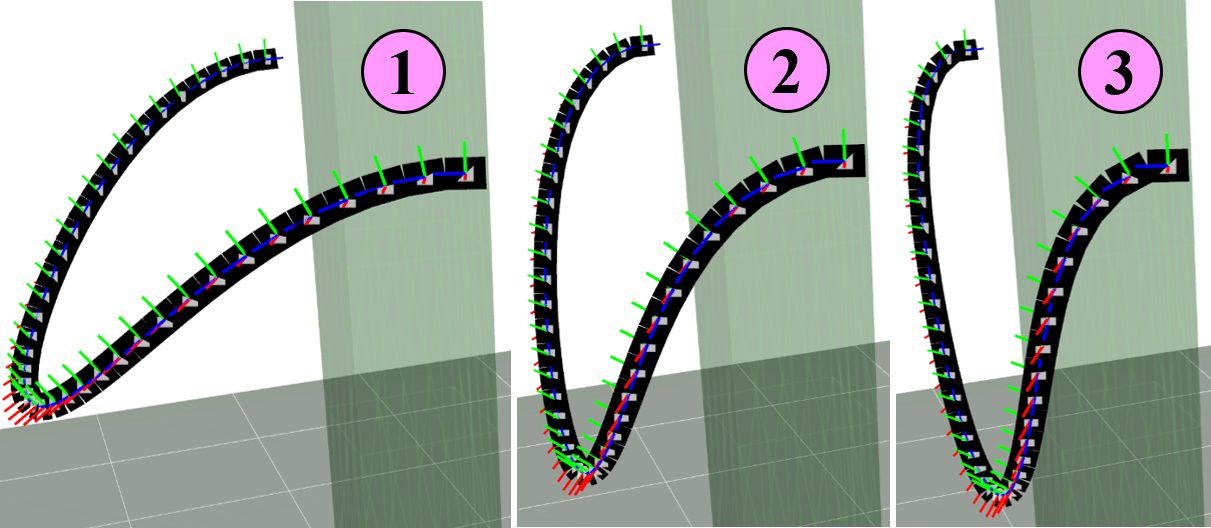}
  \end{minipage}
  \begin{minipage}[c]{0.35\columnwidth}
    \caption{%
    Simulated DLOs \CircledMagenta{1}--\CircledMagenta{3} under gravity, with each end fixed at the same position. 
    Lower stiffness values produce more pronounced sag.
    }
    \label{fig:stiffness-parameters}
  \end{minipage}
\end{figure}
Previously, all experiments used a single set of stiffness parameters resembling a tent pole (\CircledMagenta{0}). To evaluate robustness, we tested lower-stiffness DLOs (\CircledMagenta{1}, \CircledMagenta{2}, \CircledMagenta{3}) approximating TPU rubber, nylon rope, and hemp rope, as listed in Table~\ref{tab:stiffness-parameters}. Fig.~\ref{fig:stiffness-parameters} shows these DLOs in equilibrium under gravity with theirs ends are fixed.
We then designed three tasks (\CircledCyan{5}--\CircledCyan{7}), inspired by \cite{2024yu} (Fig.~\ref{fig:different-stiffness-tasks}), each featuring different initial and final DLO states. Task \CircledCyan{5} involves bypassing a block; Task \CircledCyan{6} requires flipping the DLO between two blocks (similar to Task \CircledCyan{2}); and Task \CircledCyan{7} navigates a narrow gap. Although \CircledCyan{7} is the biggest challenge for DLO \CircledMagenta{1}, it demands less bending for \CircledMagenta{2} and \CircledMagenta{3}, which can remain in tension and complete the task more easily.

\begin{table}[!hbtp]
  \centering
  \begin{threeparttable}[b]
    \begin{tabular}{@{}l c c c c@{}}
        \toprule
          \makecell{\textbf{DLO}}
          & \makecell{\textbf{Task}}
          & \makecell{\textbf{Success}\\\textbf{Rate}}
          & \makecell{\textbf{Replan}\\\textbf{Need}\\\textbf{Rate}}
          & \makecell{\textbf{Total}\\\textbf{Num. of}\\\textbf{Replans}}\\
        \midrule
        \multirow{3}{*}{\CircledMagenta{1}}
          & \CircledCyan{5} & 20/20 &  2/20 & 2  \\
          & \CircledCyan{6} & 20/20 & 16/20 & 24  \\
          & \CircledCyan{7} & 20/20 &  8/20 & 9  \\
        \midrule
        \multirow{3}{*}{\CircledMagenta{2}}
          & \CircledCyan{5} & 20/20 &  5/20 & 5  \\
          & \CircledCyan{6} & 20/20 & 17/20 & 24  \\
          & \CircledCyan{7} & 20/20 &  3/20 & 3  \\
        \midrule
        \multirow{3}{*}{\CircledMagenta{3}}
          & \CircledCyan{5} & 20/20 &  5/20 & 9  \\
          & \CircledCyan{6} & 20/20 & 20/20 & 27  \\
          & \CircledCyan{7} & 20/20 &  0/20 & 0  \\
        \bottomrule
    \end{tabular}
  \end{threeparttable}
  \caption{%
    Success and replanning rates for tasks \CircledCyan{5}--\CircledCyan{7} with DLOs \CircledMagenta{1}--\CircledMagenta{3}. 
    Lower stiffness generally increases replan frequency, except in configurations allowing near-tension DLO states.
  }
  \label{tab:different-stiffness-results-executions}
  \vspace{-0.1cm}
\end{table}

\begin{figure*}[!htbp]
  \centering
  \includegraphics[width=0.99\textwidth]{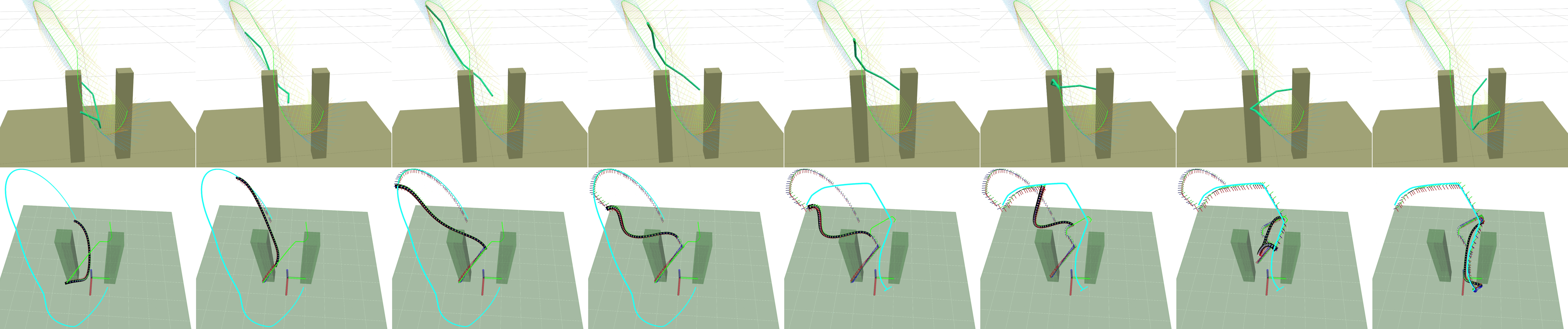}
  \caption{%
    Task \CircledCyan{6} with DLO \CircledMagenta{2}, highlighting the need for replanning. 
  The top row shows the planner’s simplified, gravity-free model; 
  the bottom row shows real execution at $t=\{0,16,24,64,72,84,90,110\}\,\mathrm{s}$ (left to right). 
  For the first three snapshots, the execution closely follows the plan. 
  At $t=64\,\mathrm{s}$, however, the path directs the DLO into the obstacle, leaving the controller stuck. 
  Replanning is triggered, and by $t=72\,\mathrm{s}$, a new path successfully navigates around the obstacle, 
  compensating for unmodeled gravity and low stiffness.
  }
  \label{fig:different-stiffness-replanning-example}
  \vspace{-0.3cm}
\end{figure*}

Each of these nine cases was run 20 times (Table~\ref{tab:different-stiffness-results-executions}). While all trials succeeded, lower stiffness generally led to more replanning. For example, Task \CircledCyan{5} required 2, 5, and then 9 total replans for \CircledMagenta{1}, \CircledMagenta{2}, and \CircledMagenta{3}, respectively, and Task \CircledCyan{6} triggered replanning for 16, 17, and finally all 20 trials. An exception occurs when the DLO stays in tension (e.g., Task \CircledCyan{7} with \CircledMagenta{2} or \CircledMagenta{3}), requiring few or no replans.

Fig.~\ref{fig:different-stiffness-replanning-example} illustrates a Task \CircledCyan{6} run with DLO \CircledMagenta{2}, highlighting why replanning occurs. The planner assumes fully controllable, gravity-free joints, so the actual DLO—affected by gravity—sags differently. Although it follows the planned path for the first three snapshots, at $t=64\,\mathrm{s}$ the path directs the DLO into the obstacle. The local controller deviates to avoid collision but eventually gets stuck, prompting a replan. By $t=72\,\mathrm{s}$, a newly generated plan circumvents the obstacle, compensating for the unmodeled gravity and low stiffness.
Such plan–execution mismatches are more frequent for lower-stiffness DLOs, especially when they are in a relaxed or highly bent state. Nevertheless, once replanning is enabled, all feasible tasks eventually succeed. Across many trials, we observed no scenario in which the system failed indefinitely.

\subsection{Real-world Tasks}
\subsubsection{Hardware Setup}
\label{subsubsec:hardware_setup}
\begin{figure}[!hbtp]
  \centering
  \includegraphics[width=1.0\columnwidth]{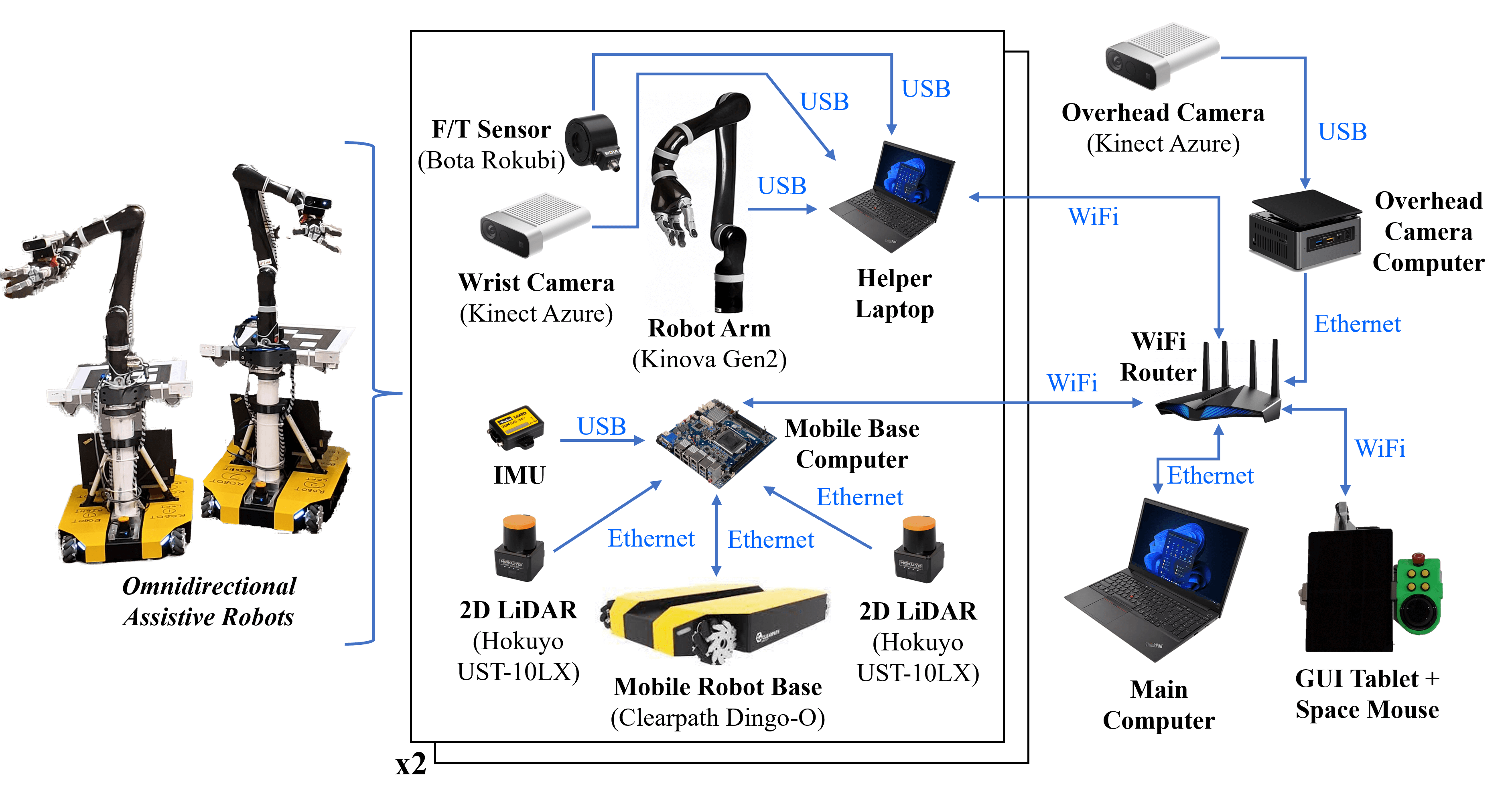}
  \caption{%
    \textbf{Hardware architecture} for the real-world experiments. 
    Each mobile manipulator consists of a 3-DoF omnidirectional base and a 6-DoF arm
    with onboard LiDARs, IMU, and an F/T sensor. 
    An overhead RGB-D camera, Wi-Fi router, GUI tablet, 
    and main computer complete the setup for ROS communications.
  }
  \label{fig:hardware-architecture}
  \vspace{-0.3cm}
\end{figure}

Fig.~\ref{fig:hardware-architecture} illustrates our hardware architecture, which comprises two mobile manipulators. Each manipulator has a 3-DoF omnidirectional base~\cite{dingo} and a 6-DoF robotic arm~\cite{kinova}. A sensor fusion suite~\cite{ros_robot_localization} combines data from fiducial markers, overhead camera, IMU~\cite{imu}, and odometry for localization. Two LiDARs~\cite{hokuyo} on the mobile bases detect obstacles, and an F/T sensor~\cite{rokubi} at each wrist measures stress. A Wi-Fi router, GUI tablet, and a main computer manage ROS communications and user interactions.
Each end effector accepts spatial velocity commands (linear and angular), distributing them between the arm and base via an inbuilt redundancy resolution. For smaller or nearby motions, the arm moves independently; for larger or distant targets, the omnidirectional base is prioritized. Intermediate scenarios blend both arm and base commands smoothly. Each base also uses local obstacle avoidance to handle objects on the floor, ensuring overall safety. We omit further implementation details for brevity.

\subsubsection{Tent-Building Scene for Real Robots}
\begin{figure}[!htbp]
  \centering
  \includegraphics[width=0.8\columnwidth]{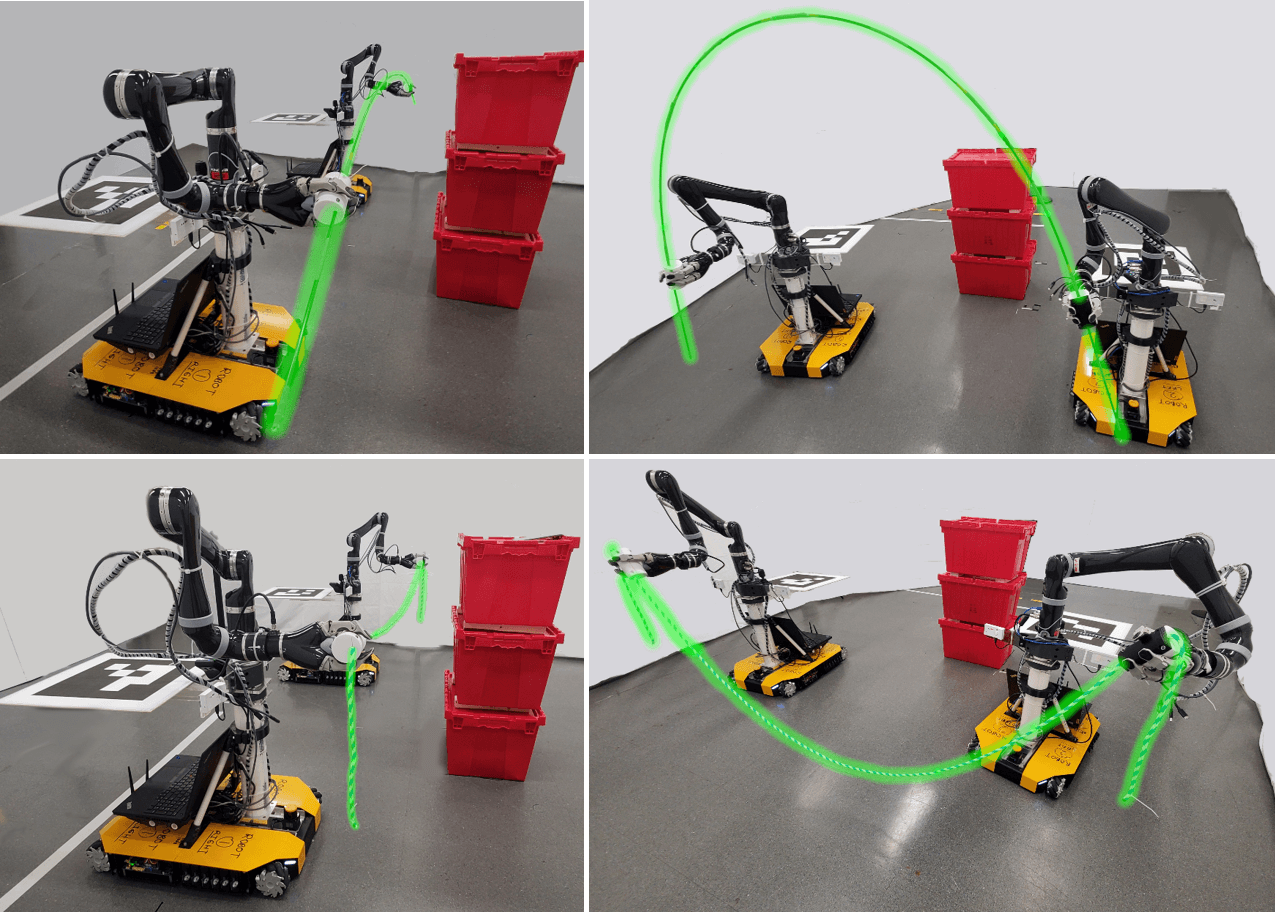}
  \caption{%
    Representative \textbf{initial} (left) and \textbf{target} (right) \textbf{configurations} 
  for the real-robot tent-building tasks. 
  A stiff tent pole (top) 
  and a soft rope (bottom) 
  are both carried over an obstacle made from stacked red tote boxes to a designated goal.
  The DLOs are highlighted in green for easier visibility.
  }
  \label{fig:real-robot-task-descriptions}
  \vspace{-0.3cm}
\end{figure}
We adapt the tent-building scenario from earlier sections to a real-world $(3.7\times5.0)\,\mathrm{m}$ workspace, 
with the origin at one corner. 
A stack of three tote boxes measuring $(0.56\times0.4\times1.0)\,\mathrm{m}$ 
serves as an obstacle, positioned at $(1.9,\,2.2)\,\mathrm{m}$ in the XY-plane. 
Two DLOs are tested: 
a tent pole of length $3.353\,\mathrm{m}$, thickness $7\,\mathrm{mm}$, density $1793\,\mathrm{kg/m^3}$, and Young’s/shear moduli $30\,\mathrm{GPa}/10\,\mathrm{GPa}$; 
and a rope of equal length, but $9\,\mathrm{mm}$ thickness, $0.086\,\mathrm{kg}$ mass, and Young’s/shear moduli $3\,\mathrm{MPa}/1\,\mathrm{MPa}$. 
Both DLOs begin behind the obstacle at varying distances $d$ and heights $h$, 
grasped at segments 5 and 34 ($\sim46\,\mathrm{cm}$ away from each tip), 
and are discretized into 40 segments for the PBD simulation.
For the tent pole, initial shapes include straight, U-shaped (bent upward), or N-shaped (bent downward). 
For the rope, the initial state is either straight under tension or sagging into an M-shape under gravity (Fig.~\ref{fig:real-robot-task-descriptions}). 
Each DLO must be carried over the obstacle to two virtual grommets at $(0.9,\,3.8)\,\mathrm{m}$ and $(2.9,\,3.8)\,\mathrm{m}$, 
with final orientations of $15^\circ$ (tent pole) or $0^\circ$ (rope) relative to $+z$. 

\begin{figure}[!t]
  \centering
  \includegraphics[width=1.0\columnwidth]{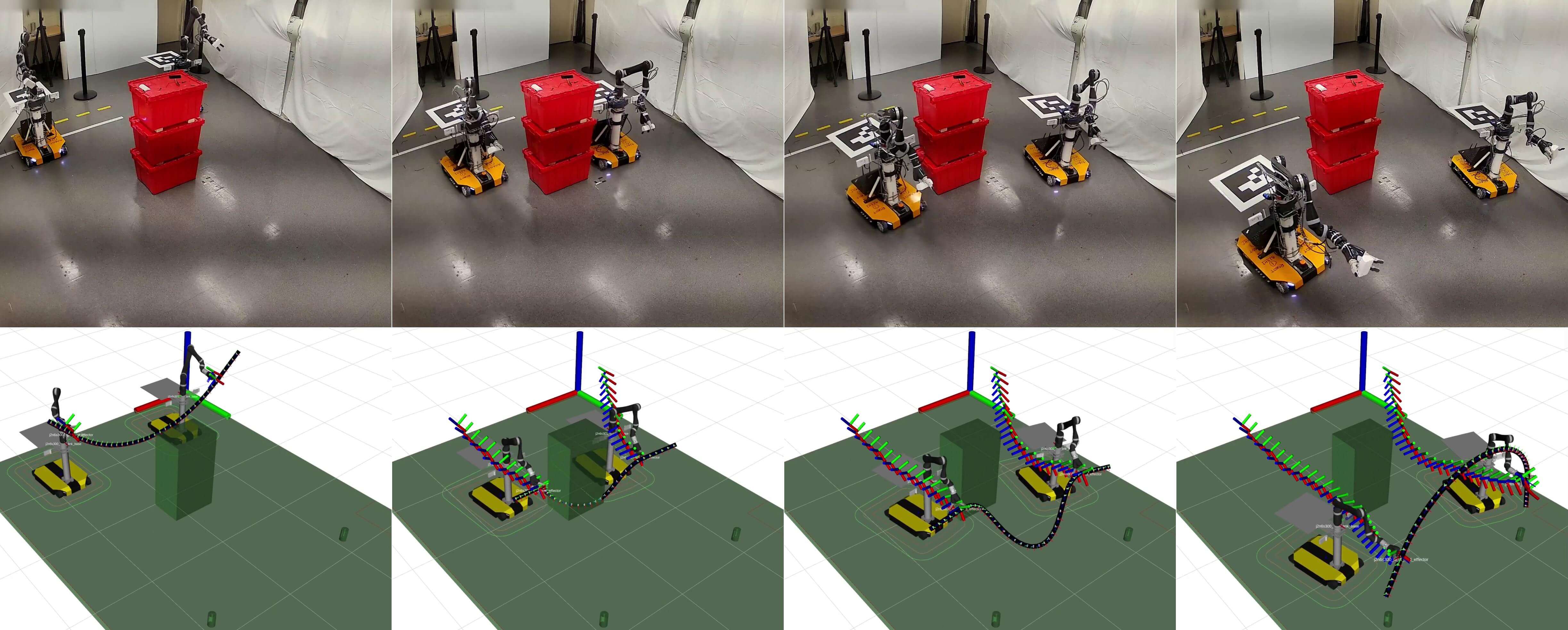}
  \caption{%
    Snapshots from the real-robot experiment using \emph{only the nominal controller (all safety features disabled)}. 
    Although the tent pole tips reach their final poses (rightmost image), 
    the path crosses obstacles (second image) and can cause overstress (third image). 
    To prevent hardware damage in this demonstration, the tent pole is simulated in RViz (bottom row) and not physically present, 
    with its state determined by the actual robots’ end-effector localizations.
  }
  \label{fig:real-robot-issues-without-safety}
  \vspace{-0.3cm}
\end{figure}
\begin{figure}[!b]
  \centering
  \includegraphics[width=1.0\columnwidth]{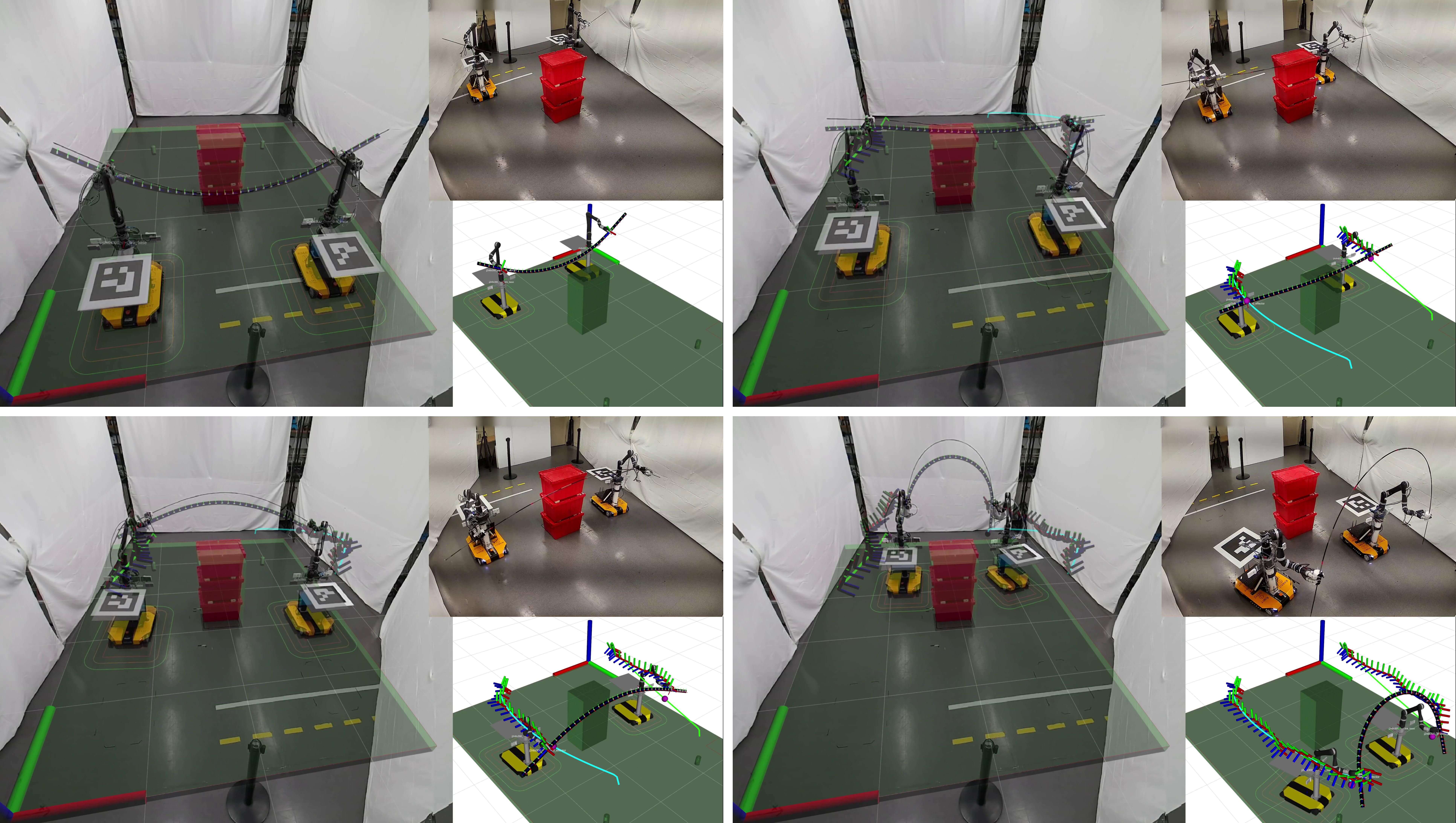}
  \caption{%
    (Left to right, top to bottom) Snapshots of the real-robot execution with a \textbf{stiff tent pole}, 
    safety features, and planning enabled at $t=\{0,21,32,72\}\,\mathrm{s}$. 
    Each frame includes three views: 
    \emph{(top right)} a real camera image, 
    \emph{(bottom right)} the corresponding RViz angle, 
    and \emph{(left)} a superimposed real-plus-RViz image from another viewpoint, highlighting the close match between the physical and simulated DLO. 
    No additional shape-matching is performed during execution, yet the task finishes without collisions or overstress.
  }
  \label{fig:real-robot-stiff-success}
\end{figure}

\subsubsection{Case 1: Virtual Tent Pole with Real Robots}
To verify the nominal controller, hardware functionality, and obstacle-scene setup, we first run an experiment using a \emph{virtual} tent pole. The actual robots move as though grasping a physical DLO, but no real tent pole is attached. Instead, the DLO state is simulated based on real end-effector positions. As shown in Fig.~\ref{fig:real-robot-issues-without-safety}, the nominal controller moves the virtual tent pole from an initially U-shaped configuration (tips oriented upward) to an N-shaped configuration (tips oriented downward), demonstrating control of both orientation and position. However, the resulting path crosses obstacles and permits high-stress states, highlighting the need for safety features described in this paper.

\subsubsection{Case 2: Real Stiff Tent Pole with Real Robots}
\label{subsubsec:Case-2-Real-Stiff-Tent-Pole-with-Real-Robots}
Next, we attach an actual stiff tent pole to the robots and repeat the previous task—from a concave-up to a concave-down configuration—this time with safety and planning features enabled. The planner pipeline, using a three-link approximation, generated a guidance path in $0.72\,\mathrm{s}$, with the per-robot paths shown in Fig.~\ref{fig:real-robot-stiff-success}.
During execution, the robots track the paths locally, avoiding the obstacle by moving in a relatively straight line over it at $t=21\,\mathrm{s}$, then gradually bending the stiff rod downward starting around $t=32\,\mathrm{s}$ until the task concludes. For most of the motion, neither obstacle nor overstress constraints activate, so there is little deviation from the guidance path. After $t=32\,\mathrm{s}$, the nominal input transitions from path tracking to tip control, diverging from the original path to reach the final poses.
Superimposed real-plus-RViz images in Fig.~\ref{fig:real-robot-stiff-success} show a close correspondence between the physical DLO and the simulated PBD model, achieved with no additional shape adjustments after setting the material parameters (Section~\ref{subsec:param_identification}). In other words, execution proceeds in an effectively open-loop manner with respect to the DLO’s exact shape, relying on accurate robot poses, grasp points, and well-tuned parameters. Some discrepancies appear in the final heavily bent configuration, mainly due to accumulated inaccuracies in robot localization and parameter estimation. Nonetheless, the task completes successfully with no collisions or overstress, demonstrating the robustness of our method. Further discussion of these discrepancies is provided in the failure cases section.

\subsubsection{Case 3: Real Rope with Real Robots}
\begin{figure}[!b]
  \centering
  \includegraphics[width=1.0\columnwidth]{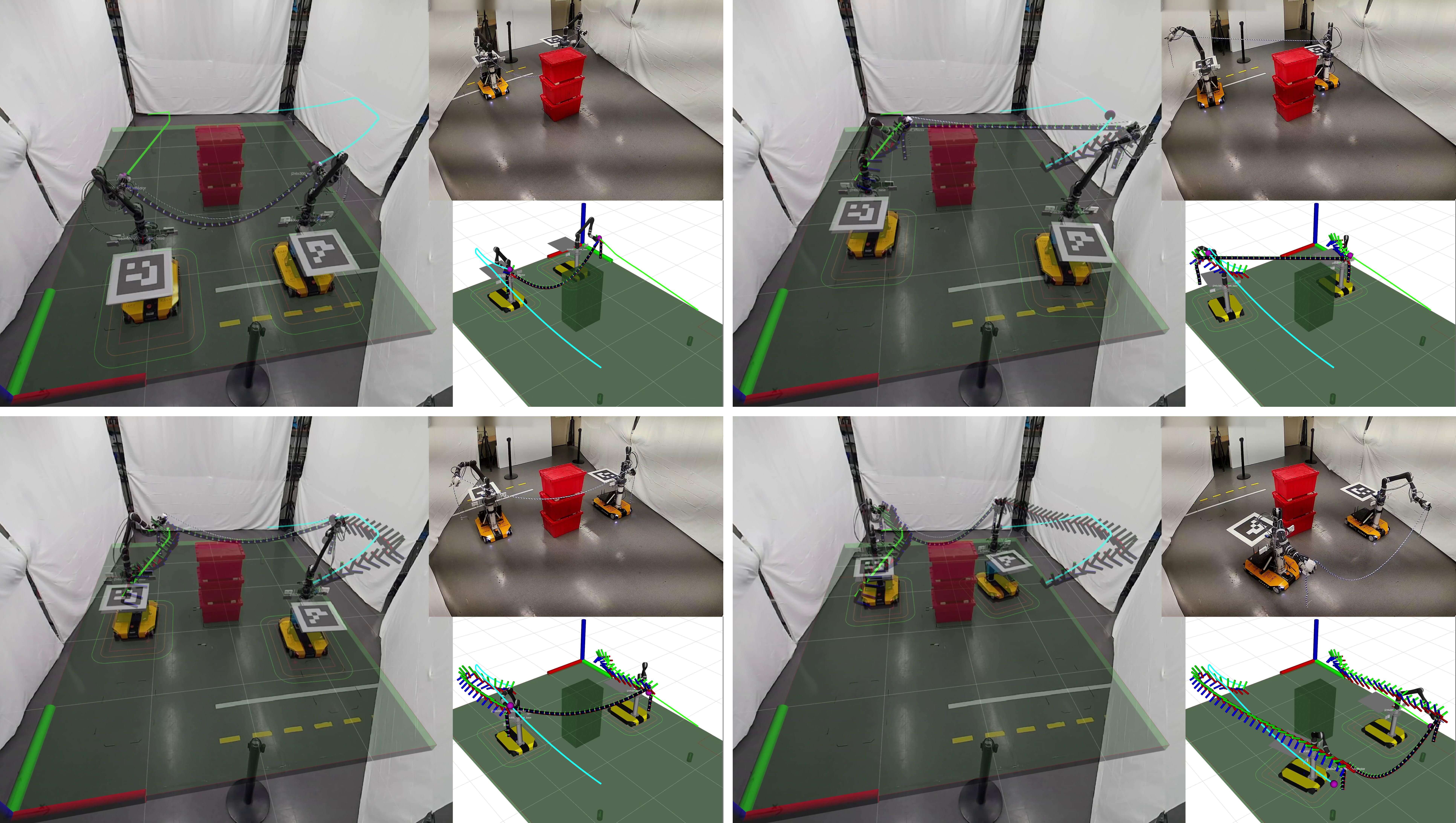}
  \caption{%
    (Left to right, top to bottom) Snapshots of the real-robot execution with a \textbf{rope}, 
    safety features, and planning enabled at $t=\{0,95,112,145\}\,\mathrm{s}$. 
    The rope is brought to tension to avoid the obstacle while deviating from the guidance path to prevent overstress. Towards the end, the nominal control shifts from path tracking to tip control with safety, again, yielding to deviations from the guidance path. Overall the task completes successfully without any safety violations.
  }
  \label{fig:real-robot-rope-success}
\end{figure}
Finally, we attach a rope to the robots. Compared to the tent pole’s smoother U-shape, this rope starts and ends in more pronounced M-shaped configurations, causing the planner to use a seven-link approximation. Generating the guidance path in this scenario took $2.65\,\mathrm{s}$, with per-robot paths shown in Fig.~\ref{fig:real-robot-rope-success}.
During execution, the rope is held under tension while following the planned path around the obstacle. Around $t=95\,\mathrm{s}$, the robot following the cyan path in Fig.~\ref{fig:real-robot-rope-success} visibly diverges from the plan to avoid overstress, triggered by overlapping constraints from obstacle avoidance, stress prevention, and nominal control. Although these competing constraints slow the robots, the local controller does not become stuck, and no replanning is required.
Once the obstacle is cleared, actual robot poses gradually converge to the planned path. Similar to the tent-pole case, after $t=112\,\mathrm{s}$ the nominal input transitions from path tracking to tip control, diverging further from the original path to achieve the final poses. 
Despite minor discrepancies between the actual and simulated DLO states—caused by modeling and localization inaccuracies—the task completes successfully with no failures, underscoring the robustness of the method.

\subsubsection{Overall Real-Robot Experiment Performance}
We conducted 10 trials each for the tent pole and the rope, setting the initial position at $(d,h)=(0.6,0.4)\,\mathrm{m}$ behind the obstacle. Table~\ref{tab:real-robot-experiments-planning-performance} summarizes the planning results, showing that all planning requests successfully generated a guidance path. The tent pole required only a three-link approximation, while the rope needed seven segments to capture its more pronounced M-shaped profile. Although this higher segment count lowered the rope approximation error to $3.96\,\mathrm{cm}$ (versus $5.34\,\mathrm{cm}$ for the tent pole), it roughly doubled the planning time from $0.73\,\mathrm{s}$ to $1.47\,\mathrm{s}$. Both times are short enough to be practical in real-time applications. After smoothing, path lengths averaged $2.44\,\mathrm{m}$ for the tent pole and $2.76\,\mathrm{m}$ for the rope, reflecting efficient obstacle avoidance optimization.
Table~\ref{tab:real-robot-experiments-execution-performance} details execution performance. The local controller ran at $26.7\,\mathrm{Hz}$ for the tent pole and $28.9\,\mathrm{Hz}$ for the rope—slower than simulation due to additional data traffic, processing, and logging in the real-robot system, but still real-time capable. All trials completed with final errors under $10\,\mathrm{cm}$. However, rope experiments lasted about $30\,\mathrm{s}$ longer than tent-pole experiments, averaging $165.6\,\mathrm{s}$. As seen in the sample rope case, conflicting constraints (e.g., obstacle avoidance vs.\ overstress avoidance) slowed the robots and introduced extra maneuvering. This is also shown by the number of trials requiring replanning: while no tent-pole experiments triggered replanning, four out of ten rope trials did—consistent with observations in Section~\ref{subsubsec:tests-with-different-stiffness}.

\begin{table}[!htbp]
    \centering
    \begin{subtable}{1\columnwidth}
        \centering
        \resizebox{\columnwidth}{!}{
        \begin{threeparttable}[b]
            \begin{tabular}{@{}l c c c c c@{}}
                \toprule
                    \makecell{\textbf{DLO}\\\textbf{Type}}
                    & \makecell{\textbf{Num}\\\textbf{of}\\\textbf{Segs}}
                    & \makecell{\textbf{Approx}\\\textbf{Error}\\\textbf{(cm)\tnote{a}}}
                    & \makecell{\textbf{Planning}\\\textbf{Time}\\\textbf{(s)\tnote{a}\tnote{a}}}
                    & \makecell{\textbf{Path}\\\textbf{Length}\\\textbf{(m)\tnote{a}}}
                    & \makecell{\textbf{Success}\\\textbf{Rate}}\\
                \midrule
                Tent Pole   & 3             & 5.34 $\pm$ 0.10  & 0.73$\pm$0.10 & 2.44$\pm$0.19 & 10/10\\ 
                Rope        & 7             & 3.96 $\pm$ 1.88  & 1.47$\pm$0.68 & 2.76$\pm$0.31 & 10/10\\
                \bottomrule
            \end{tabular}
            \begin{tablenotes}
                \item[a] Mean value \(\pm\) standard deviation are provided.
            \end{tablenotes}
        \end{threeparttable}
        }
        \caption{Planning performance metrics for real-robot experiments.}
        \label{tab:real-robot-experiments-planning-performance}
    \end{subtable}
    \\
    \vspace{0.1cm}
    \begin{subtable}{1\columnwidth}
        \centering
        \resizebox{\columnwidth}{!}{
        \begin{threeparttable}[b]
            \begin{tabular}{@{}l c c c c c c c@{}}
                \toprule
                    \makecell{\textbf{DLO}\\\textbf{Type}}
                    & \makecell{\textbf{Control}\\\textbf{Rate}\\\textbf{(Hz)\tnote{a}}}
                    & \makecell{\textbf{Replan}\\\textbf{Need}\\\textbf{Rate}}
                    & \makecell{\textbf{Task}\\\textbf{Error}\\\textbf{(cm)\tnote{a}}}
                    & \makecell{\textbf{Execution}\\\textbf{Time}\\\textbf{(s)\tnote{a}}}
                    & \makecell{\textbf{Success}\\\textbf{Rate}}\\
                \midrule
                Tent Pole  & 26.7$\pm$2.5 & 0/10            & 9.23$\pm$3.21    & 135.4 $\pm$ 24.4 & 10/10        \\
                Rope       & 28.9$\pm$2.2 & 4/10            & 7.53$\pm$2.93    & 165.6 $\pm$ 35.7 & 10/10         \\
                \bottomrule
            \end{tabular}
            \begin{tablenotes}
                \item[a] Mean value \(\pm\) standard deviation are provided.
            \end{tablenotes}
        \end{threeparttable}
        }
        \caption{Execution performance metrics for real-robot experiments.}
        \label{tab:real-robot-experiments-execution-performance}
    \end{subtable}
  
    \caption{Overall performance summary of real-robot experiments.}
    \label{tab:real-robot-experiment-performance}
    \vspace{-0.3cm}
\end{table}

\subsubsection{Failure Cases}
\begin{figure}[!hbtp]
  \centering
  \includegraphics[width=1.0\columnwidth]{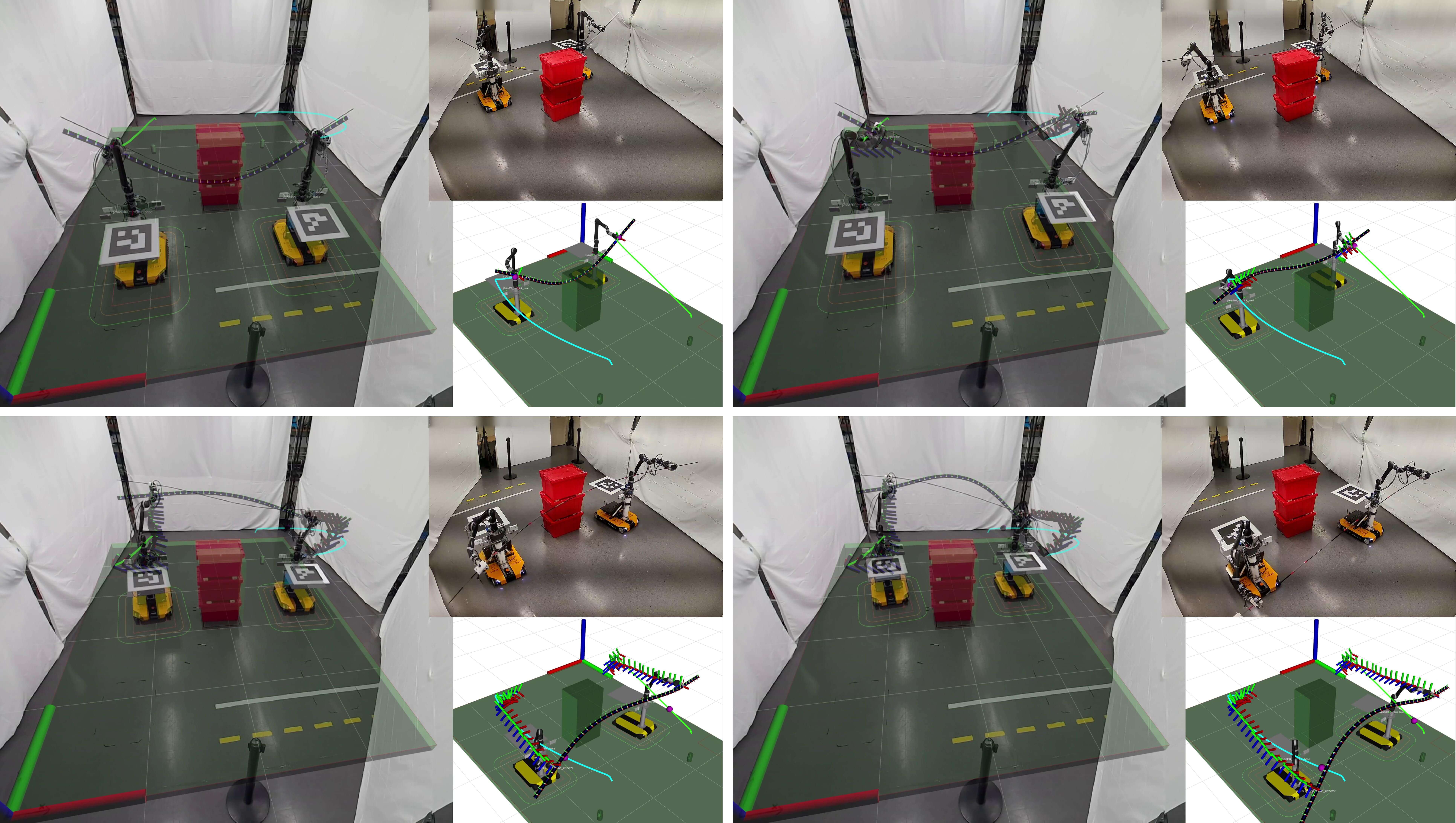}
  \caption{%
    \textbf{Failure example} during real-robot execution with a \textbf{stiff tent pole}, \textbf{caused by violating the rigid grasping assumption}. The end effector on the cyan path applies a deliberately loose grip. Snapshots at $t=\{0,44,67,76\}\,\mathrm{s}$ (left to right, top to bottom) show that although the obstacle is bypassed successfully, slippage occurs around $t=67\,\mathrm{s}$ as the end effectors move closer for task completion. The task fails as the local controller assumes rigid grasping and lacks closed-loop feedback on the actual DLO state.
  }
  \label{fig:real-robot-stiff-fail}
  \vspace{-0.3cm}
\end{figure}

\begin{figure}[!hbtp]
  \centering
  \begin{minipage}[c]{0.60\columnwidth}
    \includegraphics[width=0.99\textwidth]{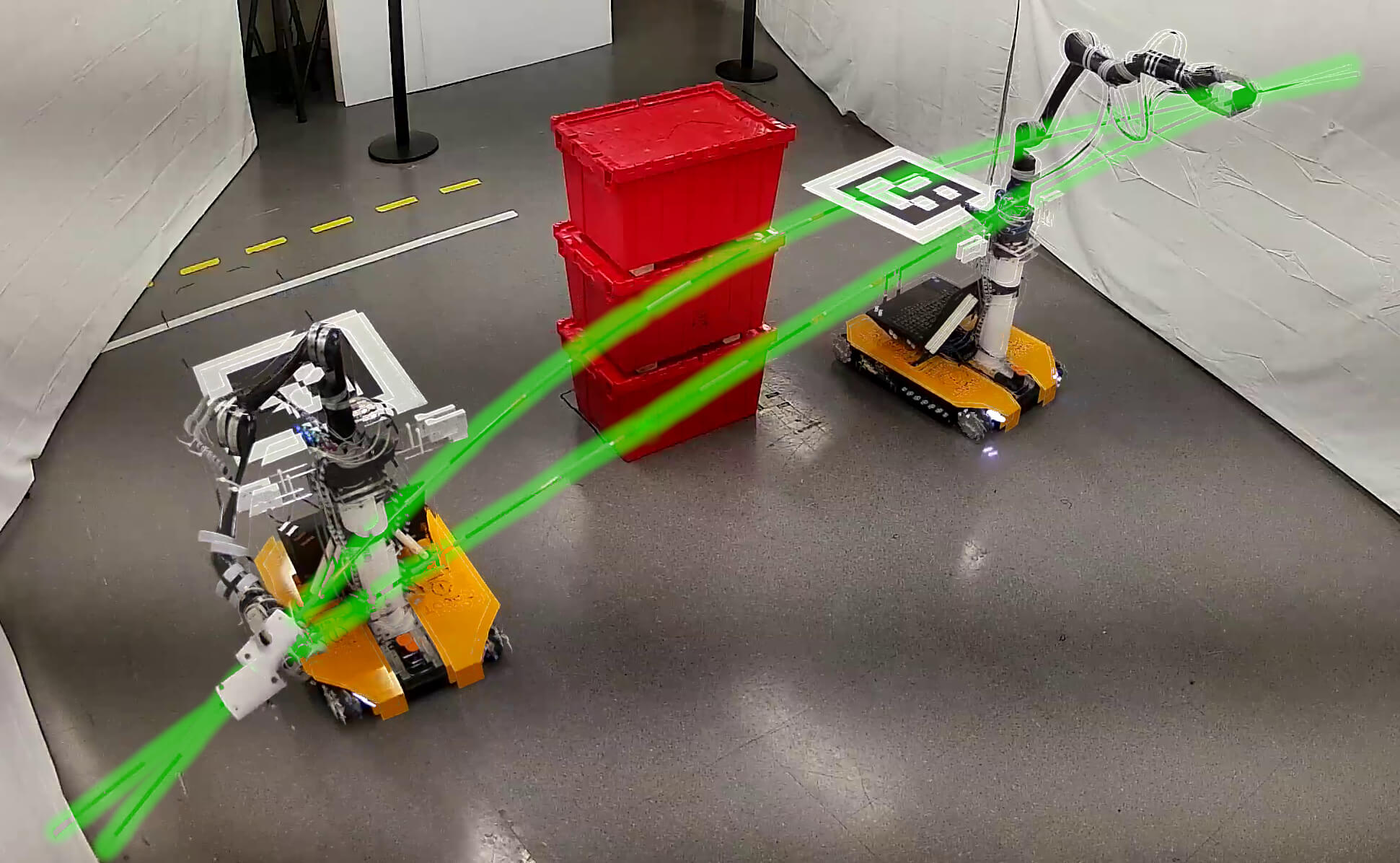}
  \end{minipage}
  \begin{minipage}[c]{0.35\columnwidth}
    \caption{%
    Closer view of the \textbf{tent pole slippage} at $t=67,\mathrm{s}$, showing before and after states with overlayed images. The DLO is highlighted in green for better visibility.
    }
    \label{fig:real-robot-stiff-grasping-slipped}
  \end{minipage}
  \vspace{-0.3cm}
\end{figure}

The importance of the assumptions in the proposed method (see Section~\ref{sec:problem_statement}) becomes evident in the following failure scenarios.
\paragraph{Violation of Rigid Grasping Assumption}
We repeated the tent-pole experiment with safety and planning enabled but intentionally loosened one robot’s end-effector grasp. As shown in Fig.~\ref{fig:real-robot-stiff-fail}, the robots successfully bypassed the obstacle from $t=44\,\mathrm{s}$ to $t=60\,\mathrm{s}$ and transitioned the pole from concave-up to slightly concave-down. However, after $t=60\,\mathrm{s}$—when the nominal controller shifted from path tracking to tip control—the end effectors moved closer to form an N-shape, increasing lateral force at the grasp points. At $t=67\,\mathrm{s}$, the loose end effector slipped, releasing the pole (Fig.~\ref{fig:real-robot-stiff-grasping-slipped}). Past that moment, the real DLO state diverged significantly from the simulation, which assumes rigid grasping (final snapshot at $t=76\,\mathrm{s}$).
This failure underscores the critical role of the rigid grasping assumption. Although relaxing stress limits in the controller might mitigate slippage by lowering lateral force, overly lenient constraints can impede task completion in scenarios demanding higher bending forces. Thus, rigid gripping or a closed-loop system that tracks real and simulated DLO states is necessary to avoid such failures.


\begin{figure}[!htbp]
    \centering
    \begin{subfigure}[h]{0.8\columnwidth}
        \centering
        \includegraphics[width=1.0\linewidth]{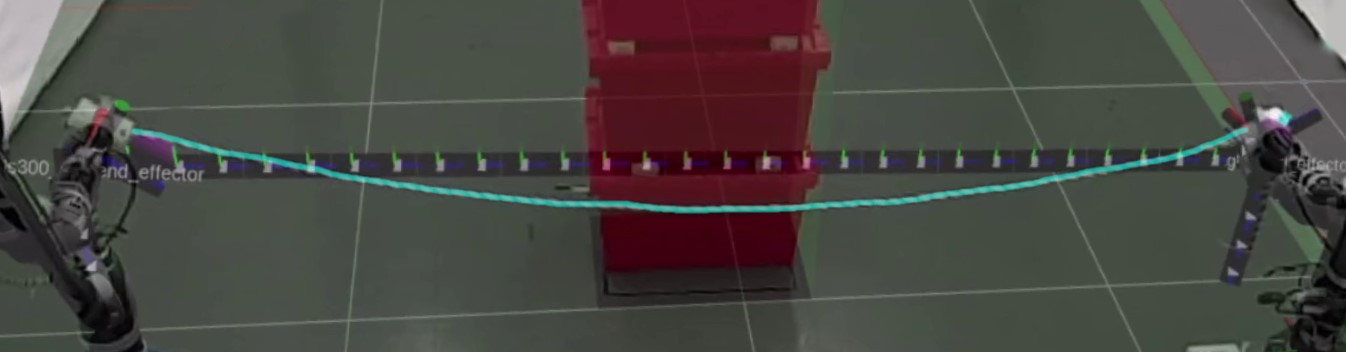}
        \caption{Initial configuration showing the mismatch. 
            The simulated DLO assumes a straight, taut state, while the actual rope (highlighted in cyan) is loose and sagging between the robots.}
        \label{fig:real-robot-rope-fail-reason}
    \end{subfigure}
    \\
    \vspace{0.2cm}
    \begin{subfigure}[h]{1.0\columnwidth}
        \centering
        \includegraphics[width=1.0\linewidth]{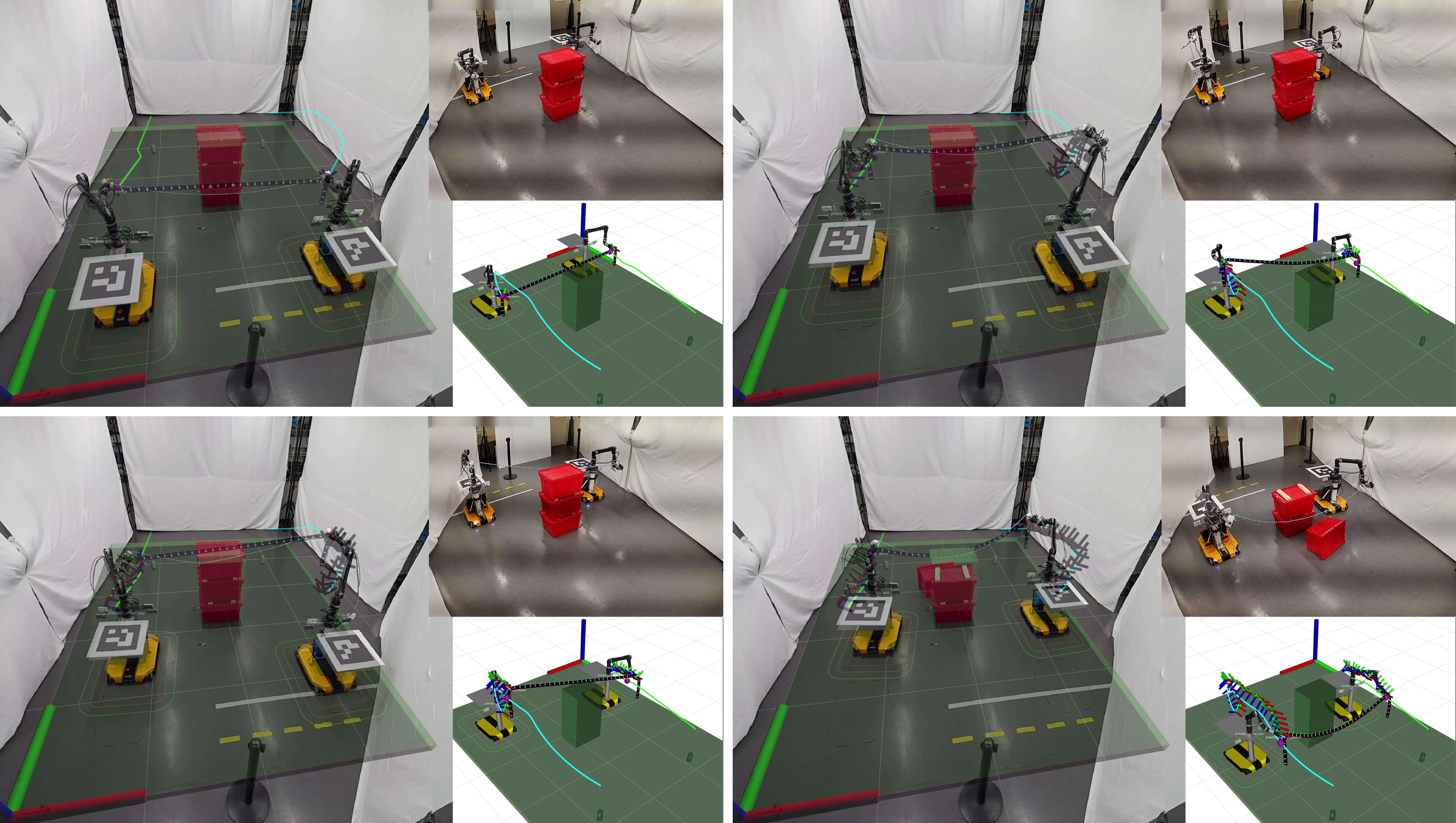}
        \caption{Snapshots at $t=\{0,53,62,77\}\,\mathrm{s}$.
        The execution shows the consequences of the state mismatch: the actual rope gradually hooks onto the obstacle by $t=62\,\mathrm{s}$, eventually knocking over the top tote box by $t=77\,\mathrm{s}$.} 
        \label{fig:real-robot-rope-fail-snapshots}
    \end{subfigure}
    \caption{%
    \textbf{Failure example} in real-robot execution with a \textbf{rope}, \textbf{caused by a mismatch between real and simulated DLO states}. 
    The simulation assumes a straight DLO based on robot poses and grasp points, but the actual rope is loose at the start, leading to a sagged state that persists throughout the execution. This mismatch causes obstacle hooking and task failure.
  }
    \label{fig:fig:real-robot-rope-fail}
    \vspace{-0.5cm}
\end{figure}

\paragraph{Real DLO - Simulated DLO mismatch}
We repeated the rope experiment with safety and planning enabled but reduced the obstacle avoidance margins. Additionally, the rope was deliberately left slack, yielding a sagged initial configuration, while the simulation assumed a taut DLO based on robot poses and grasp points (Fig.~\ref{fig:real-robot-rope-fail-reason}). This mismatch persisted throughout execution (Fig.~\ref{fig:real-robot-rope-fail-snapshots}). At $t=53\,\mathrm{s}$, just before traversing the obstacle, the simulation placed the rope above the obstacle while the actual rope was below the obstacle’s highest edge. By $t=62\,\mathrm{s}$, the rope had hooked onto the obstacle, although the controller is unaware since the simulation assumes the rope is clear. Finally, at $t=77\,\mathrm{s}$, the top tote box fell from the obstacle stack as the robots continued moving.
This failure highlights the significance of adhering to assumptions and safety measures (Section~\ref{subsubsec:Case-2-Real-Stiff-Tent-Pole-with-Real-Robots}). The real–simulated mismatch arose from compounding effects: pose inaccuracies at both the robot and end effector, localization errors, and DLO stiffness misestimations. Without a closed loop on the exact DLO shape, these errors can accumulate. Even if state mismatches are minimized, safety offsets remain essential in both planning and local control to accommodate delays or rapid robot motions violating quasi-static assumptions. Because DLO vision-based sensing is often occlusion-prone, hooking events can go undetected. Integrating real-time F/T data with the simulation stress estimates would allow the system to detect large, unexpected deviations between actual and simulated stress values, thus enabling an immediate stop as a safety measure.

\section{Conclusion}
\label{sec:conclusion}
\looseness=-1
This paper presented a novel, efficient approach for real-time manipulation of deformable linear objects (DLOs) using multiple robot agents—meeting a critical need in both industrial and domestic settings. Our framework integrates a standard planning pipeline with a real-time local controller based on Control Barrier Functions (CBFs), Position-Based Dynamics (PBD) simulations, and ROS. As demonstrated by extensive simulations and real-robot experiments in obstacle-rich environments, it ensures operational safety while offering a flexible, high-performance solution for various DLO types.
To the best of our knowledge, this is the first study to show multiple mobile manipulators collaboratively and smoothly controlling all degrees of freedom (DoFs) of a deformable object. Our planning pipeline efficiently generates guidance paths, outperforming state-of-the-art methods in time performance, while the local controller uses path tracking as a soft constraint and seamlessly transitions to tip control when necessary. By applying CBFs, the controller prevents collisions and avoids overstress in real time, relying on force/torque (F/T) readings rather than implicit geometric constraints. It effectively handles both positional and orientational aspects of the DLO.
Several limitations remain. The local controller lacks a closed loop on actual DLO states, and material properties and gravity are not explicitly considered in the global planner. These issues could be addressed by implementing a robust, real-time DLO state tracker to handle occlusions and sensor noise, and by extending the planner (e.g., TrajOpt) with custom cost functions within a Tesseract-based environment. Future work may also involve incorporating the full robot collision geometry, handling dynamic obstacles, and extending this framework to two-dimensional deformable objects such as fabrics, sheet metals, and composite layouts—further enhancing path feasibility, robustness, and versatility.
Despite these limitations, our extensive experiments confirm the proposed method’s efficiency and effectiveness. Overall, this study marks a significant step toward broader real-world applicability, providing a more flexible, tunable, and customizable solution than existing state-of-the-art methods. Such adaptability simultaneously improves safety and efficiency for tasks that require cooperative manipulation of deformable objects.


\addtolength{\textheight}{-2.40cm}  
\bibliographystyle{IEEEtran}
\bibliography{99_bibliography/bib}

 \ifANONREVIEW
\else
    \vspace{-22pt}
    \begin{IEEEbiography}[{\includegraphics[width=1in,height=1.25in,clip,keepaspectratio]{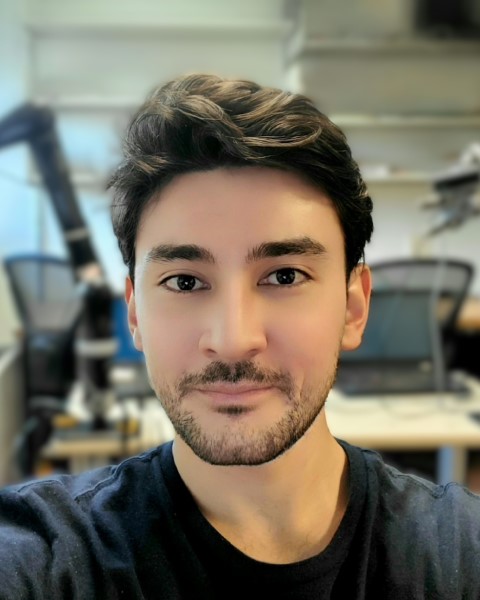}}]{Burak Aksoy}
    received the B.S. degrees in Mechatronics Engineering and Computer Science from Sabanci University, Istanbul, Turkey, in 2018. He is currently working toward the Ph.D. degree in Computer Systems Engineering at Rensselaer Polytechnic Institute (RPI), Troy, NY, USA. 
    His research interests include real-time collaborative multi-robot manipulation, with a focus on tasks involving deformable objects and complex manufacturing applications.
    \end{IEEEbiography}
    \begin{IEEEbiography}[{\includegraphics[width=1in,height=1.25in,clip,keepaspectratio]{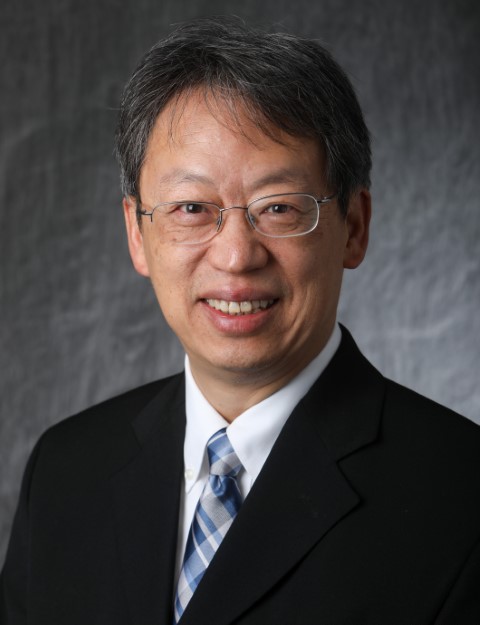}}]{John T. Wen}
    (M’76–F’01) received his B.Eng. from McGill University, M.S. from the University of Illinois, and Ph.D. from Rensselaer Polytechnic Institute (RPI). 
    He has held leadership roles at RPI, including Head of Electrical, Computer, and Systems Egnineering (2019-present), Head of Industrial and Systems Engineering (2013–2018) and Director of the Center for Automation Technologies and Systems (2005–2013). He led RPI's participation in the Advanced Robotics for Manufacturing (ARM) Institute and served on its Technical Advisory Council (2017-2020). His research focuses on control theory and applications in robotics, material processing, thermal management, biochronicity, and optomechatronics.
    \end{IEEEbiography}
    

\fi

\vfill

\end{document}